\documentclass[letterpaper]{article} 
\usepackage{aaai24}  
\usepackage{times}  
\usepackage{helvet}  
\usepackage{courier}  
\usepackage[hyphens]{url}  
\usepackage{graphicx} 
\usepackage{natbib}  
\usepackage{caption} 
\usepackage{algorithm}
\usepackage{algorithmic}

\usepackage{newfloat}
\usepackage{listings}
\DeclareCaptionStyle{ruled}{labelfont=normalfont,labelsep=colon,strut=off} 
\floatstyle{ruled}
\newfloat{listing}{tb}{lst}{}
\floatname{listing}{Listing}
\usepackage{adjustbox}
\title{Beyond TreeSHAP:\\Efficient Computation of Any-Order Shapley Interactions for Tree Ensembles
}
\author{
    Maximilian Muschalik\textsuperscript{\rm 1,}\equalcontrib,
    Fabian Fumagalli\textsuperscript{\rm 2,}\equalcontrib,
    Barbara Hammer\textsuperscript{\rm 2},
    Eyke Hüllermeier\textsuperscript{\rm 1}
}
\affiliations{
    \textsuperscript{\rm 1}LMU Munich, MCML Munich, D-80539 Munich, Germany\\
    \textsuperscript{\rm 2}Bielefeld University, CITEC, D-33619 Bielefeld, Germany\\
    maximilian.muschalik@lmu.de, ffumagalli@techfak.uni-bielefeld.de, bhammer@techfak.uni-bielefeld.de, eyke@.lmu.de
}

\usepackage{amsmath,amssymb,amsthm,mathtools}
\usepackage{mdframed}
\usepackage{booktabs}
\newtheorem{definition}{Definition}
\newtheorem{theorem}{Theorem}
\newtheorem{prop}{Proposition}
\newtheorem{lemma}{Lemma}

\newcommand{\fset}{N}
\newcommand{\fnum}{n}

\begin{document}

\maketitle

\begin{abstract}

While shallow decision trees may be interpretable, larger ensemble models like gradient-boosted trees, which often set the state of the art in machine learning problems involving tabular data, still remain black box models. As a remedy, the Shapley value (SV) is a well-known concept in explainable artificial intelligence (XAI) research for quantifying additive feature attributions of predictions. The model-specific TreeSHAP methodology solves the exponential complexity for retrieving exact SVs from tree-based models. Expanding beyond individual feature attribution, Shapley interactions reveal the impact of intricate feature interactions of any order. In this work, we present TreeSHAP-IQ, an efficient method to compute any-order additive Shapley interactions for predictions of tree-based models. TreeSHAP-IQ is supported by a mathematical framework that exploits polynomial arithmetic to compute the interaction scores in a single recursive traversal of the tree, akin to Linear TreeSHAP. We apply TreeSHAP-IQ on state-of-the-art tree ensembles and explore interactions on well-established benchmark datasets.

\end{abstract}

\section{Introduction}
Tree-based ensemble methods, in particular gradient-boosted trees \cite{Friedman.2001}, such as XGBoost \cite{Chen.2016} or LightGBM \cite{Ke.2017}, are among the most popular machine learning (ML) models and often achieve state-of-the-art (SOTA) performance on tabular data without extensive hyperparameter tuning \cite{ShwartzZiv.2022}.
These ensemble methods utilize intricate prediction functions by employing tree structures of high depth, thereby obstructing interpretation of the model's internal reasoning.
Yet, understanding a model's prediction is necessary for safe and reliable deployment, alongside addressing ethical and regulatory considerations \cite{Adadi.2022}.
Additive feature attributions, which split the individual features' contributions to the prediction, are a prevalent approach to improving the local interpretation of ML models \cite{Lundberg.2017,Covert.2021,Chen.2023}.
However, in complex real-world applications, such as bioinformatics \cite{Lunetta.2004,Boulesteix.2012,Winham.2012,Wright.2016} or language-related tasks \cite{Tsang.2020} features only attain meaningfulness when \emph{interacting} with other features.
In such scenarios, information about interactions complements additive feature attributions, which only show part of the picture \cite{Wright.2016}.

In this work, we are interested in model-specific local XAI measures for tree-based models, such as XGBoost.
In particular, the extension of predominant attribution measures based on the Shapley value (SV) \cite{Shapley.1953} to any-order additive Shapley-based interactions to explain single predictions locally.
Our work extends path dependent TreeSHAP \cite{Lundberg.2020}, which exploits the structure of trees to reduce time complexity from exponential to polynomial, to any-order Shapley-based interactions.

\begin{figure}[t]
    \centering
    \includegraphics[width=0.97\columnwidth]{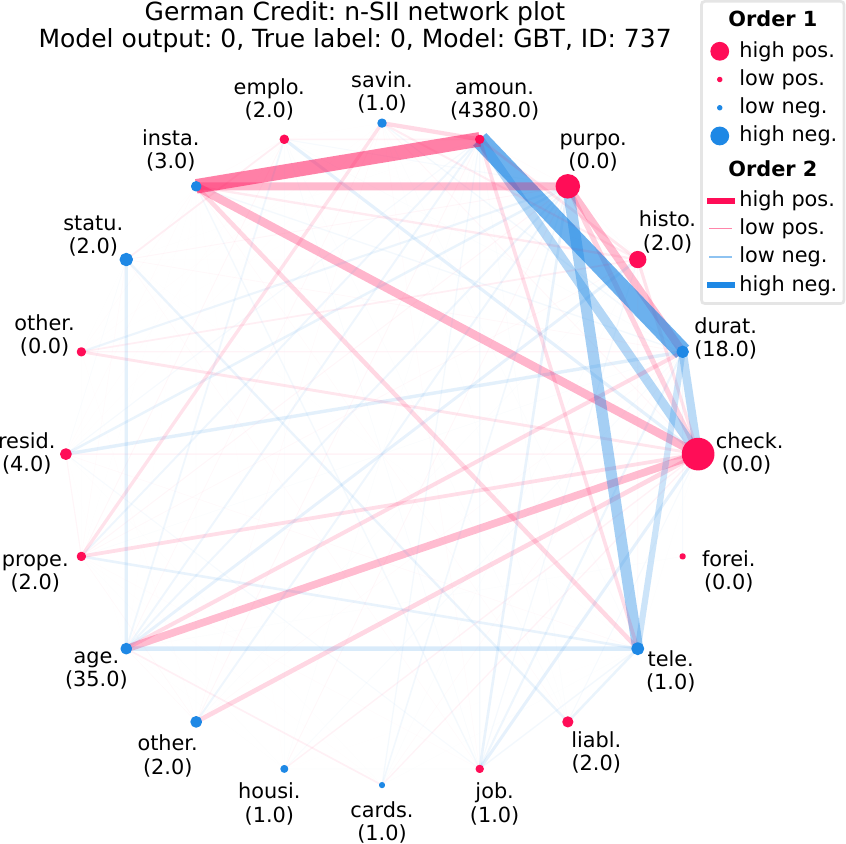}
    \caption{Network Plot after \cite{Inglis.2022} for a test instance of the \emph{German Credit} dataset for visualizing local feature attribution and interaction.}\label{fig_network_plot_intro}
\end{figure}

\begin{figure*}[t]
    \centering
    \begin{minipage}{0.19\textwidth}
        \centering
        \textbf{Maximum Order $s_0=1$ \\(SV)}
    \end{minipage}
    \hfill
    \begin{minipage}{0.8\textwidth}
        \includegraphics[width=\textwidth]{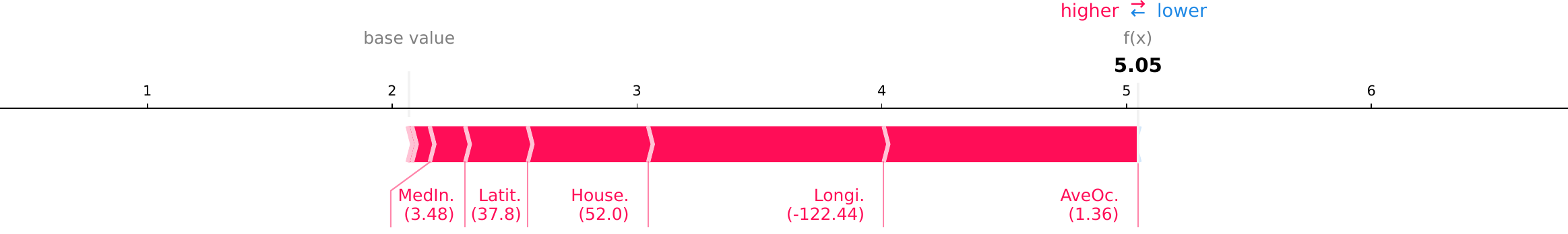}
    \end{minipage}
    \begin{minipage}{0.19\textwidth}
        \centering
        \textbf{Maximum Order $s_0=2$ \\ (n-SII)}
    \end{minipage}
    \hfill
    \begin{minipage}{0.8\textwidth}
        \includegraphics[width=\textwidth]{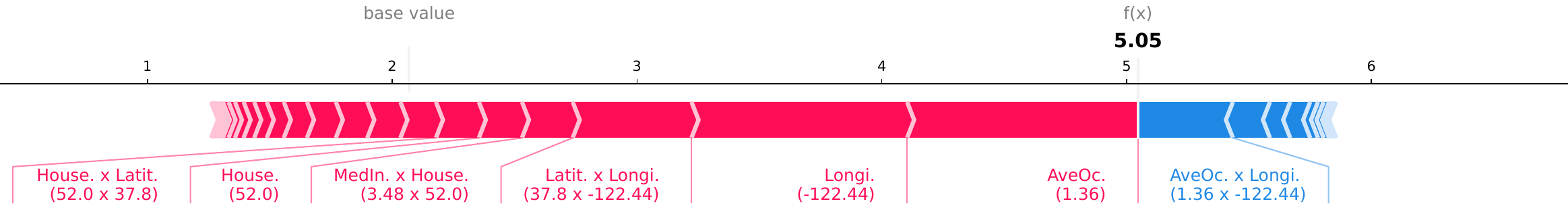}
    \end{minipage}
    \begin{minipage}{0.19\textwidth}
        \centering
        \textbf{Maximum Order $s_0=3$ \\ (n-SII)}
    \end{minipage}
    \hfill
    \begin{minipage}{0.8\textwidth}
        \includegraphics[width=\textwidth]{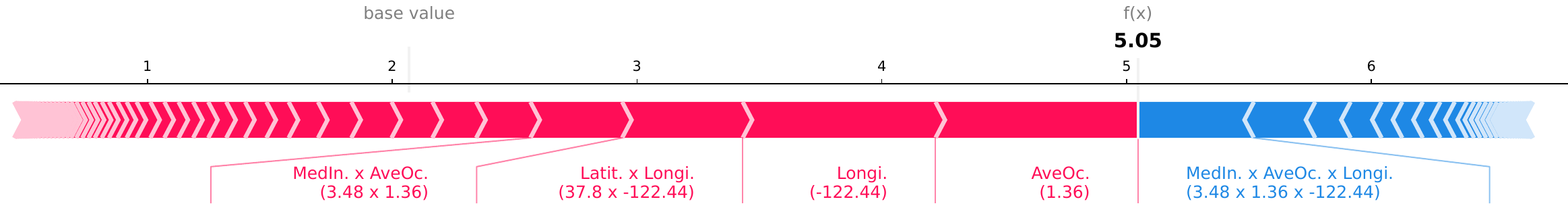}
    \end{minipage}
    \caption{Force plots of positive (red) and negative (blue) SVs and n-SII scores for an instance of the \emph{California} dataset
    The \emph{longit.} feature has a high contribution, describing the proximity to the ocean, which affects the price. \emph{TreeSHAP} ($s_0=1$) reveals this contribution. It also shows that \emph{latit.} contributed positively. \emph{TreeSHAP-IQ}, e.g. $s_0\geq2$, reveals that this contribution can be (mostly) attributed to the interaction \emph{latit. x longit.}, which reveals that the \emph{exact location}, and not \emph{latit.}, is meaningful.
    }
    \label{fig_force_plots}
\end{figure*}

\subsubsection{Related Work.}
The SV \cite{Shapley.1953} is a concept from cooperative game theory that has been proposed for model-agnostic explanations for local \cite{Strumbelj.2014,Lundberg.2017} and global \cite{Casalicchio.2018,Covert_Lundberg_Lee_2020} interpretation.
In a model-agnostic setting, efficient approximations techniques, based on Monte Carlo \cite{Castro.2009,Castro.2017,Kolpaczki.2023,Fumagalli.2023} or the representation of the SV as a constrained weighted least square problem \cite{Lundberg.2017,Covert.2021,Jethani.2022} have been proposed to overcome the exponential complexity.
For tree-based models the SV can be computed in polynomial time using TreeSHAP \cite{Lundberg.2020} with more efficient variants \cite{Yang.2021}. 
Linear TreeSHAP \cite{Yu.2022} establishes a theoretical foundation that connects the computation to polynomial arithmetic, achieving SOTA computational and storage efficiency.

Limitations of the SV due to correlations and interactions have been widely studied by \citet{Slack.2020}, \citet{Sundararajan.2020b}, and \citet{Kumar.2020,Kumar.2021}.
Extensions to interactions have been proposed with the \emph{Shapley Interaction Index} (SII) \cite{Grabisch.1999}, its aggregation as \emph{n-Shapley Values} (n-SII) \cite{Bord.2023}, the \emph{Shapley Taylor Interaction Index} (STI) \cite{Sundararajan.2020} and the \emph{Faithful Shapley Interaction Index} (FSI) \cite{Tsai.2022}.
All of these are subsumed in the broad class of the \emph{Cardinal Interaction Index} (CII) \cite{Grabisch.1999}.
Model-agnostic approximations have been proposed for general CIIs \cite{Fumagalli.2023}, STI \cite{Sundararajan.2020}, SII and for FSI \cite{Tsai.2022}.
Local pairwise interactions for tree-based models were computed by \citet{Lundberg.2020} and for interventional SHAP by \citet{DBLP:conf/aaai/ZernBK23}.

Other interaction scores were introduced by \citet{Tsang.2020}, \citet{Zhang.2021}, \citet{DBLP:conf/fat/PatelSZ21}, \citet{Harris.2022}, and \citet{Hiabu.2023}.
Interaction scores are further linked to functional decomposition \cite{Hooker.2004,Hooker.2007,Lengerich.2020,Herbinger.2023}.
For tree-based models, limitations of feature attribution measures \cite{Wright.2016}, and efficient implementations for interactions \cite{Lengerich.2020,Hiabu.2023} were discussed.

So far, any-order Shapley interactions have only been studied in a model-agnostic setting, where the exponential complexity problem is approximately solved.
Tree-based approaches have not considered the efficient computation of local any-order Shapley interactions.

\subsubsection{Contribution.}
Our main contributions include;
\begin{enumerate}
    \item \textit{TreeSHAP-IQ} (Section \ref{sec_treeshapiq}): An efficient algorithm for computing any-order SII scores for tree ensembles.
    TreeSHAP-IQ is supported by a mathematical framework based on polynomial arithmetic, akin to Linear TreeSHAP (Section \ref{sec_background}).
    \item \textit{Unified Framework}: Application of TreeSHAP-IQ to the broad class of any-order CIIs.
    \item \textit{Application}: We efficiently implement TreeSHAP-IQ on SOTA tree-based models, such as XGBoost, and showcase how interaction scores enrich single feature attribution measures on several benchmark datasets (Section \ref{sec_experiments}).
\end{enumerate}

\section{Local Shapley-Based Explanations}\label{sec_background}
Local Shapley-based explanations consider a model $f$ on an $\fnum$-dimensional feature space $\mathcal X$ with features $\fset := \{1,\dots,\fnum\}$.
The goal is to explain the prediction $f(x) \in \mathbb{R}$ for a selected explanation point $x \in \mathcal X$ and find an additive attribution $\phi = (\phi[1],\dots,\phi[\fnum]) \in \mathbb{R}^{\fnum}$, such that $f(x) = b_0 + \sum_{i \in \fset} \phi[i]$, where $b_0 \in \mathbb R$ is the baseline prediction, i.e. the prediction of $x$, if no feature information is available.
To compute a unique attribution score $\phi[i]$ for each feature $i \in \fset$, we extend the model with subsets of features $f: \mathcal X \times \mathcal P(\fset) \to \mathbb R$, where $\mathcal P(\fset)$ is the power set of $\fset$ and $f(x,T)$ refers to the prediction of $f$ at $x$, if only the features in $T \subseteq \fset$ are known.
In the following, if we omit the subset, then $T=\fset$, i.e. $f(x) := f(x,\fset)$. 
We further omit the explanation point $x$ if it is clear from context, and set $f(T) := f(x,T)$ and $b_0 := f(x,\emptyset)$.
The contribution of each feature $i \in \fset$ is then the SV \cite{Shapley.1953}
\begin{equation*}\label{eq_SV}
    \phi(f,i) := \sum_{T \subset \fset \setminus \{i\}} \frac 1 {\fnum \cdot \binom{\fnum-1}{\vert T \vert}}\Big[f(T \cup \{i\})-f(T)\Big].
\end{equation*}
The SVs define the unique attribution measure satisfying the following axioms: linearity (in $f$), symmetry (ordering does not impact $\phi$), dummy (no impact on $f$ implies $\phi(f,i)=0$) and efficiency $f(x) = b_0 + \sum_{i \in \fset}\phi(f,i)$ \cite{Shapley.1953}.

In many real-world applications, single feature importance scores are not sufficient to understand a model, where features become only meaningful when \emph{interacting} with others.
The SV does not give any information about such \emph{interactions} between two or more features.
The SII has been the first extension of the SV to interactions of feature subsets.

\begin{definition}[SII, \citeauthor{Grabisch.1999}~\citeyear{Grabisch.1999}]\label{def_SII}
The SII for an interaction $S \subseteq \fset$ is defined as
\begin{align*}
I^{\text{SII}}(f,S) := \sum_{T \subseteq \fset\setminus S} \frac{1}{(\fnum-\vert S \vert+1) \cdot\binom{\fnum-\vert S \vert}{\vert T \vert}} \delta_{S}(f,T),
\end{align*}
where $\delta_{S}$ is the S-derivative of $f$ for $T \subseteq \fset \setminus S$, i.e.
\begin{equation*}
    \delta_{S}(f,T) := \sum_{L\subseteq S}(-1)^{\vert S \vert - \vert L \vert}f(T \cup L).
\end{equation*}
\end{definition}

The SII is the unique attribution measure that fulfills the (generalized) linearity, symmetry and dummy axiom, as well as a novel recursive axiom that links higher to lower order interactions \cite{Grabisch.1999}.
In contrast to the SV, the SII does not fulfill the (generalized) efficiency axiom, which states that the sum of interaction scores (including $b_0$) up to a \emph{maximum order $s_0$} equals the model prediction $f(x)$.
This axiom is particularly useful in the ML context.
Recently, \citet{Bord.2023} proposed a specific aggregation, known as n-SII of order $s_0$, which yields a unique index that satisfies the (generalized) efficiency axiom.
A more general class constitutes the CII, where it was shown that every interaction index fulfilling the linearity, symmetry and dummy axiom can be represented as a CII \cite[Proposition 5]{Grabisch.1999}.
Other CIIs were proposed that introduce a unique interaction index of order $s_0$ and require the efficiency axiom directly, such as the STI \cite{Sundararajan.2020} or the FSI \cite{Tsai.2022}.
While the computation of the SV and SIIs are of exponential complexity, it has been shown that the complexity for the SV can be reduced to polynomial time in the case of tree-based models.

\subsection{The Shapley Value for Tree Ensembles}
For tree-based models the computational complexity of the SV can be drastically reduced by utilizing the additive tree structure.
Furthermore, there exists a natural way to handle missing features, which can be used to define the extended model $f(x,T)$.
For simplicity, we consider in the following a single decision tree, where ensembles of trees can be similarly computed due to the linearity of the SV.

\subsubsection{Notation.}
\begin{figure}[t]
    \centering
    \includegraphics[width=0.95\columnwidth]{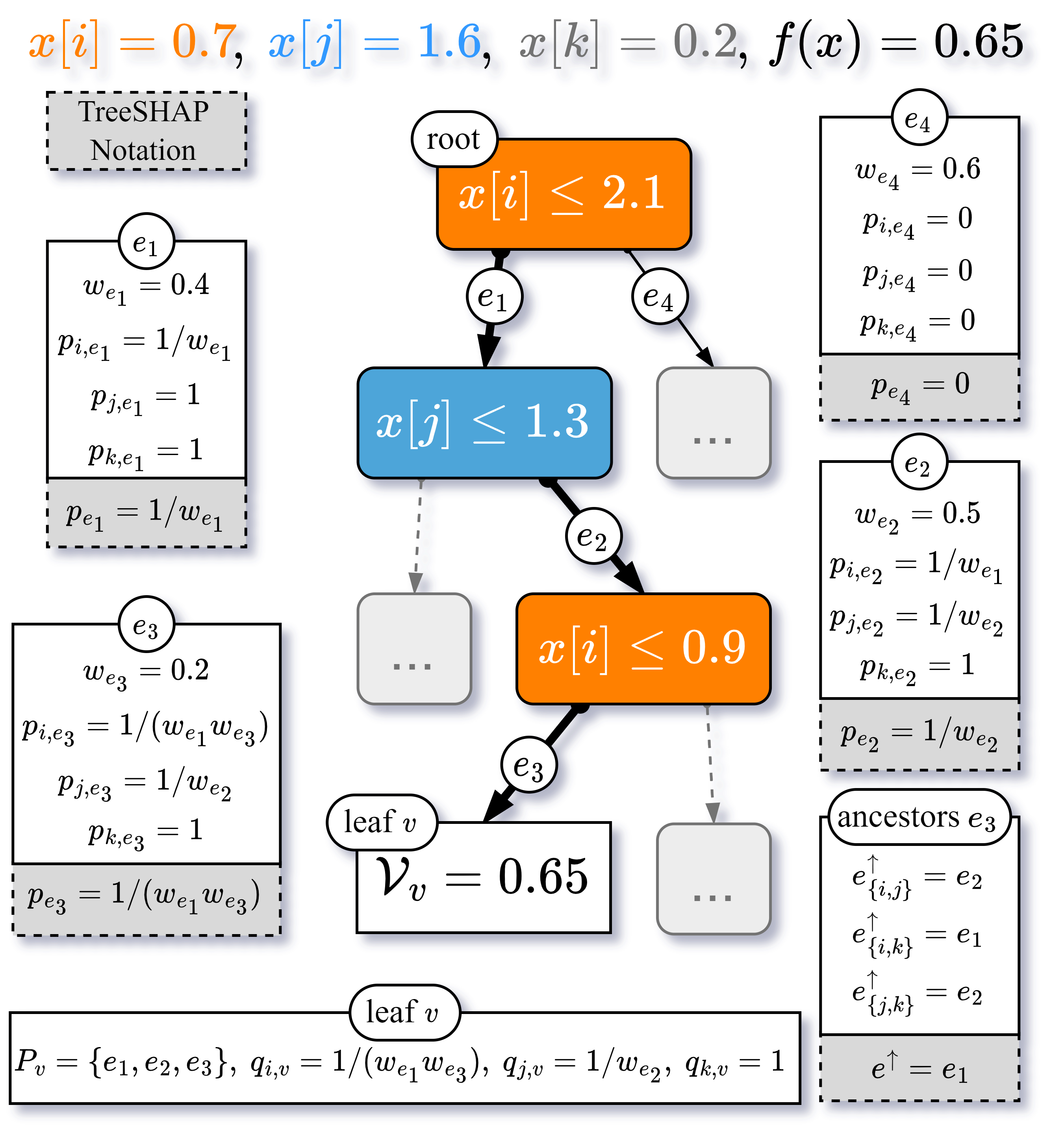}
    \caption{Notations in TreeSHAP-IQ and Linear TreeSHAP.}
    \label{fig_notation}
\end{figure}

We consider a decision tree $\mathcal T=(V,E)$ as a rooted directed tree with a set of vertices $V$, referred to as decision nodes, and edges $E$.
The root node is denoted as $r \in V$.
Each decision node consists of a split feature $i \in \fset$ with a threshold value and predictions $\mathcal V_v$ at the leaf nodes.
For each node $v \in V$, we let $P_v$ be the set of edges from the root node to $v$ and $\mathcal L(v)$ the set of leaf nodes reachable from $v$, where $\mathcal L(\mathcal T)=\mathcal L(r)$ is the set of all leaf nodes in the tree.
For every edge $e \in E$ going from $u$ to $v$, we denote $u$ as the tail of $e$ and $v$ as the head of $e$, $h(e)$.
We consider a weighted tree with weights $w_e \in (0,1)$ for every edge $e \in E$, which is defined as the proportion of observed data points at the tail of $e$, that split to the head of $e$.
Additionally, we label each edge $e \in E$ with the feature associated with the tail of $e$, i.e. the feature that was used to split the observations on the decision node at the source of $e$.
Further, $P_{i,v}$ and $E_i$ are the edges in $P_v$ and $E$ with label $i \in \fset$.
Our notation for decision trees is illustrated in Figure~\ref{fig_notation}.

We also require polynomial arithmetic and refer to the set of polynomials with maximum degree $d$ and coefficients in $\mathbb{R}$ as $\mathbb{R}[x]_d$.
Polynomial multiplication is denoted with $\odot$ and division with $\lfloor \frac a b \rfloor$ or $\lfloor a / b \rfloor$.
We denote with $\langle x,y \rangle$ the inner product of two vectors $x,y \in \mathbb{R}^d$ and refer to the inner product of the coefficients, if polynomials are considered.

\subsubsection{Extended Model $f(x,T)$ for Decision Trees.} 
A decision tree can be decomposed into distinct decision rules $R^v: \mathcal X \to \mathbb R$ for each leaf $v \in \mathcal L(\mathcal T)$, which predict $\mathcal V_v$, if $x$ reaches $v$ and zero otherwise.
Note that each $R^v$ induces a subspace of $\mathcal X$ at which the prediction of $f$ is constant.
The decision tree is thus given as $f(x) = \sum_{v \in \mathcal L(\mathcal T)} R^v(x)$.
We now define $R^v_T$, the prediction rule restricted to a set of \emph{active} features $T \subseteq \fset$, where the remaining are considered to be unknown.
If the split feature is unknown, we split based on the weights $w_e$, which is a common practice \cite{Yu.2022}.
If all features are unknown, we define $R^v_\emptyset := R^v_\emptyset(x) := \mathcal V_v \prod_{e \in P_v} w_e$.
When adding feature $i \in \fset$ to the active set, the product of associated weights is replaced by the split criterion.
This is formalized as a recursive property $R^v_{T \cup \{i\}} = q_{i,v}(x) R^v_T(x)$, where $q_{i,v}$ is the marginal effect of adding $i$ to the active set.
To define $q_{i,v}$, we let $x \in \pi_i(R^v)$, if $x[i]$ satisfies each split criterion regarding features $i$ in the path of $v$, i.e. $\pi_i(R^v)$ is the region of feature $i$ in the induced subspace of $\mathcal X$ by $R^v$.
For $v$ the marginal effect of adding feature $i \in \fset$ is then defined as
\begin{equation}
    q_{i,v}(x) := \mathbf{1}(x \in \pi_i(R^v))\prod_{e \in P_{i,v}} \frac{1}{w_e},
\end{equation}
where $\mathbf 1(\cdot)$ is the indicator function.
Furthermore, for $P_{i,v}=\emptyset$ we define $q_{i,v}=1$.
The restricted rule is thus defined as
\begin{align}\label{eq_R_def}
    &R^v_T(x) := R^v_\emptyset \prod_{j \in T} q_{j,v}(x).
\end{align}
For a tree $\mathcal T_f$ and $T\subseteq \fset$ the restricted model at $x$ is then
\begin{align*}
    f(T) := f(x,T) := \sum_{v \in \mathcal L(\mathcal T_{f})} R^v_T(x).
\end{align*}
In the following, we omit the argument $x$ in the notation.
We proceed to compute the SV of $f(T)$, which is known as \emph{path dependent} TreeSHAP \cite{Lundberg.2020}.

\subsubsection{Linear TreeSHAP.}
TreeSHAP exploits the tree structure to compute the SV in polynomial time \cite{Lundberg.2020}.
Linear TreeSHAP improved this computation and provided a theoretical framework by linking the computation to polynomial arithmetic \cite{Yu.2022}.
Plugging (\ref{eq_R_def}) into the definition of the SV and using the fact that $q_{j,v}-1=0$, if a feature does not appear in the path, yields
\begin{equation}\label{eq_R_pre}
    \phi(R^v,i) = (q_{i,v}-1) \sum_{T \subseteq \mathcal{F}(R^v)\setminus \{i\}} \frac {R_\emptyset^v} {\fnum \cdot \binom{\fnum-1}{\vert T \vert}} \prod_{j \in T} q_{j,v},
\end{equation}
where $\mathcal{F}(R^v)$ is the set of all features that appear in $R^v$.
It was shown that this sum can be efficiently stored using the coefficients of a specific polynomial.
\begin{definition}[Summary Polynomial (SP), \citeauthor{Yu.2022}~\citeyear{Yu.2022}]    
The SP of leaf node $v$ is $G^\text{SP}_{v}(y) := R^v_\emptyset \prod_{j \in \mathcal{F}(R^v)} \left(q_{j,v}+y\right)$.
\end{definition}

For feature $i \in \fset$, $R_\emptyset^v \sum_{S \subset \mathcal{F}(R)\setminus\{i\}}^{\vert S \vert = k} \prod_{j \in S} q_{j,v}$ is the coefficient of $y^{d-k-1}$ in $\frac G {q_{i,v} + y}$, where $d := \vert \mathcal{F}(R^v) \vert$ is the number of features in each path.
Note that this corresponds to the non-weighted terms in the sum of (\ref{eq_R_pre}) for $k=0,\dots,d-1$.
The SV of a single decision rule can thus be represented as
\begin{equation}\label{eq_SV_R}
    \phi(R^v,i) = (q_{i,v} - 1) \psi\left(\left\lfloor\frac{G_v}{q_{i,v}+y}\right\rfloor\right),
\end{equation}
where $\psi: \mathbb{R}[x]_d \to \mathbb{R}$ is a function that properly weights the coefficients, such that it corresponds to the sum in (\ref{eq_R_pre}).
It is formally defined \cite{Yu.2022} as
\begin{align}\label{eq_psi}
    \psi_d(A) := \frac{\langle A, B_d \rangle}{d+1} \text{ with } B_d(y) := \sum_{k=0}^d \binom{d}{k}^{-1} y^k.
\end{align}
We write $\psi(A)=\psi_d(A)$, where $d$ is the degree of $A$. 
It was then shown that $\psi$ is additive and scale invariant.
\begin{prop}[\citeauthor{Yu.2022} \citeyear{Yu.2022}]
For $\psi$ and $p,q \in \mathbb{R}[x]_d$,
\begin{equation*}
    \psi_d(p+q) = \psi_d(p) + \psi_d(q) \text{ and } \psi\left( p \odot (1+y)^k\right) = \psi(p).
\end{equation*} 
\end{prop}
Using (\ref{eq_SV_R}) based on leaf nodes, a representation of the SV in terms of edges is presented, which is explicitly computed by recursively traversing the tree.
For this representation, the SP is extended to every edge in the path of $P_{i,v}$ as
\begin{equation*}
    G_u := \bigoplus_{v \in \mathcal L(u)} G_v \text{ with } G^1 \oplus G^2 := G^1 + G^2 \odot (1+y)^{d_1-d_2},
\end{equation*}
where the order is such that $d_1 > d_2$, i.e. $\oplus$ is an operation on the set of polynomials $\oplus: \mathbb{R}[x]_{d_1} \times \mathbb{R}[x]_{d_2} \to \mathbb{R}[x]_{\max(d_1,d_2)}$ that sums the polynomial while scaling them to the same degree.
Note that due to the properties of $\psi$, we have $\psi(G_u) = \sum_{v \in \mathcal L(u)} \psi(G_v)$.
For edge $e \in E$ and its feature $i$, we further introduce the inter-path value of $q_{i,v}$ as

\begin{equation*}
    p_{e} := \mathbf{1}\left(x \in \pi_{h(e)}\right) \prod_{e' \in P_{i,h(e)}} \frac 1 {w_{e'}}.
\end{equation*}
Note that $p_{e^*} = q_{i,v}$ if $e^*$ is the last edge in $P_{i,v}$.
An edge-based representation of the SV is then provided.
\begin{theorem}[\citeauthor{Yu.2022} \citeyear{Yu.2022}]\label{thm_linear_treeshap}
Let $i \in \fset$ and denote for $e$ the closest ancestor in the set $E_i$ by $e^\uparrow$, where $e^\uparrow = \, \perp$ and $p_{i,\perp} = 1$ in case it does not exist. Then,
\begin{align*}
    \phi(f,i) = &\sum_{e \in E_i}\left(p_{e} -1\right) \psi\left(\left\lfloor\frac{G_{h(e)}}{y+p_{e}}\right\rfloor\right) \\&-  \left(p_{{e^\uparrow}} -1\right) \psi\left(\left\lfloor\frac{G_{h(e)} \odot (y+1)^{d_{e^\uparrow}-d_e}}{y+p_{e^\uparrow}}\right\rfloor\right).
\end{align*}
\end{theorem}
Using this edge-based representation, Linear TreeSHAP computes the SV by traversing once through the tree.
To improve efficiency, the SP is stored in a multipoint interpolation form.
For more details, we refer to Appendix~B.

\section{TreeSHAP-IQ: Computation of Local Shapley Interactions for Tree Ensembles}\label{sec_treeshapiq}
Computing the exact SV for tree ensembles can reliably quantify the impact of single features on the model's predictions.
However, in many applications, certain features become only meaningful when \emph{interacting} with other features.
In this case, the SV is not sufficient to understand how the model predicts, and more complex explanations in terms of Shapley interactions are necessary.
In the following, we propose TreeSHAP Interaction Quantification (TreeSHAP-IQ), an efficient algorithm for computing any-order SII scores, which follows naturally by extending the SP to interactions.

TreeSHAP-IQ can further be applied to the broad class of CIIs \cite{Grabisch.1999}, which we briefly discuss in Section \ref{sec_extension_cii}.
All proofs are deferred to Appendix~A.

\subsection{Theoretical Foundation of TreeSHAP-IQ}\label{sec_treeshapiq_theory}
We now present the theoretical foundation of TreeSHAP-IQ.
The notations in this section extend on Linear TreeSHAP \cite{Yu.2022} and are illustrated in Figure~\ref{fig_notation}.
We compute the S-derivative for $R^v$ and $T \subseteq \fset \setminus S$ as
\begin{equation}\label{eq_sii_delta}
    \delta_{S}(R^v,T) = R^v_T \sum_{L \subset S} (-1)^{\vert S \vert - \vert L \vert}\prod_{j \in L} q_{j,v},
\end{equation}
which follows from (\ref{eq_R_def}) and the recursive property.
We thus represent the SII for a single decision rule as follows.
\begin{prop}\label{prop_SII_R}
For a leaf $v$ in $\mathcal T_f$, it holds $I^{\text{SII}}(R^v,S)=$
    \begin{align*}
       \left(\sum_{L \subseteq S} (-1)^{\vert S \vert - \vert L \vert}\prod_{j \in L} q_{j,v}\right)\psi\left(\left\lfloor\frac{G_v}{\prod_{j \in S}(q_{j,v}+y)}\right\rfloor\right).
    \end{align*}
\end{prop}
Proposition \ref{prop_SII_R} yields a compact representation in terms of leaf nodes and decision rules, which reduces to the representation of (\ref{eq_SV_R}) for single feature subsets.
Similar to Linear TreeSHAP, the representation of SII in terms of leaf nodes is not suitable for efficient computation.
We thus again establish an edge-based representation, similar to Theorem \ref{thm_linear_treeshap}.
By Proposition \ref{prop_SII_R}, the computation of an interaction for a subset $S \subset \fset$ requires knowledge of all $q_{i,v}$ with $i \in S$, which have to be tracked during the traversal of the tree.
We thus first extend the inter-path values $p_{e}$ to every feature as
\begin{equation*}
    p_{i,e} := \mathbf{1}\left(x \in \pi_{i,h(e)}\right) \prod_{e' \in P_{i,h(e)}} \frac 1 {w_{e'}},
\end{equation*}
where $x \in \pi_{i,u}$ if $x[i]$ satisfies each decision criterion in $P_{i,u}$.
Note that $p_{j,e}=p_e$ and $\pi_{j,h(e)}=\pi_{h(e)}$, if $j$ is the label of $e$.
Our goal in the following is to provide an algorithm similar to Linear TreeSHAP that traverses the decision tree once and recursively computes the interaction scores.
The SP thereby remains unchanged, but we introduce further polynomials of order $\vert S \vert$ to efficiently maintain the sum as well as the denominator in Proposition \ref{prop_SII_R}.
\begin{definition}[Interaction Polynomial (IP)]
The IP of $S \subset \fset$ and edge $e$ is $H^\text{IP}_{S,e}(y) := \prod_{j \in S} (p_{j,e} - y)$.
\end{definition}
Note that the coefficient of $y^k$ in $H_{S,e}$ is exactly $\sum_{L \subset S}^{\vert L \vert=\vert S \vert - k} (-1)^{\vert L \vert} \prod_{j \in L} p_{j,e}$ for $k=0,\dots,\vert S \vert$.
Therefore, the sum of the coefficients of the IP equals the sum in (\ref{eq_sii_delta}).
We thus define the coefficient sum.

\begin{definition}[Coefficient sum $\kappa$]
We define the function $\kappa_d: \mathbb{R}[x]_d \to \mathbb{R}$ as $\kappa_{d}(A) := \langle A , y^d+\dots+y+1\rangle$. 
We write $\kappa(p) = \kappa_d(p)$, where $d$ is the degree of $p$.
\end{definition}

Applying $\kappa$ to the IP yields the following properties.

\begin{prop}\label{prop_kappa}
    For the sum of coefficients of the IP, it holds 
    \begin{align}
        \kappa(H^\text{IP}_{S,e}) = \sum_{L \subseteq S} (-1)^{\vert S \vert - \vert L \vert}\prod_{j \in L} p_{j,e}.
    \end{align}
    If there exists $j \in S$ with $p_{j,e}=1$, then $\kappa(H^\text{IP}_{S,e})=0$.
\end{prop}
Proposition \ref{prop_kappa} shows that $\kappa(H^\text{IP}_{S,e})$ corresponds to the edge-based representation of the sum in Proposition \ref{prop_SII_R}.
If $e$ is the last edge in $P_{j,v}$, then $p_{j,e} = q_{j,v}$ for all $j \in \fset$ and thus $\kappa(H^\text{IP}_{S,e})$ retrieves the sum in Proposition \ref{prop_SII_R}.
Furthermore, if $p_{j,e}=1$, then it is intuitive that all inter-path contributions with $j \in S$ are zero, since $j$ does not impact the model's prediction in this part of the tree.
This property allows us to update interaction scores only if all features of the subset have occurred in the path.
We further describe the quotient in Proposition~\ref{prop_SII_R} using another polynomial of order $\vert S \vert$.

\begin{definition}[Quotient Polynomial (QP)]
The QP of $S \subset \fset$ and edge $e$ is $H^\text{QP}_{S,e}(y) := \prod_{j \in S} (p_{j,e} + y)$.
\end{definition}

If $e^*_S$ is the last edge in $P_v$ of leaf node $v$ that contains any feature of $S$, then $p_{j,e^*_S}=q_{j,v}$ for every $j \in S$ and hence we can rewrite Proposition \ref{prop_SII_R} using Proposition \ref{prop_kappa} as
\begin{equation}\label{eq_SII_R_poly}
     I^{\text{SII}}(R^v,S) = \kappa(H^\text{IP}_{S,e^*_S}) \psi\left(\left\lfloor G_v/ H^\text{QP}_{S,e^*_S}\right\rfloor\right).
\end{equation}

Clearly, Proposition \ref{prop_SII_R} reduces to (\ref{eq_SV_R}) for the case of the SV.
In contrast to the SV, the edge-based computation includes all inter-path values of $p_{j,e}$ with $j \in S$.
To extend Theorem \ref{thm_main}, we therefore need to extend the notion of ancestor edges to ancestors with respect to a subset $S \subset \fset$.

\begin{prop}\label{prop_SII_R_poly}
     For a decision rule $R^v$ of a leaf node $v$ and a subset $S \subset \fset$, let $P_{S,v} := \bigcup_{i \in S} P_{i,v}$ and  $e^\uparrow_S$ as the closest ancestor of $e$ in $P_{S,v}$. The SII of $R^v$ is then given by
\begin{align*}
    I^{\text{SII}}(R^v,S) = &\sum_{e \in P_{S,v}} \kappa(H^\text{IP}_{S,e}) \psi\left(\left\lfloor\frac{G_v \odot (y+1)^{d_e-d_v}}{ H^\text{QP}_{S,e}}\right\rfloor\right) \\&-   \kappa(H^\text{IP}_{S,e_S^\uparrow}) \psi\left(\left\lfloor\frac{G_v \odot (y+1)^{d_{e^\uparrow_S}-d_v}}{H^\text{QP}_{S,e_S^\uparrow}}\right\rfloor\right).
\end{align*}  
\end{prop}

Using Proposition \ref{prop_SII_R_poly}, we can state our main theorem.

\begin{theorem}\label{thm_main}
    For $S \subset \fset$, let $E_S := \bigcup_{i\in S}E_i$ be the set of edges that split on any feature in $S$, and denote the closest ancestor of $e$ in $P_{S,v}$ as $e_S^\uparrow$.
    The SII is then computed as
\begin{align*}
    I^{\text{SII}}(f,S) = &\sum_{e \in E_S} \kappa(H^\text{IP}_{S,e}) \psi\left(\left\lfloor\frac{G_{h(e)}}{ H^\text{QP}_{S,e}}\right\rfloor\right) \\&-   \kappa(H^\text{IP}_{S,e_S^\uparrow}) \psi\left(\left\lfloor\frac{G_{h(e)} \odot (y+1)^{d_{e^\uparrow_S}-d_e}}{H^\text{QP}_{S,e_S^\uparrow}}\right\rfloor\right).
\end{align*}
\end{theorem}

Note that for $S=\{i\}$, Theorem \ref{thm_main} reduces to Theorem \ref{thm_linear_treeshap}.

\subsubsection{Implementation of TreeSHAP-IQ.}
Theorem \ref{thm_main} allows for an efficient computation of the SII, with the SP being handled alike to Linear TreeSHAP.
The IQ and QP are updated for each interaction subset that contains the feature of $e$.
We again use the multipoint interpolation form to store and update the polynomials $G_v, H^\text{IP}_{S,e}$, and $H^\text{QP}_{S,e}$.
TreeSHAP-IQ traverses the decision tree once for every explanation point.
At each edge (decision node), TreeSHAP-IQ updates all interactions that contain the currently encountered feature,
$\binom{\fnum-1}{\vert S \vert - 1}$ in total.
However, the update can be restricted to those interactions, where all features have been observed in the path.
We refer to Appendix~B for more details.

\subsubsection{Complexity of TreeSHAP-IQ}
Consider $m$ explanation points, $\ell_{\mathcal{T}} := \vert \mathcal L(\mathcal{T}) \vert$ as the number of leaves and $d_{\max}$ as the maximum depth of the tree.

Linear TreeSHAP has a computational complexity of $\mathcal{O}(m \cdot \ell_{\mathcal{T}} \cdot d_{\max})$ and storage complexity of $\mathcal O(d_{\max}^2 + \fnum)$ \cite{Yu.2022}.
We now consider the complexity of TreeSHAP-IQ, if all interactions of order $s := \vert S \vert$ are computed.
In contrast to Linear TreeSHAP and the SP, where only the current feature value has to be updated, TreeSHAP-IQ needs to update the IP, the QP and the interaction scores for all interaction subsets that contain the currently observed feature.
This increases the computational complexity by a factor of $\binom{\fnum-1}{s-1}$.
Furthermore, all interaction scores have to be stored, requiring storage of $\binom{\fnum}{s}$.
To store the IQ and QP, we require further a storage capacity of $\mathcal O(d_{\max}^2\cdot\binom{\fnum}{s})$.
The computational complexity is thus summarized as follows.
\begin{mdframed}
\textbf{TreeSHAP-IQ complexity for the SII of order $s$}
\begin{align*}
 &\text{Computational Complexity} &\text{Storage Complexity} 
 \\
 &\mathcal{O}\left(m \cdot \ell_{\mathcal T} \cdot d_{\max} \cdot \binom{\fnum-1}{s-1}\right)
 &\mathcal{O}\left(d_{\max}^2 \cdot \binom{\fnum}{s}\right)
\end{align*}    
\end{mdframed}

For the computation of the SV, the computational complexity of TreeSHAP-IQ is similar to Linear TreeSHAP.
The storage capacity is increased by a factor $n$, as we store the IP and QP for every feature.
Moreover, for pairwise interactions, TreeSHAP-IQ mirrors the complexity of the computation proposed by \citet{Lundberg.2020} using Linear TreeSHAP.
However, our method distinguishes itself by relying on a single initialization of the tree parameters.

\subsection{Extending TreeSHAP-IQ to General CIIs}\label{sec_extension_cii}
TreeSHAP-IQ can be extended to the broad class of CIIs.
A CII is defined as $I^{\text{CII}}(f,S) := \sum_{T \subseteq \fset \setminus S} w^{\text{CII}}_{s}(\vert T \vert) \delta_S(f,T)$ with non-negative weights $w_s^{\text{CII}}$ that depend on the interaction order $s := \vert S \vert$ \cite{Grabisch.1999,Fumagalli.2023}.
This includes other approaches of extending the SV to interactions, such as STI and FSI, as well as Banzhaf interactions \cite{DBLP:conf/fat/PatelSZ21}.
Observe from the proofs, that different weights in CIIs solely impact the SP, and in particular $\psi$.
To extend the SP for CIIs, we let $d := \vert \mathcal{F}(R^v) \vert$ and scale $G_v$ to the degree of $\fnum$, which does not impact $\psi$ due to the scale invariance.
We then observe
\begin{align*}
 \psi\left(\left\lfloor \frac{G_v \odot (1+y)^{\fnum - d}}{\prod_{j \in S}(1+q_{j,v})}\right\rfloor\right)
 =  R_\emptyset^v \sum_{T \subseteq \fset \setminus S} w^{\text{SII}}_{s}(t) \prod_{j \in S} q_{j,v},
\end{align*}
where $w^{\text{SII}}_s(t)$ is the CII weight for SII, cf. Definition \ref{def_SII}.
Recall from (\ref{eq_psi}) that these weights are retrieved from the polynomial $B_{\fnum-s}$.
Thus, we generalize $\psi_d^{\text{CII}}: \mathbb{R}[x]_d \to \mathbb{R}$ to
\begin{align*}
    \psi^{\text{CII}}_d(A) := \langle A, W_d \rangle \text{ with } W^{\text{CII}}_{d}(y) := \sum_{k=0}^d w^{\text{CII}}_{\fnum-d}(k) y^k.
\end{align*}
If $G_v$ is scaled to degree $\fnum$, then $\psi^{\text{CII}}_d$ is always evaluated with a polynomial of degree $d = \fnum - \vert S \vert$.
Further, note that for SII, we have $\psi_d(A) \equiv \psi^{\text{SII}}_d(A)$, where the quotient $d+1$ is included in the weights, i.e. $W^{\text{SII}}_{d}(y)= B_d/(d+1)$.

\subsubsection{Implementation of CIIs in TreeSHAP-IQ}
Using $\psi^{\text{CII}}$, any CII can be computed by TreeSHAP-IQ.
In contrast to $\psi \equiv \psi^{\text{SII}}$, the scale invariance does not hold for CIIs.
Therefore, the SP cannot be reduced to the degree $d := \vert \mathcal{F}(R^v) \vert$.
However, if we maintain the SP at the maximum degree $\fnum$, then all previous results apply.
If the SP is stored in multipoint interpolation form, then this merely requires a multiplication with the corresponding term of $(y+1)^{\fnum-d}$, which can be efficiently precalculated.
Thus, the computational complexity is not affected by this extension.
Provided $d_{\max} \geq \fnum$, the storage complexity is not affected either.

\begin{table}[t]
    \centering
    \adjustbox{max width=\columnwidth}{\begin{tabular}{@{}lcccc@{}}
    \toprule
    \textbf{Datasets} & \textbf{\# Instances} & \textbf{\# Features} & \textbf{Target} & \textbf{Speed-Up}\\ \midrule
    \emph{Credit} & $1\,000$ & $20$ & $\{0,1\}$ & $\sim 10^4$\\
    \emph{Bank} & $45\,211$ & $16$ & $\{0,1\}$ & $\sim 10^3$\\
    \emph{Adult} & $45\,222$ & $14$ & $\{0,1\}$ & $\sim 10^3$\\
    \emph{Bike} & $17\,379$ & $12$ & $\mathbb{R}$ & $\sim 10^1$\\ 
    \emph{COMPAS} & $6\,172$ & $11$ & $\{0,1\}$ & $\sim 10^2$\\
    \emph{Titanic} & $891$ & $9$ & $\{0,1\}$ & $\sim 10^1$\\
    \emph{California} & $20\,640$ & $8$ & $\mathbb{R}$ & $\sim 1$\\    
    \bottomrule
    \end{tabular}
    }
    \caption{Overview of datasets and speed-up compared to a naive computation}\label{tab_datasets}
\end{table}

\begin{figure}[t]
    \centering
    \includegraphics[width=\columnwidth]{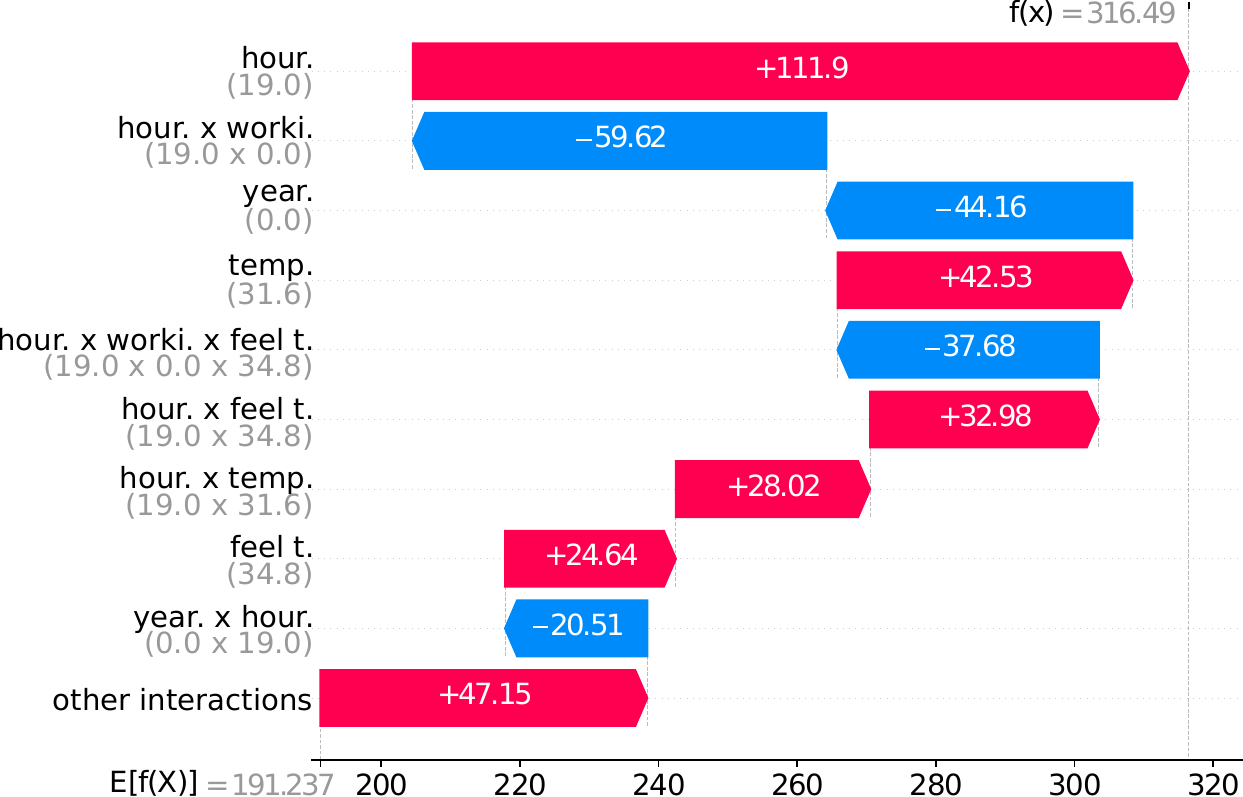}\caption{Waterfall chart for n-SII scores with $s_0=3$ and a prediction of the \emph{Bike} regression dataset.}\label{fig_waterfall}
\end{figure}

\section{Experiments}\label{sec_experiments}

We apply TreeSHAP-IQ\footnote{All experimental code and the technical appendix can be found at: \url{github.com/mmschlk/TreeSHAP-IQ}.} on XGBoost (XBG) \cite{Chen.2016}, gradient-boosted trees (GBTs), random forest (RF), and decision tree (DT) algorithms on the \emph{German Credit} \cite{Hofmann.1994}, \emph{Bank} \cite{Moro.2011}, \emph{Adult Census} \cite{Kohavi.1996}, \emph{Bike} \cite{FanaeeT.2014}, \emph{COMPAS} \cite{Angwin.2016}, \emph{Titanic} \cite{Dawson.1995}, and \emph{California} \cite{Kelley.1997} datasets, see Table~\ref{tab_datasets}.
For further experimental results, including a run-time analysis and detailed information on the datasets, models, and pre-processing steps, we refer to Appendix~C.
We compute additive interactions for single predictions using TreeSHAP-IQ with n-SII of different order.

\begin{figure}[t]
    \centering
    \includegraphics[width=\columnwidth]{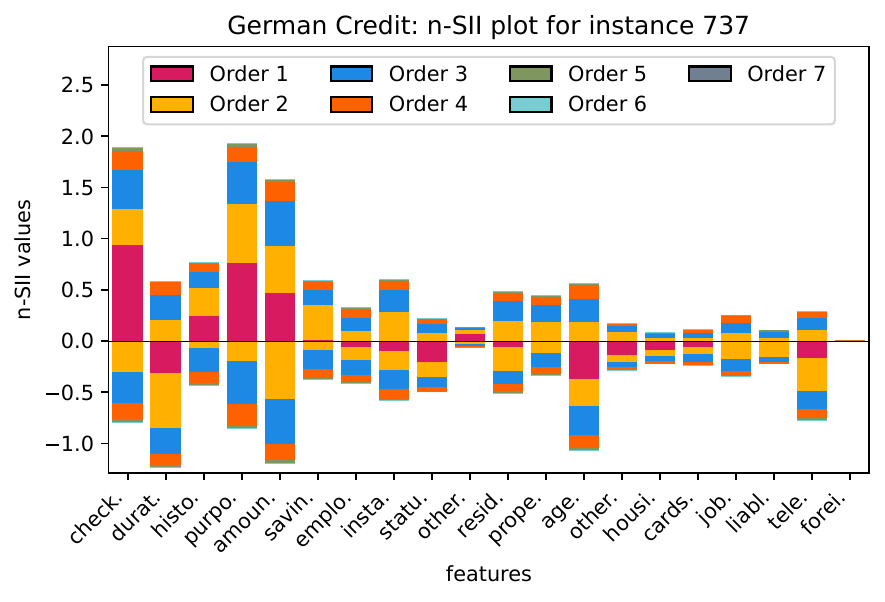}
    \caption{Visualization of positive and negative n-SII scores per feature with $s_0=7$ for an observation in \emph{German Credit}.}
    \label{fig_n_SII}
\end{figure}

\paragraph{TreeSHAP-IQ Reveals Intricate Feature Interactions.}
Using TreeSHAP-IQ, we examine the model's prediction based on higher order interaction effects. 
We distinguish n-SII scores that positively (red) and negatively (blue) impact the prediction.
In Figure~\ref{fig_network_plot_intro}, we visualize n-SII with $s_0=2$.
The width of the network vertices (order 1) and the network edges (order 2) describes the absolute value of the corresponding n-SII scores.
We observe that there exist features that strongly impact the prediction individually, such as the information about a non-existing \emph{checking account}.
However, the present \emph{credit amount} strongly impacts the prediction only in interaction with the given \emph{installment rate} (positively) and \emph{duration} (negatively).

The force plots in Figure~\ref{fig_force_plots} illustrate how the additive local explanations change, if higher order interactions are considered.
We consider the n-SII scores for $s_0=1,2,3$ for the \emph{California} housing dataset and an XGBoost regressor.
The force plot displays the positive and negative interaction scores starting from the predicted value to the left and right, respectively, sorted by their absolute value.
We observe that individual feature effects, such as \emph{Longitude}, reduce when higher order interactions are considered.
The interaction of \emph{Longitude} and \emph{Latitude} reveals the importance of the geographic location of this instance.

The waterfall chart in Figure~\ref{fig_waterfall} displays the explanations of n-SII with order $s_0=3$ for an instance in the \emph{bike} dataset.
For this instance, it can be seen that the interaction of the evening \emph{hour} with a non-\emph{working} day affects the prediction negatively, whereas the interaction with both \emph{temperature} features contribute positively.

\paragraph{n-SII Plots Quantify Interactions of Each Feature.}

To assess the strength of interaction per individual feature, we utilize the visualization of n-SII values presented by \citet{Bord.2023}.
We compute exact n-SII scores up to order $s_0=7$ for the \emph{German Credit} dataset.
The positive and negative interactions are distributed equally onto each feature in the subset and displayed on the positive and negative axes, respectively.
The sum of all stacked bars results in the SV of each feature \cite{Bord.2023}.
In Figure~\ref{fig_n_SII}, we observe that the interaction effects diminish at order 5, with interactions of orders 6 and 7 being virtually absent. 
Assuming that interactions decay with higher order, this visualization can be used to find the maximum order to explain the prediction (i.e. $s_0 = 5$ from Figure~\ref{fig_n_SII}).

\section{Limitations}
TreeSHAP-IQ applies to the broad class of CIIs, provided that its representation in terms of a weighted sum of discrete derivatives is known.
For FSI, this representation is only explicitly known for top-order interactions \cite{Tsai.2022}, as FSI is motivated as a solution to a constrained weighted least square problem.
Similar to the SV, Shapley interactions strongly rely on how absent features are modeled.
In our work, we considered the \emph{path dependent} feature perturbation \cite{Lundberg.2020}, which is linked to the \emph{observational} approach \cite{Chen.2020}.
The \emph{interventional} approach \cite{Lundberg.2020} can be computed with TreeSHAP-IQ, akin to TreeSHAP, but similarly increases the computational complexity by the number of samples used in the background dataset.
In this case, more efficient variants should be used instead \cite{DBLP:conf/aaai/ZernBK23}.
Both paradigms yield different explanations, where the appropriate choice should be carefully done depending on the application \cite{Chen.2020}.

\section{Conclusion and Future Work}
We presented TreeSHAP-IQ, an efficient method to compute any-order additive Shapley interactions that locally explain single predictions for general ensembles of trees.
Akin to SOTA Linear TreeSHAP \cite{Yu.2022}, our algorithm is based on a solid theoretical foundation that exploits polynomial arithmetic. 
We applied TreeSHAP-IQ on SOTA ML models, such as XGBoost \cite{Chen.2016}, and several benchmark datasets.
We demonstrated that TreeSHAP-IQ reveals intricate feature interactions, which enrich Shapley-based feature attribution.

Utilizing well-known visualization and aggregation techniques from machine learning \cite{Lundberg.2017,Bord.2023} and statistics \cite{Inglis.2022} we presented these scores in a manner that is easily understandable and interpretable.
While interactions are widely studied in statistics, explaining local predictions using interaction scores, in particular with Shapley-based interactions, is an emerging line of research in the field of XAI.
Due to the exponentially increasing number of interactions, we provided intuitive visualizations to present TreeSHAP-IQ scores to practitioners.
Nevertheless, it would be beneficial to explore further human-centered post-processing techniques and visualizations, as well as rigorously evaluate the explanatory capabilities of TreeSHAP-IQ with user studies, especially to validate quantitatively that the user's understanding increases when higher order explanations are presented.
Additionally, the n-SII scores define a local generalized additive model (GAM) \cite{Bord.2023} that could be further linked to functional decomposition \cite{Hiabu.2023}.

\section*{Acknowledgements}

We sincerely thank the anonymous reviewers for their work and helpful comments.
We gratefully acknowledge funding by the Deutsche Forschungsgemeinschaft (DFG, German Research Foundation): TRR 318/1 2021 – 438445824.

\bibliography{references}

\appendix
\onecolumn
\setcounter{secnumdepth}{2}

\section{Proofs}\label{sec_appx_proofs}

\subsection{Proof of Proposition \ref{prop_SII_R}}
For a leaf $v$ in $\mathcal T_f$, it holds $I^{\text{SII}}(R^v,S)=$
    \begin{align*}
       \left(\sum_{L \subseteq S} (-1)^{\vert S \vert - \vert L \vert}\prod_{j \in L} q_{j,v}\right)\psi\left(\left\lfloor\frac{G_v}{\prod_{j \in S}(q_{j,v}+y)}\right\rfloor\right).
    \end{align*}
\begin{proof}
    Let $R^v$ be a decision rule of leaf node $v \in \mathcal L(\mathcal T)$.
    For the S-derivative it follows by the recursive property and (\ref{eq_R_pre}) that
    \begin{equation*}
    \delta_{S}(R^v,T) = R^v_T \sum_{L \subseteq S} (-1)^{\vert S \vert - \vert L \vert}\prod_{j \in L} q_{j,v}.
\end{equation*}
Plugging this representation into the definition of SII with
\begin{equation*}
    \omega(t) := \frac{1}{(\fnum-\vert S \vert+1) \cdot\binom{\fnum-\vert S \vert}{\vert T \vert}}
\end{equation*}
yields
\begin{align*}
I^{\text{SII}}(R^v,S) := \sum_{T \subseteq \fset\setminus S} \omega(\vert T \vert) \delta_{S}(f,T)
= \left(\sum_{L \subseteq S} (-1)^{\vert S \vert - \vert L \vert}\prod_{j \in L} q_{j,v}\right) \sum_{T \subseteq \fset\setminus S} \omega(\vert T \vert)R^v_T.
\end{align*}
This sum is further simplified to
\begin{align}\label{appx_proof_lem_sum}
    \sum_{T \subseteq \fset\setminus S} \omega(\vert T \vert)R^v_T = R^v_\emptyset\sum_{T \subseteq \fset\setminus S}  \omega(\vert T \vert)\prod_{j \in T} q_{j,v} 
    = R^v_\emptyset\sum_{k=0}^{\fnum-\vert S \vert} \omega(k) \sum^{\vert T \vert = k}_{T \subseteq \fset\setminus S} \prod_{j \in T} q_{j,v}.
\end{align}
We thus need to show that this term is equal to $\psi$ applied on
\begin{align*}
    \frac{G^\text{SP}_{v}(y)}{\prod_{j \in S}(q_{j,v}+y)} &= R^v_\emptyset \frac{\prod_{j \in \mathcal{F}(R^v)} \left(q_{j,v}+y\right)}{\prod_{j \in S}(q_{j,v}+y)}
    = R^v_\emptyset \prod_{j \in \mathcal{F}(R^v) \setminus S} \left(q_{j,v}+y\right).
\end{align*}
With $d := \vert \mathcal{F}(R^v) \vert$ using the scale invariance of the SP, we can scale this polynomial to the degree of $\fnum$, i.e. multiplying it with $(1+y)$ for every feature not present in $\mathcal{F}(R^v)$, i.e. $\fset \setminus \mathcal{F}(R^v)$.
This yields with the above that
\begin{align*}
    &\frac{G^\text{SP}_{v}(y) \odot (1+y)^{\fnum-d}}{\prod_{j \in S}(q_{j,v}+y)}
    =  R^v_\emptyset \left(\prod_{j \in \mathcal{F}(R^v) \setminus S}q_{j,v}+y\right) \odot (1+y)^{\fnum-d}
    = R^v_\emptyset \prod_{j \in \fset \setminus S}\left(q_{j,v}+y\right),
\end{align*}
where the last line follows from the fact that $q_{j,v}=1$ if the feature is not present in the path.
As $\psi$ weights the coefficients of this polynomial, we proceed by evaluating the coefficients.
When writing the product as a sum, every combination $T \subseteq \fset\setminus S$ of features appears exactly once, and thus
\begin{align*}
    \prod_{j \in \fset \setminus S}\left(q_{j,v}+y\right) &= \sum_{T \subseteq \fset\setminus S} \left(\prod_{j \in T} q_{j,v}\right) y^{\fnum-\vert T \vert}
    = \sum_{k=0}^{\fnum-\vert S \vert} \left(\sum^{\vert T \vert = k}_{T \subseteq \fset\setminus S} \prod_{j \in T} q_{j,v}\right) y^{\fnum-\vert S \vert - k}
    = \sum_{k'=0}^{\fnum-\vert S \vert} \left(\sum^{\vert T \vert = \fnum-\vert S \vert - k'}_{T \subseteq \fset\setminus S} \prod_{j \in T} q_{j,v}\right) y^{k'}
\end{align*}
Hence, it follows for the polynomial of degree $\fnum-\vert S\vert$ that
\begin{align*}
    \psi\left(\frac{G^\text{SP}_{v}(y)}{\prod_{j \in S}(q_{j,v}+y)}\right) 
    &= \psi\left(\frac{G^\text{SP}_{v}(y) \odot (1+y)^{\fnum-d}}{\prod_{j \in S}(q_{j,v}+y)}\right)
    \\
    &= \frac{R^v_\emptyset}{\fnum-\vert S \vert + 1} 
    \left\langle \sum_{k=0}^{\fnum-\vert S \vert} \left(\sum^{\vert T \vert = \fnum-\vert S \vert - k}_{T \subseteq \fset\setminus S} \prod_{j \in T} q_{j,v}\right) y^{k}, B_{\fnum-\vert S\vert}\right\rangle
    \\
    &= \frac{R^v_\emptyset}{\fnum-\vert S \vert + 1} \sum_{k=0}^{\fnum-\vert S \vert} \left(\sum^{\vert T \vert = \fnum-\vert S \vert - k}_{T \subseteq \fset\setminus S} \prod_{j \in T} q_{j,v}\right) \frac{1}{\binom{\fnum-\vert S \vert}{k}}
    \\
    &=\frac{R^v_\emptyset}{\fnum-\vert S \vert + 1} \sum_{k=0}^{\fnum-\vert S \vert} \left(\sum^{\vert T \vert = k}_{T \subseteq \fset\setminus S} \prod_{j \in T} q_{j,v}\right) \frac{1}{\binom{\fnum-\vert S \vert}{\fnum-\vert S \vert - k}}
    \\
    &=\frac{R^v_\emptyset}{\fnum-\vert S \vert + 1} \sum_{k=0}^{\fnum-\vert S \vert} \left(\sum^{\vert T \vert = k}_{T \subseteq \fset\setminus S} \prod_{j \in T} q_{j,v}\right) \frac{1}{\binom{\fnum-\vert S \vert}{k}}
    \\
    &= R^v_\emptyset\sum_{k=0}^{\fnum-\vert S \vert} \omega(k) \sum^{\vert T \vert = k}_{T \subseteq \fset\setminus S} \prod_{j \in T} q_{j,v},
\end{align*}
which is equal to (\ref{appx_proof_lem_sum}) and finishes the proof.
\end{proof}

\subsection{Proof of Proposition \ref{prop_kappa}}
    For the sum of coefficients of the IP, it holds 
    \begin{align}
        \kappa(H^\text{IP}_{S,e}) = \sum_{L \subseteq S} (-1)^{\vert S \vert - \vert L \vert}\prod_{j \in L} p_{j,e}.
    \end{align}
    If there exists $j \in S$ with $p_{j,e}=1$, then $\kappa(H^\text{IP}_{S,e})=0$.
\begin{proof}
    We first compute the coefficients of the IP as
    \begin{align*}
        H^\text{IP}_{S,e}(y) &= \prod_{j \in S} (p_{j,e} - y)
        = \sum_{k=0}^{\vert S\vert} (-1)^{\vert S\vert -k}  \left(\sum_{L\subseteq S}^{\vert L \vert = k} \prod_{j \in L} p_{j,e} \right) y^{\vert S \vert - k}
        =\sum_{k=0}^{\vert S\vert} (-1)^{k}  \left(\sum_{L\subseteq S}^{\vert L \vert = \vert S \vert - k}\prod_{j \in L} p_{j,e} \right) y^{k}.
    \end{align*}
Hence, 
\begin{align*}
   \kappa(H^\text{IP}_{S,e}) &= \sum_{k=0}^{\vert S\vert} (-1)^{k} \sum_{L\subseteq S}^{\vert L \vert = \vert S \vert - k} \prod_{j \in L} p_{j,e}
   =(-1)^{\vert S \vert} \sum_{k=0}^{\vert S\vert} (-1)^{k} \sum_{L\subseteq S}^{\vert L \vert = k} \prod_{j \in L} p_{j,e}
    = \sum_{L\subseteq S} (-1)^{\vert S \vert - \vert L \vert} \prod_{j \in L} p_{j,e},
\end{align*}
which finishes the first part of the proof.

Now, let $p_{j_0,e}=1$ for some $j_0 \in S$. Then,
\begin{align*}
    \kappa(H^\text{IP}_{S,e}) &=  \sum_{L\subseteq S} (-1)^{\vert S \vert - \vert L \vert} \prod_{j \in L} p_{j,e}
    \\
    &=\sum_{L\subseteq S \setminus \{j_0\}} (-1)^{\vert S \vert - \vert L \vert} \prod_{j \in L} p_{j,e}
    + \sum_{L\subseteq S \setminus \{j_0\}} (-1)^{\vert S \vert - \vert L  \cup \{j_0\} \vert} \prod_{j \in L  \cup \{j_0\}} p_{j,e}
    \\
    &=\sum_{L\subseteq S \setminus \{j_0\}} (-1)^{\vert S \vert - \vert L \vert} \prod_{j \in L} p_{j,e}
    \\
    &- \sum_{L\subseteq S \setminus \{j_0\}} (-1)^{\vert S \vert - \vert L \vert} \prod_{j \in L} p_{j,e}
    \\
    &= 0,
\end{align*}
which finishes the proof.
\end{proof}

\subsection{Proof of Proposition \ref{prop_SII_R_poly}}
     For a decision rule $R^v$ of a leaf node $v$ and a subset $S \subset \fset$, let $P_{S,v} := \bigcup_{i \in S} P_{i,v}$ and  $e^\uparrow_S$ as the closest ancestor of $e$ in $P_{S,v}$. The SII of $R^v$ is then
\begin{align*}
    I^{\text{SII}}(R^v,S) = \sum_{e \in P_{S,v}} \kappa(H^\text{IP}_{S,e}) \psi\left(\left\lfloor\frac{G_v \odot (y+1)^{d_e-d_v}}{ H^\text{QP}_{S,e}}\right\rfloor\right) -   \kappa(H^\text{IP}_{S,e_S^\uparrow}) \psi\left(\left\lfloor\frac{G_v \odot (y+1)^{d_{e^\uparrow_S}-d_v}}{H^\text{QP}_{S,e_S^\uparrow}}\right\rfloor\right).
\end{align*}  
\begin{proof}
By Proposition \ref{prop_SII_R}, we need to show that the right hand side is equal to
\begin{align}\label{appx_proof_prop_rule}
       \left(\sum_{L \subseteq S} (-1)^{\vert S \vert - \vert L \vert}\prod_{j \in L} q_{j,v}\right)\psi\left(\left\lfloor\frac{G_v}{\prod_{j \in S}(q_{j,v}+y)}\right\rfloor\right).
    \end{align}
We can simplify the sum on the right hand side, as all terms except the last edge $e_S^* \in P_{S,v}$ cancel out, to
\begin{align*}
    &\sum_{e \in P_{S,v}} \kappa(H^\text{IP}_{S,e}) \psi\left(\left\lfloor\frac{G_v \odot (y+1)^{d_e-d_v}}{ H^\text{QP}_{S,e}}\right\rfloor\right) -   \kappa(H^\text{IP}_{S,e_S^\uparrow}) \psi\left(\left\lfloor\frac{G_v \odot (y+1)^{d_{e^\uparrow_S}-d_v}}{H^\text{QP}_{S,e_S^\uparrow}}\right\rfloor\right)
    \\
    &=  \kappa(H^\text{IP}_{S,e^*_S}) \psi\left(\left\lfloor\frac{G_v \odot (y+1)^{d_{e^*_S}-d_v}}{ H^\text{QP}_{S,e^*_S}}\right\rfloor\right)
    \\
    &= \kappa(H^\text{IP}_{S,e^*_S}) \psi\left(\left\lfloor\frac{G_v \odot (y+1)^{d_{e^*_S}-d_v}}{ \prod_{j \in S} (p_{j,e^*_S}+y)}\right\rfloor\right)
    \\
    &= \kappa(H^\text{IP}_{S,e^*_S}) \psi\left(\left\lfloor\frac{G_v }{\prod_{j \in S}(p_{j,e^*_S}+y)}\right\rfloor \odot (y+1)^{d_{e^*_S}-d_v}\right)
    \\
    &= \kappa(H^\text{IP}_{S,e^*_S}) \psi\left(\left\lfloor\frac{G_v }{\prod_{j \in S} (p_{j,e^*_S}+y)}\right\rfloor\right),
\end{align*}
where we used the scale invariance of $\psi$ and the definition of the QP.
For the last edge $e^*_S$ in the path $P_{S,v}$ it holds that $p_{i,e^*_S}=q_{i,v}$ for all $i \in S$ and thus the argument in $\psi$ is equal to the argument in (\ref{appx_proof_prop_rule}).
Furthermore, for the IP, we have by Proposition \ref{prop_kappa} that
\begin{align*}
    \kappa(H^\text{IP}_{S,e^*_S}) &= \sum_{L \subseteq S} (-1)^{\vert S \vert - \vert L \vert}\prod_{j \in L} p_{j,e^*_S}
    =\sum_{L \subseteq S} (-1)^{\vert S \vert - \vert L \vert}\prod_{j \in L} q_{j,v},
\end{align*}
which concludes the proof.
\end{proof}

\subsection{Proof of Theorem \ref{thm_main}}
    For $S \subset \fset$ let $E_S := \bigcup_{i\in S}E_i$ be the set of edges that split on any feature in $S$ and denote $e_S^\uparrow$ as the closest ancestor of $e$ in $P_{S,v}$.
    The SII is then computed as
\begin{align*}
    I^{\text{SII}}(f,S) = \sum_{e \in E_S} \kappa(H^\text{IP}_{S,e}) \psi\left(\left\lfloor\frac{G_{h(e)}}{ H^\text{QP}_{S,e}}\right\rfloor\right) -   \kappa(H^\text{IP}_{S,e_S^\uparrow}) \psi\left(\left\lfloor\frac{G_{h(e)} \odot (y+1)^{d_{e^\uparrow_S}-d_e}}{H^\text{QP}_{S,e_S^\uparrow}}\right\rfloor\right).
\end{align*}

\begin{proof}
    The model $f$ can be represented as a sum of decision rules as $f(T) = \sum_{v \in \mathcal L(\mathcal T_{f})} R^v_T$ and thus by the linearity of SII and Proposition \ref{prop_SII_R_poly}
    \begin{align*}
         I^{\text{SII}}(f,S) &= \sum_{v \in \mathcal L(\mathcal T_{f})} I^{\text{SII}}(R^v,S) 
         \\
         &= \sum_{v \in \mathcal L(\mathcal T_{f})} \sum_{e \in P_{S,v}} \kappa(H^\text{IP}_{S,e}) \psi\left(\left\lfloor\frac{G_v \odot (y+1)^{d_e-d_v}}{ H^\text{QP}_{S,e}}\right\rfloor\right) - \kappa(H^\text{IP}_{S,e_S^\uparrow}) \psi\left(\left\lfloor\frac{G_v \odot (y+1)^{d_{e^\uparrow_S}-d_v}}{H^\text{QP}_{S,e_S^\uparrow}}\right\rfloor\right)
         \\
       &= \sum_{v \in \mathcal L(\mathcal T_{f})} \sum_{e \in P_{S,v}} \kappa(H^\text{IP}_{S,e}) \psi\left(\left\lfloor\frac{G_v \odot (y+1)^{d_e-d_v}}{ H^\text{QP}_{S,e}}\right\rfloor\right) - \kappa(H^\text{IP}_{S,e_S^\uparrow}) \psi\left(\left\lfloor\frac{G_v \odot (y+1)^{(d_e-d_v)+(d_{e^\uparrow_S}-d_e)}}{H^\text{QP}_{S,e_S^\uparrow}}\right\rfloor\right),       
    \end{align*}
    where we added $0=d_e-d_e$ to the scaling factor.
    Now we have that $v \in \mathcal L(\mathcal T_f) \wedge e \in E_S \Leftrightarrow v \in \mathcal L(h(e))$, i.e. if $v$ is a leaf and $e$ and edge in its path, then $v$ is also reachable from the head of $e$, $h(e)$, and vice versa.
    We can thus change the summation to
    \begin{align*}
        I^{\text{SII}}(f,S) &= \sum_{e \in E_S} \sum_{v \in \mathcal L(h(e))} \kappa(H^\text{IP}_{S,e}) \psi\left(\left\lfloor\frac{G_v \odot (y+1)^{d_e-d_v}}{ H^\text{QP}_{S,e}}\right\rfloor\right) - \kappa(H^\text{IP}_{S,e_S^\uparrow}) \psi\left(\left\lfloor\frac{G_v \odot (y+1)^{(d_e-d_v)+(d_{e^\uparrow_S}-d_e)}}{H^\text{QP}_{S,e_S^\uparrow}}\right\rfloor\right).
    \end{align*}
    Note that $G_v \odot (y+1)^{d_e-d_v}$ yields a polynomial of degree $d_e$ and the degree of $H^\text{QP}_{S,e}$ is always equal to $\vert S \vert$.
    Hence, the polynomials can be summed using the additivity of $\psi$, which yields
    \begin{align*}
        \sum_{v \in \mathcal L(h(e))} \kappa(H^\text{IP}_{S,e}) \psi\left(\left\lfloor\frac{G_v \odot (y+1)^{d_e-d_v}}{ H^\text{QP}_{S,e}}\right\rfloor\right) 
        &=    \kappa(H^\text{IP}_{S,e}) \psi\left( \sum_{v \in \mathcal L(h(e))} \left\lfloor\frac{G_v \odot (y+1)^{d_e-d_v}}{ H^\text{QP}_{S,e}}\right\rfloor\right) 
        \\
        &=\kappa(H^\text{IP}_{S,e}) \psi\left(\left\lfloor\frac{\sum_{v \in \mathcal L(h(e))} G_v \odot (y+1)^{d_e-d_v}}{ H^\text{QP}_{S,e}}\right\rfloor\right) 
        \\
        &= \kappa(H^\text{IP}_{S,e}) \psi\left(\left\lfloor\frac{\bigoplus_{v \in \mathcal L(h(e))} G_v }{ H^\text{QP}_{S,e}}\right\rfloor\right) 
        \\
        &= \kappa(H^\text{IP}_{S,e}) \psi\left(\left\lfloor\frac{G_{h(e)} }{ H^\text{QP}_{S,e}}\right\rfloor\right).
    \end{align*}
    Similarly, for the other term
    \begin{align*}
         &\sum_{v \in \mathcal L(h(e))} \kappa(H^\text{IP}_{S,e_S^\uparrow}) \psi\left(\left\lfloor\frac{G_v \odot (y+1)^{(d_e-d_v)+(d_{e^\uparrow_S}-d_e)}}{H^\text{QP}_{S,e_S^\uparrow}}\right\rfloor\right)
         \\
         &= \sum_{v \in \mathcal L(h(e))} \kappa(H^\text{IP}_{S,e_S^\uparrow}) \psi\left(\left\lfloor\frac{G_v \odot (y+1)^{(d_e-d_v)+(d_{e^\uparrow_S}-d_e)}}{H^\text{QP}_{S,e_S^\uparrow}}\right\rfloor\right)
         \\
         &= \kappa(H^\text{IP}_{S,e_S^\uparrow}) \psi\left(\sum_{v \in \mathcal L(h(e))}\left\lfloor\frac{G_v \odot (y+1)^{(d_e-d_v)+(d_{e^\uparrow_S}-d_e)}}{H^\text{QP}_{S,e_S^\uparrow}}\right\rfloor\right)
         \\
         &= \kappa(H^\text{IP}_{S,e_S^\uparrow}) \psi\left(\left\lfloor\frac{\sum_{v \in \mathcal L(h(e))} G_v \odot (y+1)^{d_e-d_v}}{H^\text{QP}_{S,e_S^\uparrow}}\odot (y+1)^{d_{e^\uparrow_S}-d_e}\right\rfloor\right)
         \\
         &= \kappa(H^\text{IP}_{S,e_S^\uparrow}) \psi\left(\left\lfloor\frac{\bigoplus_{v \in \mathcal L(h(e))} G_v}{H^\text{QP}_{S,e_S^\uparrow}}\odot (y+1)^{d_{e^\uparrow_S}-d_e}\right\rfloor\right)
         \\
         &= \kappa(H^\text{IP}_{S,e_S^\uparrow}) \psi\left(\left\lfloor\frac{G_{h(e)}\odot (y+1)^{d_{e^\uparrow_S}-d_e}}{H^\text{QP}_{S,e_S^\uparrow}}\right\rfloor\right),
    \end{align*}
    which finishes the proof.
\end{proof}

\clearpage
\section{Implementations}\label{sec_appx_implementations}
In the following, we describe pseudo-code for TreeSHAP-IQ for SII (Section \ref{appx_sec_treeshapiq}) and general CIIs (Section \ref{appx_sec_treeshapiq_cii}) and Linear TreeSHAP (Section \ref{appx_sec_linear_treeshap}), as well as its efficient implementation using the multipoint interpolation form for all polynomials.

\subsection{TreeSHAP-IQ Algorithm}\label{appx_sec_treeshapiq}
The pseudo-code for TreeSHAP-IQ is outlined in Algorithm \ref{appx_alg_treeshapiq}.

\begin{algorithm}
    \caption{\textsc{\texttt{TraverseTree}}($v,C,C^\text{IP}, C^\text{QP},a_x$) -- TreeSHAP-IQ for SII}
    \begin{algorithmic}[1]\label{appx_alg_treeshapiq}
    \REQUIRE Interaction order $s$, tree $\mathcal T = (V,E)$, leaf predictions $\mathcal V_v$ and empty decision rules $R_\emptyset^v$ for all leaf nodes $v \in \mathcal L(\mathcal T)$. 
    \\
    For all edges $e \in E$ with label $i$: Weights $w_e$ and inter-path products $p^{\text{raw}}(e) := 1/\prod_{e' \in P_{i,h(e)}} w_{e'}$
    \\
    For all edges $e \in E$ with label $i$ and interaction subsets $S$ with $i \in S$: Ancestors $e^\uparrow_S$.
    \IF{$v$ is not root}{
        \STATE $e, i \gets$ edge with $v$ as head and split feature $i$ of its source (parent of $v$)
        \STATE $p_{i,e}(x) \gets a_x(v) p^{\text{raw}}(e)$
        \STATE $C \gets C \odot (y+p_{i,e}(x))$
        \STATE $H^\text{IP}_{S,e} \gets C^\text{IP}[S] \odot (-y + p_{i,e}(x))$, if $i \in S$, else $C^\text{IP}[S]$ 
        \STATE $H^\text{QP}_{S,e} \gets C^\text{QP}[S] \odot (y + p_{i,e}(x))$, if $i \in S$, else $C^\text{QP}[S]$
        \IF{$e$ has ancestor $e^\uparrow$}{
            \STATE $p_{i,e^\uparrow}(x) \gets a_x(h(e^\uparrow)) p^{\text{raw}}(e^\uparrow)$
            \STATE $C \gets \left\lfloor C/(y+p_{i,e^\uparrow}(x))\right\rfloor$
            \STATE $H^\text{IP}_{S,e} \gets \left\lfloor H^\text{IP}_{S,e} / (-y + p_{i,e^\uparrow}(x))\right\rfloor$, $\forall S: i\in S$
            \STATE $H^\text{QP}_{S,e} \gets \left\lfloor H^\text{QP}_{S,e}/(y + p_{i,e^\uparrow}(x))\right\rfloor$,  $\forall S: i\in S$
        }
        \ENDIF
    }\ENDIF
    \IF{$v$ is leaf}{
        \STATE $G_v \gets C \cdot R^v_\emptyset$
    }\ELSE{
        \STATE $v_c \gets $ child nodes $v_c$ of $v$
        \STATE $\forall v_c: a_x(v_c) \gets \mathbf{1}(x \in \pi_{v_c})$
        \STATE $\forall v_c: G_{v_c} \gets $\textsc{\texttt{TraverseTree}}($v_c, C, H^\text{IP}_{\cdot,e}, H^\text{QP}_{\cdot,e}, a_x$)
        \STATE $G_v \gets \bigoplus_{v_c} G_{v_c}$
        \FOR{all $S$ with $i\in S$}
        \STATE $I[S] \gets I[S] + \kappa(H^\text{IP}_{S,e}) \psi(\left\lfloor G_v / H^\text{QP}_{S,e}\right\rfloor$
        \IF{$e_S^\uparrow \neq \perp$}
        \STATE $I[S] \gets I[S] - \kappa(H^\text{IP}_{S,e_S^\uparrow})\psi(\left\lfloor \left(G_v \odot (1+y)^{d_{e^\uparrow_S}-d_e}\right)/H^\text{QP}_{S,e^\uparrow_S}\right\rfloor)$
        \ENDIF
        \ENDFOR
    }
    \ENDIF
    \STATE \textbf{return} $G_v$
    \end{algorithmic}
\end{algorithm}

\subsection{Implementation of TreeSHAP-IQ for general CIIs}\label{appx_sec_treeshapiq_cii}
The pseudo-code of TreeSHAP-IQ applied to an arbitrary CII is outlined in Algorithm \ref{appx_alg_treeshapiq_cii}.

\begin{algorithm}
    \caption{\textsc{\texttt{TraverseTree}}($v,C,C^\text{IP}, C^\text{QP},a_x$) -- TreeSHAP-IQ for CIIs}
    
    \begin{algorithmic}[1]\label{appx_alg_treeshapiq_cii}
    \REQUIRE Interaction order $s$, tree $\mathcal T = (V,E)$, leaf predictions $\mathcal V_v$ and empty decision rules $R_\emptyset^v$ for all leaf nodes $v \in \mathcal L(\mathcal T)$. 
    \\
    For all edges $e \in E$ with label $i$: Weights $w_e$ and inter-path products $p^{\text{raw}}(e) := 1/\prod_{e' \in P_{i,h(e)}} w_{e'}$
    \\
    For all edges $e \in E$ with label $i$ and interaction subsets $S$ with $i \in S$: Ancestors $e^\uparrow_S$.
    \IF{$v$ is not root}{
        \STATE $e, i \gets$ edge with $v$ as head and split feature $i$ of its source (parent of $v$)
        \STATE $p_{i,e}(x) \gets a_x(v) p^{\text{raw}}(e)$
        \STATE $C \gets C \odot (y+p_{i,e}(x))$
        \STATE $H^\text{IP}_{S,e} \gets C^\text{IP}[S] \odot (-y + p_{i,e}(x))$, if $i \in S$, else $C^\text{IP}[S]$ 
        \STATE $H^\text{QP}_{S,e} \gets C^\text{QP}[S] \odot (y + p_{i,e}(x))$, if $i \in S$, else $C^\text{QP}[S]$
        \IF{$e$ has ancestor $e^\uparrow$}{
            \STATE $p_{i,e^\uparrow}(x) \gets a_x(h(e^\uparrow)) p^{\text{raw}}(e^\uparrow)$
            \STATE $C \gets \left\lfloor C/(y+p_{i,e^\uparrow}(x))\right\rfloor$
            \STATE $H^\text{IP}_{S,e} \gets \left\lfloor H^\text{IP}_{S,e} / (-y + p_{i,e^\uparrow}(x))\right\rfloor$, $\forall S: i\in S$
            \STATE $H^\text{QP}_{S,e} \gets \left\lfloor H^\text{QP}_{S,e}/(y + p_{i,e^\uparrow}(x))\right\rfloor$,  $\forall S: i\in S$
        }
        \ENDIF
    }\ENDIF
    \IF{$v$ is leaf}{
        \STATE $G_v \gets C \cdot R^v_\emptyset$
    }\ELSE{
        \STATE $v_c \gets $ child nodes $v_c$ of $v$
        \STATE $\forall v_c: a_x(v_c) \gets \mathbf{1}(x \in \pi_{v_c})$
        \STATE $\forall v_c: G_{v_c} \gets $\textsc{\texttt{TraverseTree}}($v_c, C, H^\text{IP}_{\cdot,e}, H^\text{QP}_{\cdot,e}, a_x$)
        \STATE $G_v \gets \bigoplus_{v_c} G_{v_c}$
        \FOR{all $S$ with $i\in S$}
        \STATE $I[S] \gets I[S] + \kappa(H^\text{IP}_{S,e}) \psi(\left\lfloor \left(G_v \odot (1+y)^{\fnum-d_e} \right) / H^\text{QP}_{S,e}\right\rfloor$
        \IF{$e_S^\uparrow \neq \perp$}
        \STATE $I[S] \gets I[S] - \kappa(H^\text{IP}_{S,e_S^\uparrow})\psi(\left\lfloor \left(G_v \odot (1+y)^{\fnum-d_e} \right)/H^\text{QP}_{S,e^\uparrow_S}\right\rfloor)$
        \ENDIF
        \ENDFOR
    }
    \ENDIF
    \STATE \textbf{return} $G_v$
    \end{algorithmic}
\end{algorithm}

\subsection{Implementation of Linear TreeSHAP}\label{appx_sec_linear_treeshap}
The pseudo-code of Linear TreeSHAP is outlined in Algorithm \ref{appx_alg_linear_treeshap}.

\begin{algorithm}
    \caption{\textsc{\texttt{TraverseTree}}($v,C,a_x$) -- Linear TreeSHAP}
    \begin{algorithmic}[1]\label{appx_alg_linear_treeshap}
  \REQUIRE Interaction order $s$, tree $\mathcal T = (V,E)$, leaf predictions $\mathcal V_v$ and empty decision rules $R_\emptyset^v$ for all leaf nodes $v \in \mathcal L(\mathcal T)$. 
    \\
    For all edges $e \in E$ with label $i$: Weights $w_e$, inter-path products $p^{\text{raw}}(e) := 1/\prod_{e' \in P_{i,h(e)}} w_{e'}$ and ancestors $e^\uparrow$.
    \IF{$v$ is not root}{
        \STATE $e, i \gets$ edge with $v$ as head and split feature $i$ of its source (parent of $v$)
        \STATE $p_{e}(x) \gets a_x(v) p^{\text{raw}}(e)$
        \STATE $C \gets C \odot (y+p_{e}(x))$
        \IF{$e$ has ancestor $e^\uparrow$}{
            \STATE $p_{e^\uparrow}(x) \gets a_x(h(e^\uparrow)) p^{\text{raw}}(e^\uparrow)$
            \STATE $C \gets \left\lfloor C/(y+p_{i,e^\uparrow}(x))\right\rfloor$
        }
        \ENDIF
    }\ENDIF
    \IF{$v$ is leaf}{
        \STATE $G_v \gets C \cdot R^v_\emptyset$
    }\ELSE{
        \STATE $v_c \gets $ child nodes $v_c$ of $v$
        \STATE $\forall v_c: a_x(v_c) \gets \mathbf{1}(x \in \pi_{v_c})$
        \STATE $\forall v_c: G_{v_c} \gets $\textsc{\texttt{TraverseTree}}($v_c, C, a_x$)
        \STATE $G_v \gets \bigoplus_{v_c} G_{v_c}$
        \STATE $\phi[i] \gets \phi[i] + (p_e-1)\psi(\left\lfloor G_v / (p_e + y )\right\rfloor$
        \IF{$e_S^\uparrow \neq \perp$}
        \STATE $\phi[i] \gets \phi[i] - (p_e^\uparrow-1)\psi(\left\lfloor G_v/(p_{e^\uparrow}+y)\right\rfloor)$
        \ENDIF
    }
    \ENDIF
    \STATE \textbf{return} $G_v$
    \end{algorithmic}
\end{algorithm}

\subsection{Efficient Implementation of Polynomial Operations using Multipoint Interpolation}\label{sec_appx_interpolation}

Linear TreeSHAP and TreeSHAP-IQ rely on multiplication and division for polynomials, as well as evaluating the polynomial coefficients using $\psi$ and $\kappa$.
These operations can be efficiently implemented by storing the polynomial in a multipoint interpolation form.
Instead of storing the poylnomial, we store its evaluation at base points $\mathcal Y$.
Multiplication and division then transfer to vector multiplication and division.
The inner product of the coefficients of the polynomial, as required for $\psi$ and $\kappa$, can be efficiently computed using the following lemma.

\begin{lemma}[\citeauthor{Yu.2022}, \citeyear{Yu.2022}]
    Let $p,q \in \mathbb{R}[x]_d$ with coefficients $A,B$, respectively. Then $\langle p,q \rangle = \langle A , B \rangle = \langle p(Y), \mathcal V(Y)^{-1} B \rangle$, where $\mathcal V \in \mathbb{R}^{d+1 \times d+1}$ corresponds to the Vandermonde matrix with entries $(\mathcal V(Y))_{i,j} = y_i^j$. 
\end{lemma}

As proposed by \citet{Yu.2022}, we use the Chebyshev points $\mathcal Y$ to evaluate the polynomial, as they are optimal in terms of numerical stability.
Using the Chebyshev points $\mathcal Y$, the values $\mathcal V(\mathcal Y)^{-1} C$ have to be precomputed, where $C$ refers to the coefficients of $B$.
It is thus required to precompute as many values as the degree of the given polynomial.

\clearpage
\section{Experiments}
This section contains further information on the experiments and additional results like a run-time analysis in Section~\ref{app_run_time}

\subsection{Dataset and Model Descriptions}

This section contains detailed information about the datasets, the required pre-processing steps and the model fitting.

\subsubsection{Models}

The following tree-based models are used in our experiments.

\begin{itemize} 
\item{\textbf{XGBoost (XBG) \cite{Chen.2016}}}\\
XGBoost is an ensemble learning algorithm based on gradient boosting. It utilizes decision trees as base learners and optimizes a user-defined loss function through an iterative process. XGBoost incorporates regularization techniques to control model complexity and improve generalization.
In our experiments, we relied on the \texttt{XGBoost} library\footnote{\url{https://xgboost.readthedocs.io/en/stable/}}.

\item{\textbf{Gradient-Boosted Tree (GBT)}}\\
Gradient-Boosted Trees are an ensemble learning method that combines multiple weak learners, here decision trees, in a sequential manner. It trains each tree to correct the errors made by the previous ones, effectively reducing the overall prediction error.
We used the default parametrization of the \texttt{GradientBoostingClassifier} and \texttt{GradientBoostingRegressor} classes from the \texttt{scikit-learn} library \cite{Pedregosa.2011}

\item{\textbf{Random Forest (RF)}}\\
Random Forest is an ensemble learning algorithm that constructs a collection of decision trees by using bootstrapped subsets of the training data and random feature selection. The predictions from individual trees are then aggregated to make a final prediction. This technique reduces overfitting and enhances model robustness compared with single decision trees.
We used the default parametrization of the \texttt{RandomForestClassifier} and \texttt{RandomForestRegressor} classes for classification and regression tasks, respectively \cite{Pedregosa.2011}.

\item{\textbf{Decision Tree (DT)}}\\
A Decision Trees is a simple yet powerful model that makes predictions by recursively splitting the data based on the most informative features. Each internal node represents a decision based on a feature, and each leaf node represents a prediction. Decision Trees are prone to overfitting, but they serve as the fundamental building blocks for ensemble methods like Random Forest and Gradient-Boosted Trees. We used the default parameterization of the \texttt{DecisionTreeClassifier} and \texttt{DecisionTreeRegressor} classes in the \texttt{scikit-learn} library \cite{Pedregosa.2011}.
\end{itemize}

\subsubsection{Datasets}
\label{appendix_datasets}

The following dataset are used in our experiments.
The data was either directly retrieved from the cited sources or via \texttt{scikit-learn} \cite{Pedregosa.2011} or \texttt{openml} \cite{Feurer.2020}.

\begin{itemize} 
\item{\textbf{German Credit \cite{Hofmann.1994}}}\\
The German Credit dataset consists of credit applicants' information from a German bank, including 20 attributes such as age, employment status, credit history, and risk assessment outcomes. It contains 1,000 instances, and the primary prediction task involves classification to determine whether an applicant is a ``good'' or ``bad'' credit risk based on their attributes. The dataset was retrieved from the UCI repository\footnote{\url{http://archive.ics.uci.edu/dataset/144/statlog+german+credit+data}}.

\item{\textbf{Bank \cite{Moro.2011}}}\\
The Bank dataset originates from a Portuguese banking institution and includes customer-related attributes, marketing campaign details, and the outcome of customers subscribing to term deposits. It contains 45,211 instances, and the primary prediction task is classification, aiming to predict whether a customer will subscribe to a term deposit or not. We retrieved the dataset via \texttt{openml} and the identifier \emph{1461}.

\item{\textbf{Adult Census \cite{Kohavi.1996}}}\\
Also known as the ``Census Income'' or ``Adult Income'' dataset, it contains socio-demographic attributes of individuals along with their income levels. It contains 45,222 instances and 14 attributes, and the main prediction task is classification, aiming to predict whether an individual's income exceeds $50,000$ per year. We retrieved the dataset via \texttt{openml} and the identifier \emph{1590}.

\item{\textbf{Bike \cite{FanaeeT.2014}}}\\
This dataset involves information from a bike-sharing program, encompassing attributes like weather conditions, time, and bike rental counts. It contains 17,379 instances and 12 attributes, and the primary prediction task is regression, aiming to predict the count of bike rentals (a continuous value) based on the provided attributes. We retrieved the dataset via \texttt{openml} and the identifier \emph{42712}.

\item{\textbf{COMPAS \cite{Angwin.2016}}}\\
The COMPAS dataset comprises criminal defendant attributes and recidivism predictions generated by a proprietary software tool. It contains $6,172$ instances and $11$ attributes, and the main prediction task is classification, aiming to predict whether a criminal defendant is likely to recidivate or not.
We retrieved the ``simplified'' dataset from Kaggle\footnote{\url{https://www.kaggle.com/datasets/danofer/compass}}, which was presented in a blog post\footnote{\url{https://blog.fastforwardlabs.com/2017/03/09/fairml-auditing-black-box-predictive-models.html}}.

\item{\textbf{Titanic \cite{Dawson.1995}}}\\
The Titanic dataset records passenger information from the ill-fated Titanic voyage, including features like age, gender, class, and survival status. It contains 891 instances and 9 attributes, and the core prediction task is classification, aiming to predict whether a passenger survived the Titanic disaster or not. The data was retrieved from Kaggle\footnote{\url{https://www.kaggle.com/c/titanic/data}}.

\item{\textbf{California \cite{Kelley.1997}}}\\
The California housing dataset encompasses housing-related attributes for various geographical regions in California. It contains 20,640 instances and 8 attributes, and the primary prediction task is regression, aiming to predict the median value of owner-occupied homes in California based on the provided attributes. The dataset was retrieved from \texttt{scikit-learn}.
\end{itemize}

\subsubsection{Pre-processing and Model Training}

The following list contains all pre-processing steps for each dataset to reproduce the experiments. For further details we refer to the technical supplement containing all experiment scripts and these steps. We base most of the data transformation on \texttt{scikit-learn} \cite{Pedregosa.2011}.
All models were trained with a 70\%, 30\% training split and fixed random seeds.

\begin{itemize}
    \item \textbf{German Credit:} We transform the categorical columns (``checkingstatus'', ``history'', ``purpose'', ``savings'', ``employ'', ``status'', ``others'', ``property'', ``otherplans'', ``housing'', ``job'', ``tele'', and ``foreign'') into integer values via an ordinal encoding. We binarize the label into $0$ and $1$.
    \item \textbf{Bank:} We transform the categorical columns (``job'', ``marital'', ``education'', ``default'', ``housing'', ``loan'', ``contact'', ``day'', ``month'', ``campaign'', and ``poutcome'') into integer values via an ordinal encoding. We binarize the label into $0$ and $1$. Lastly, we drop rows containing null values.
    \item \textbf{Adult Census:} We drop rows containing null values and transform the categorical columns (``workclass'', ``education'', ``marital-status'', ``occupation'', ``relationship'', ``race'', ``sex'', ``native-country'', and ``education-num'') into integer values.
    \item \textbf{Bike:} We transform the categorical columns (``season'', ``year'', ``month'', ``holiday'', ``weekday'', ``workingday'', and ``weather'') into integer values via an ordinal encoding and drop rows containing null values.
    \item \textbf{COMPAS:} The dataset was used as-is and no data transformation was applied.
    \item \textbf{Titanic:} We transform the categorical columns (``Sex'', ``Ticket'', ``Cabin'', and ``Embarked'') into integer values via an ordinal encoding. We impute missing values in categorical and numerical features with the mode and median values respectively. 
    \item \textbf{California:} The dataset was used as-is and no data transformation was applied.
\end{itemize}

\begin{table}[t]
    \centering
    \begin{tabular}{@{}lccccccc@{}}
    \toprule
    \textbf{Datasets} & \textbf{\# Instances} & \textbf{\# Features} & \textbf{Target} & \multicolumn{4}{c}{\textbf{Performance} ($R^2$ or Accuracy)} \\
    & & & & \textbf{XGB} & \textbf{GBT} & \textbf{RF} & \textbf{DT} \\ \midrule
    \emph{German Credit} & $1\,000$ & $20$ & $\{0,1\}$ & 0.7542 & 0.7700 & 0.7533 & 0.6833 \\
    \emph{Bank} & $45\,211$ & $16$ & $\{0,1\}$ & 0.9064 & 0.9031 & 0.9020 & 0.8941 \\
    \emph{Adult Census} & $45\,222$ & $14$ & $\{0,1\}$ & 0.8715 & 0.8655 & 0.8568 & 0.8524 \\
    \emph{Bike} & $17\,379$ & $12$ & $\mathbb{R}$ & 0.9464 & 0.8442 & 0.8840 & 0.8796 \\
    \emph{COMPAS} & $6\,172$ & $11$ & $\{0,1\}$ & 0.6695 & 0.6776 & 0.6646 & 0.6609 \\
    \emph{Titanic} & $891$ & $9$ & $\{0,1\}$ & 0.7761 & 0.8059 & 0.7947 & 0.7723\\
    \emph{California} & $20\,640$ & $8$ & $\mathbb{R}$ & 0.8315 & 0.8317 & 0.7723 & 0.5945 \\    
    \bottomrule
    \end{tabular}
    \caption{Overview of datasets used in the experiments with performance}\label{appx_tab_datasets}
\end{table}

\clearpage
\subsection{Run-time Analysis}
\label{app_run_time}

In this section, we provide a run-time analysis that validates our theoretical finding about the runtime complexity of our algorithm.
As described in Section \ref{sec_treeshapiq_theory}, the run-time complexity of TreeSHAP-IQ for all interactions of order $s$ is $\mathcal{O}\left(m \cdot \ell_{\mathcal T} \cdot d_{\max} \cdot \binom{\fnum-1}{s-1}\right)$, where $m$ corresponds to the number of explanation points, $d_{\max}$ corresponds to the maximum depth of the tree and $\ell_{\mathcal T}$ is the number of leaves in the tree.
Clearly, there are three components that affect the run-time.
First, the number of explanation points scales linearly, which is clear and will not be further considered.
We thus will keep $m=1$ in the following.
Second, the \emph{tree complexity}, given by both, the number of leaves and depth of the tree affect the complexity jointly.
Note, that these values are highly dependent on each other.
Third, the order of interactions affects the complexity in an exponential manner, given by the binomial coefficient.
In the following, to account for irregularities in the processing times, we run every explanation $10$ times and average over the run-times and show the corresponding standard deviations.

\subsubsection{Naive Comparison}
\begin{figure*}[h]
    \centering
    \includegraphics[width=0.45\textwidth]{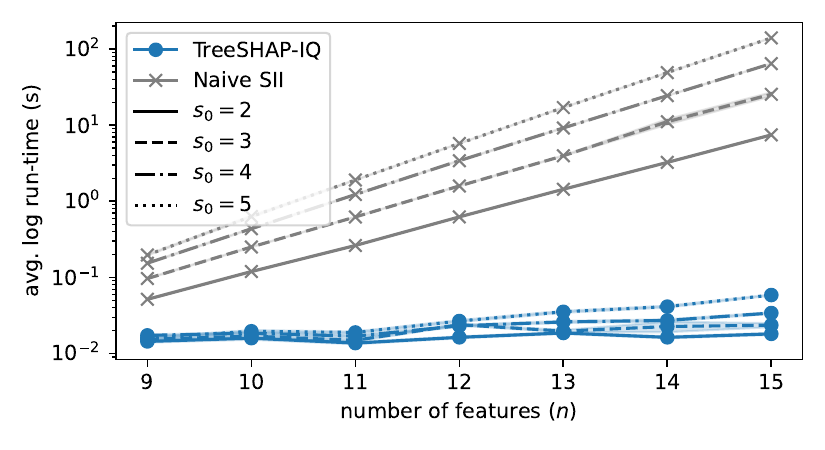}
    \caption{Log run-time of computing SII scores with TreeSHAP-IQ (blue) compared to naive calculation (grey). Naive computation scales exponentially.}
    \label{runtime_naive_comp}
\end{figure*}

We compare TreeSHAP-IQ's run-time for computing SII values up to order $s_0 \leq 5$ with a naive computation of SII.
For this comparison, we create a separate synthetic classification dataset of $5\,000$ samples and $n$ number of features for all $n \in [9, 15]$ (we use the \texttt{make\_classification} function from \texttt{sklearn}).
We fix the tree-depth to $8$ and fit a decision tree for each dataset.
We make sure that the trees all consists of approximately the same number of nodes (i.e. all trees reach the maximum depth).
We then compute the SII values up to order $s_0$ with TreeSHAP-IQ and through a naive brute force computation over all combinations of subsets.
We plot the log run-time of TreeSHAP-IQ and the naive SII computation in Figure~\ref{runtime_naive_comp}.
The comparison shows that the naive calculation scales exponentially with the number of features, while TreeSHAP-IQ scales polynomially for each interaction order $s_0$.

\subsubsection{Run-time by Tree Complexity}

\begin{figure*}[h]
    \centering
    \includegraphics[width=0.45\textwidth]{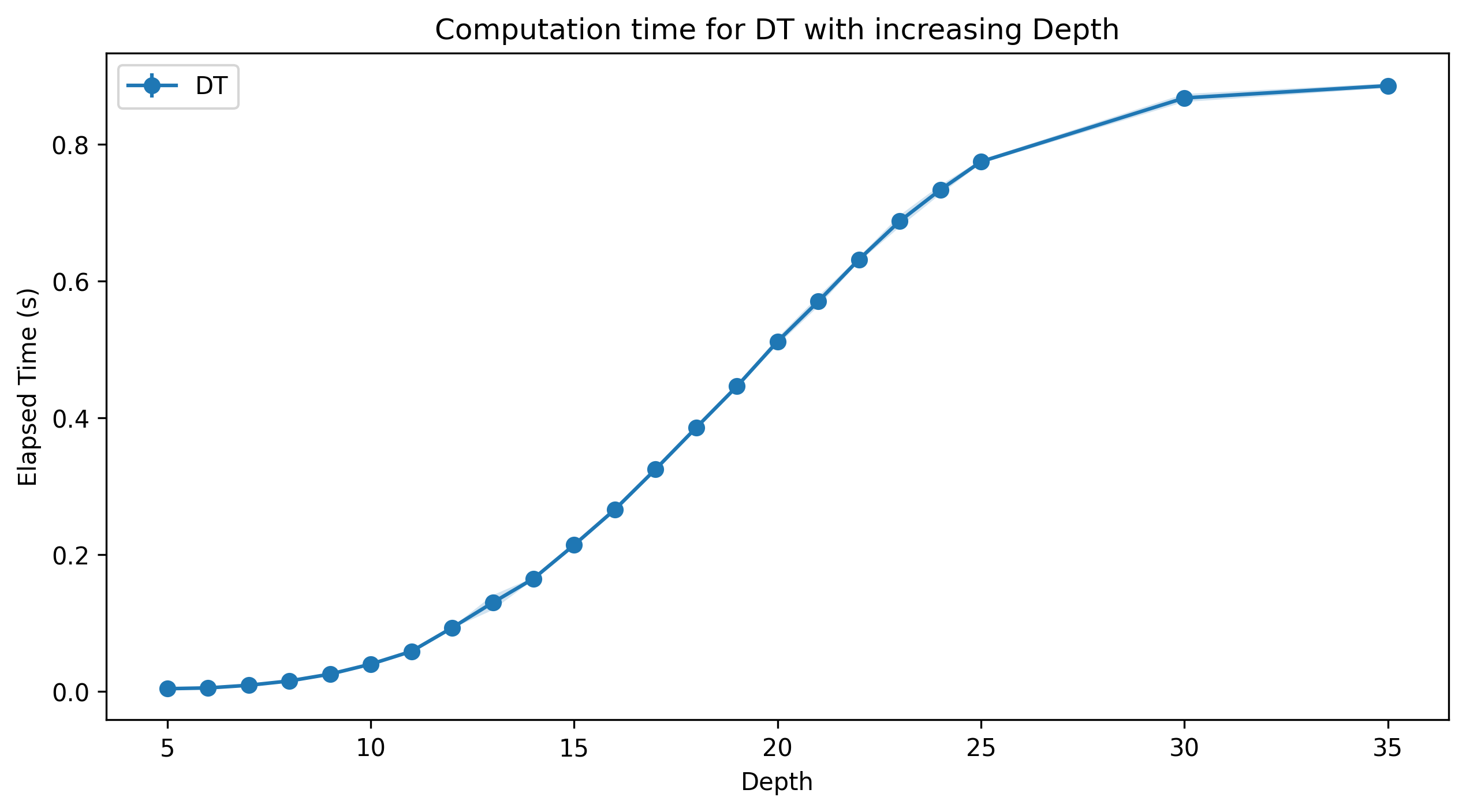}
    \includegraphics[width=0.45\textwidth]{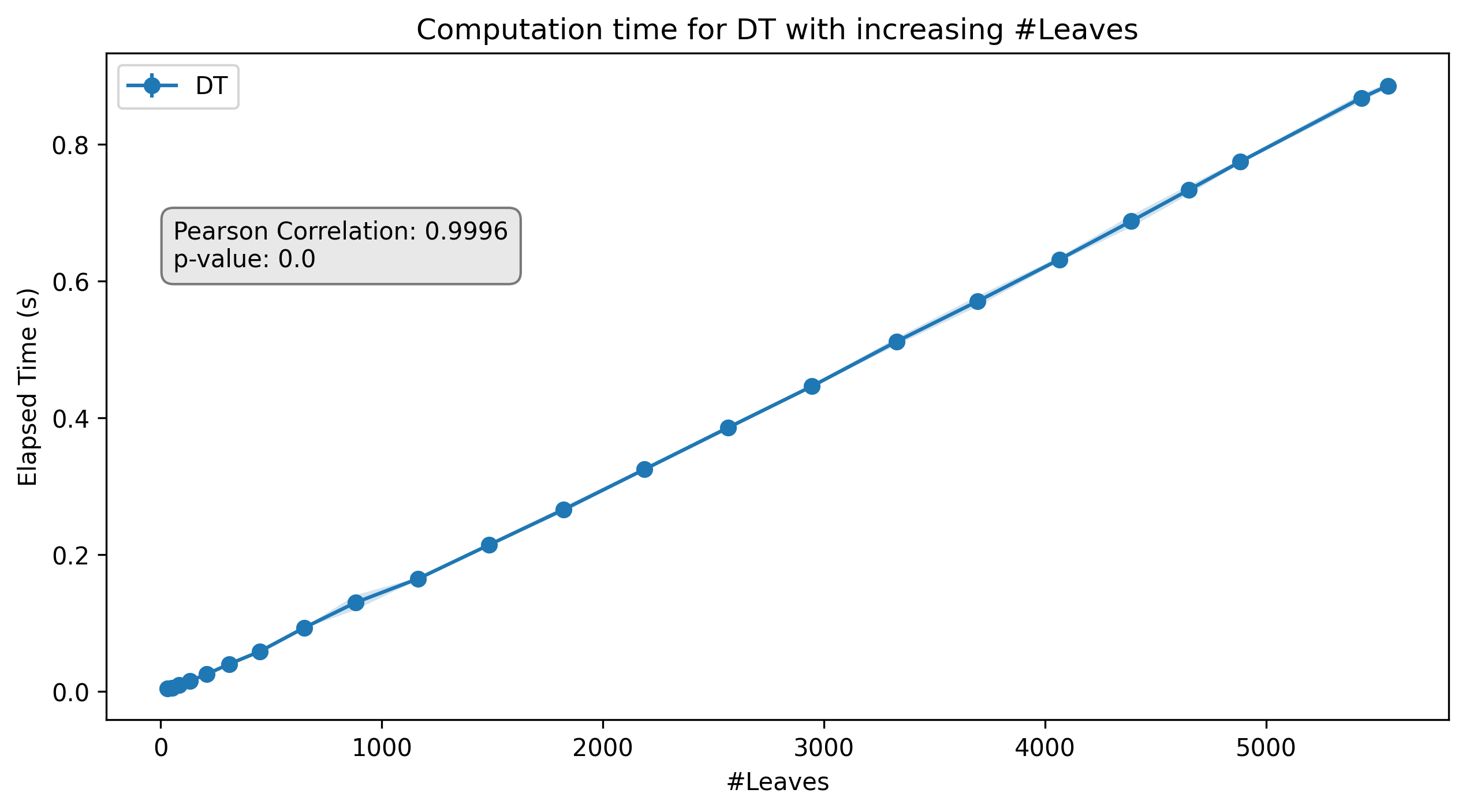}
    \\
    \includegraphics[width=0.45\textwidth]{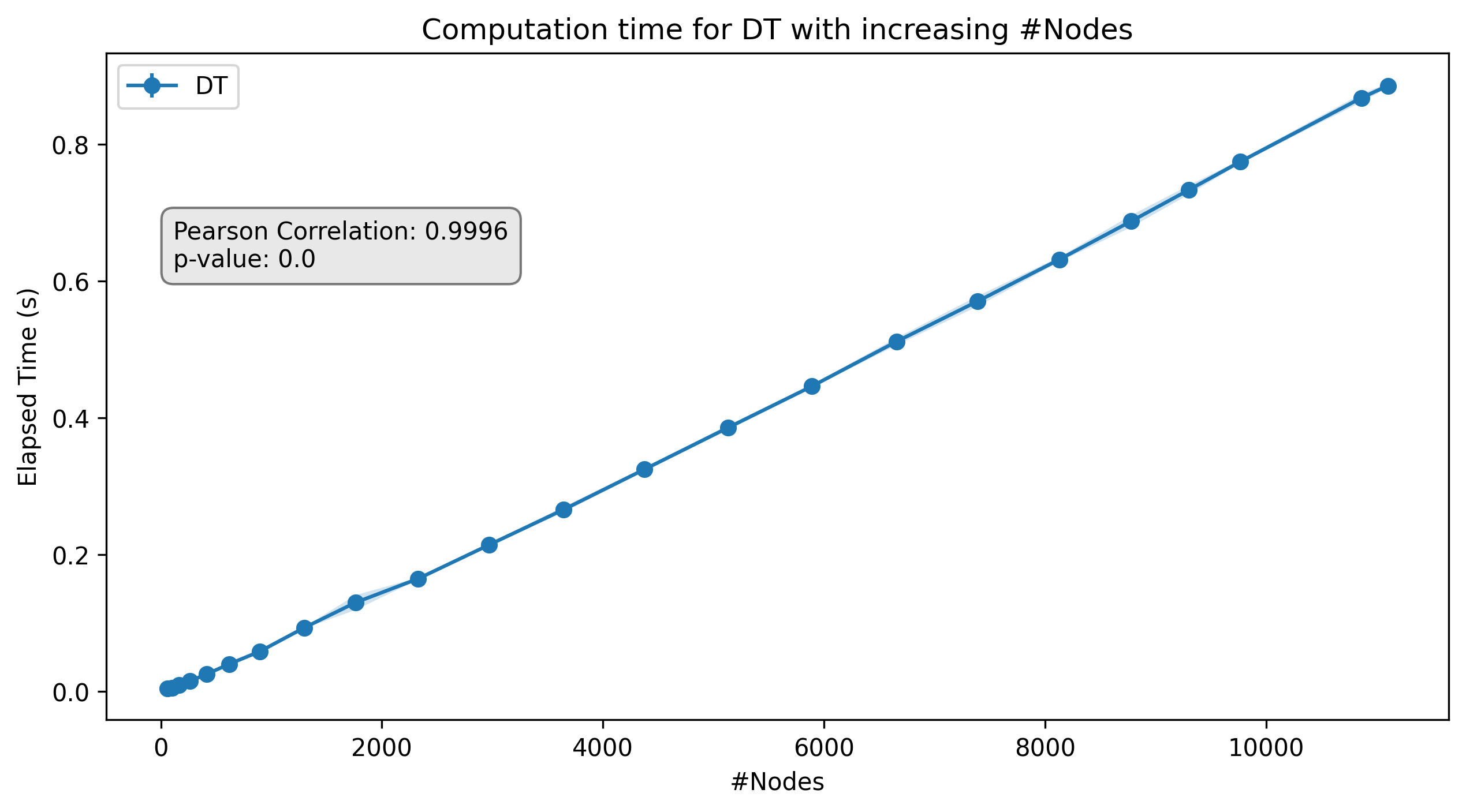}
    \includegraphics[width=0.45\textwidth]{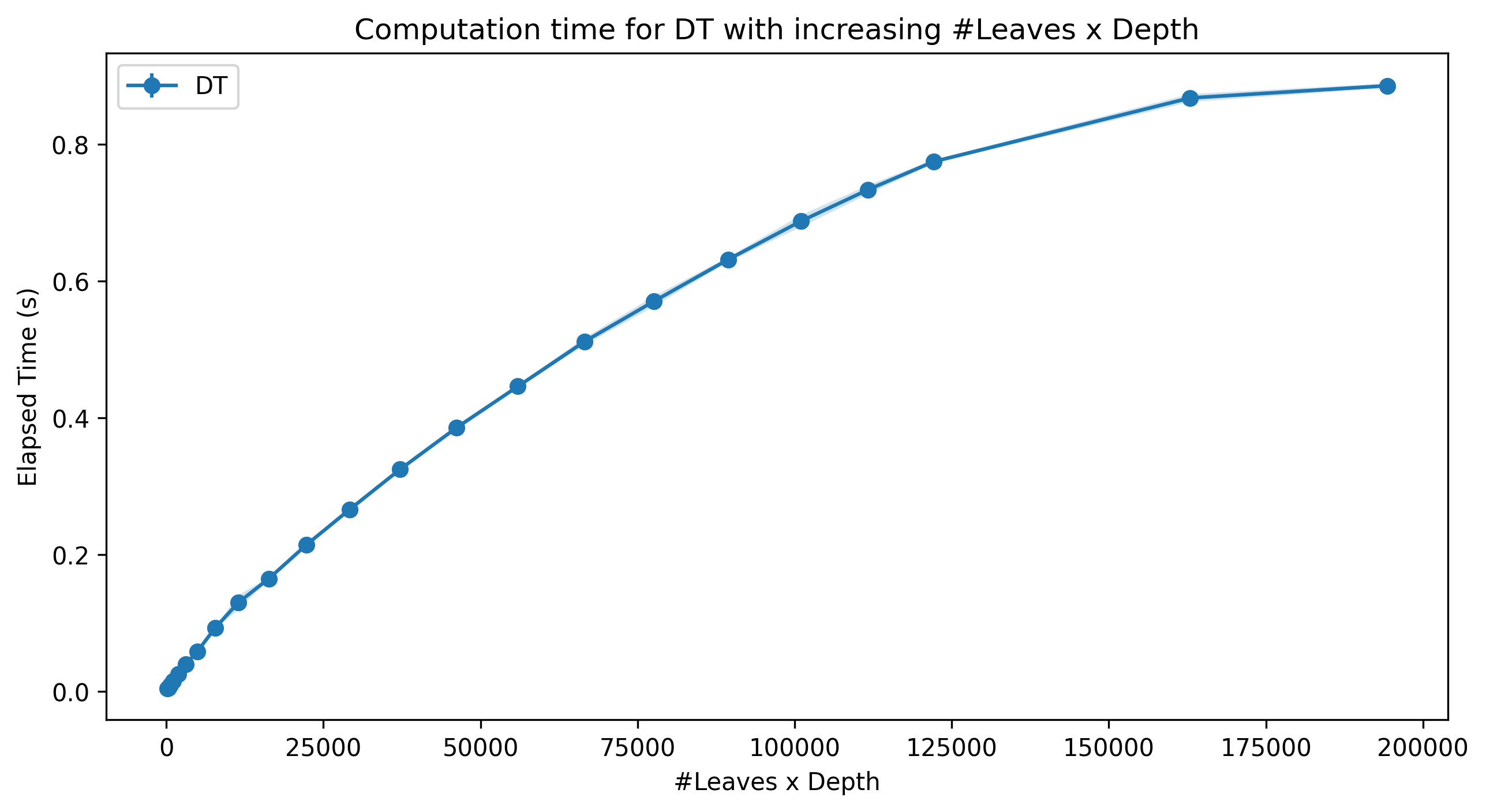}
    \caption{Run-time analysis of TreeSHAP-IQ for a single DT with varying tree complexity parameters}
    \label{fig_app_runtime_tree}
\end{figure*}

We now illustrate the run-time by the tree complexity, where we compute always pairwise interactions ($s=2$). The results are shown in Figure~\ref{fig_app_runtime_tree}.
We observe a linear relationship of the run-time and the number of vertices, as well as the number of leaves.
The run-time compared with the maximum depth of the tree admits a sub-linear behavior at higher depths.
This can be explained by the increasing number of paths in the DT that are shorter than the maximum depth, which increasingly occurs at higher depths.
Lastly, we also observe this sub-linear behavior in terms of the number of leaves times the depth, which again is explained by the overestimation of operations.

\subsubsection{Run-Time by Number of Interactions}
\begin{figure*}[h]
    \centering
    \includegraphics[width=0.45\textwidth]{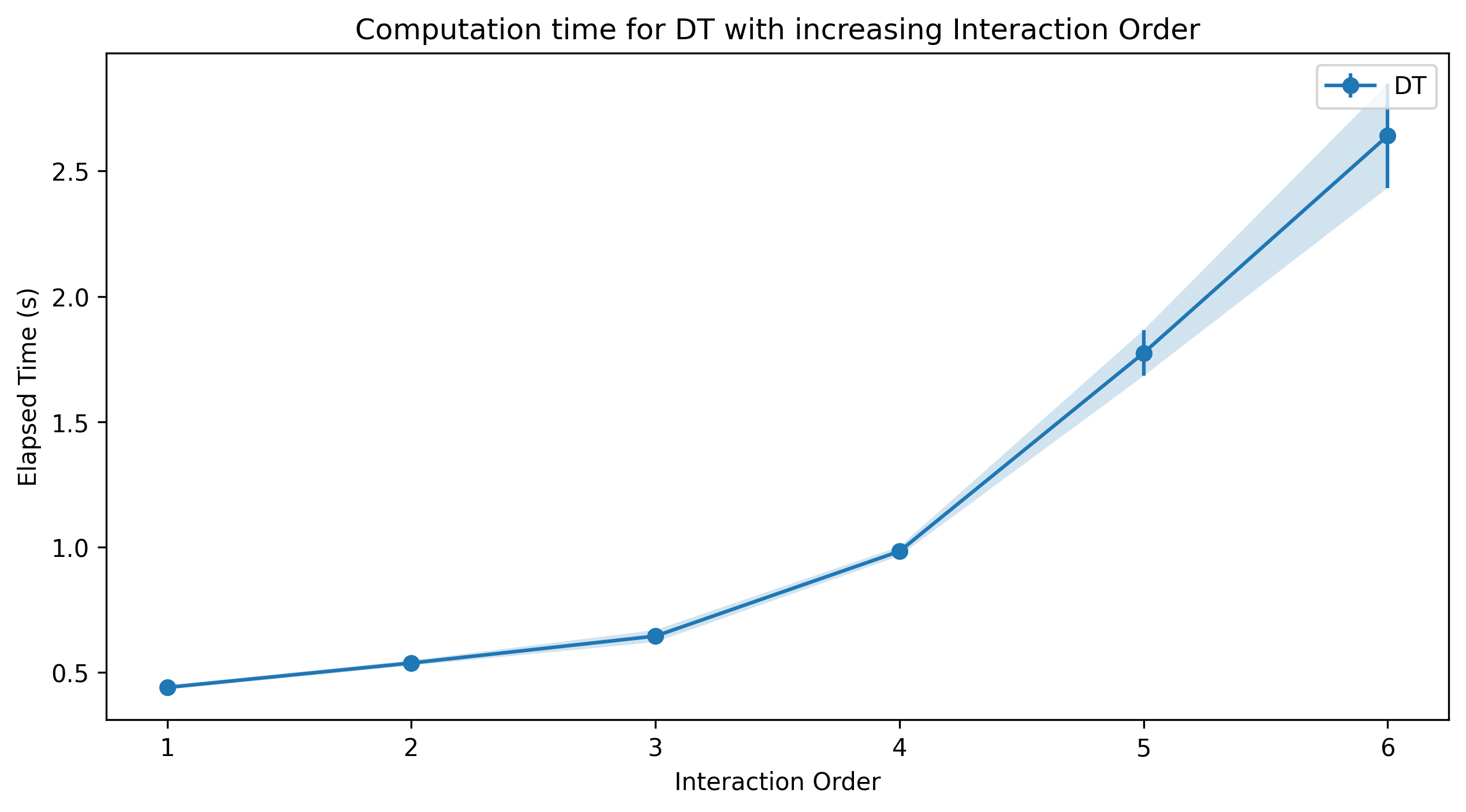}
    \caption{Run-time analysis of TreeSHAP-IQ for a single DT of depth $20$ with varying interaction order}
    \label{fig_app_runtime_order}
\end{figure*}

We illustrate the run-time depending on the number of interactions for a fixed DT of depth $20$.
The results are shown in Figure~\ref{fig_app_runtime_order}.
Note that interactions are only updated, if all features have appeared in the observed path.
This is results in far less evaluations, especially for higher order interactions.
Again, this is a consequence of the structure of the DT, where only very few paths admit the maximum depth.

\clearpage
\subsection{Additional n-SII Plots}
\begin{figure*}[h]
    \centering
    \includegraphics[width=0.4\textwidth]{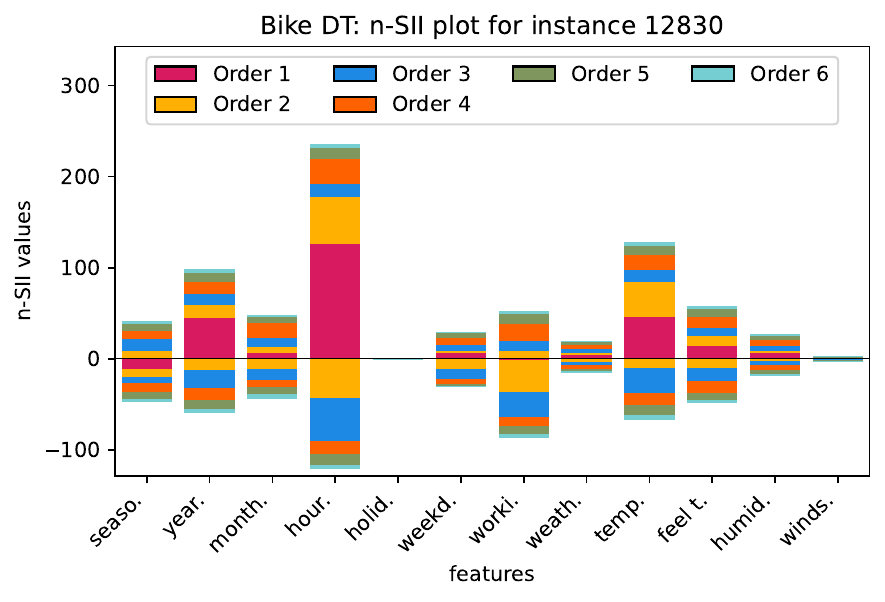}
    \includegraphics[width=0.4\textwidth]{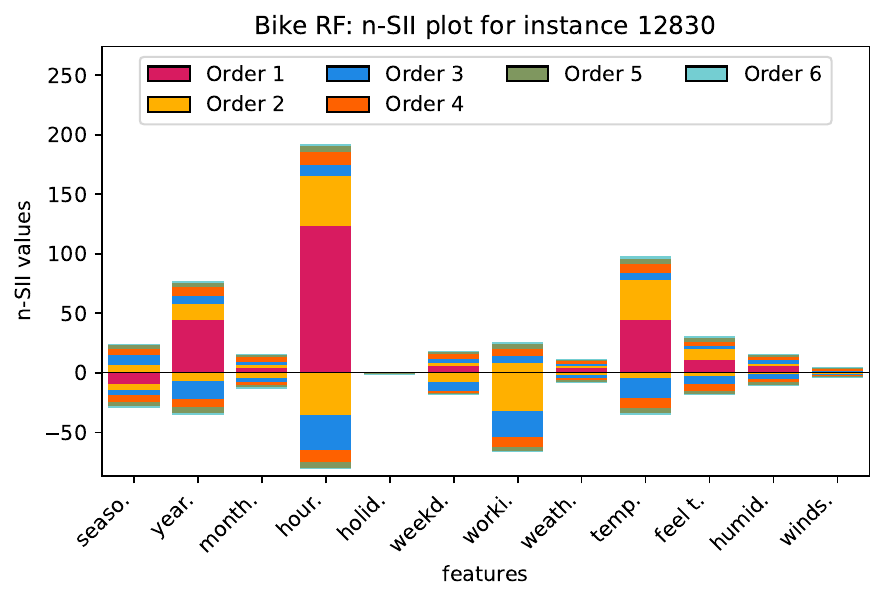}
    \\
    \includegraphics[width=0.4\textwidth]{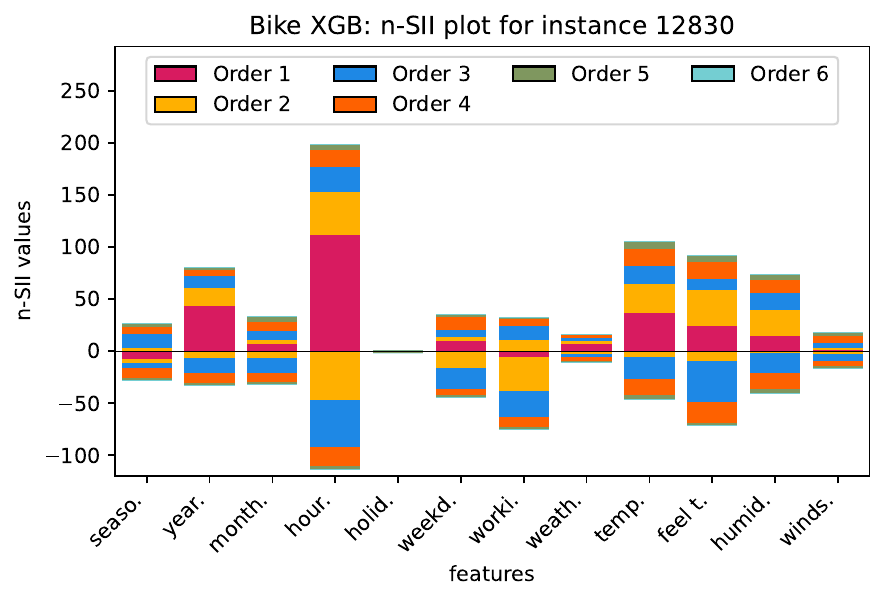}
    \includegraphics[width=0.4\textwidth]{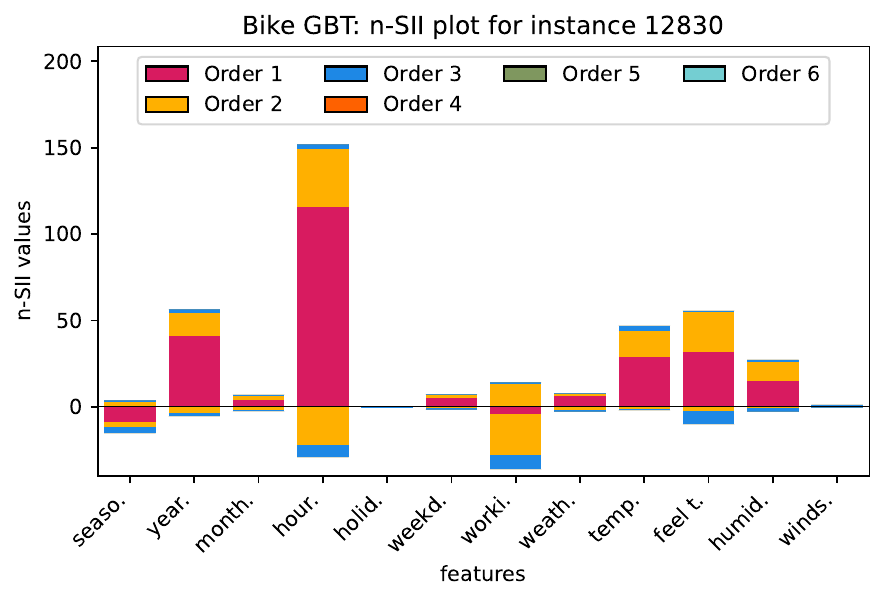}
    \caption{n-SII values up to order $s_0 = 6$ for a randomly selected instance of the \emph{Bike} dataset.}
    \label{fig_app_n_SII_bike_12830}
\end{figure*}

\begin{figure*}[h]
    \centering
    \includegraphics[width=0.4\textwidth]{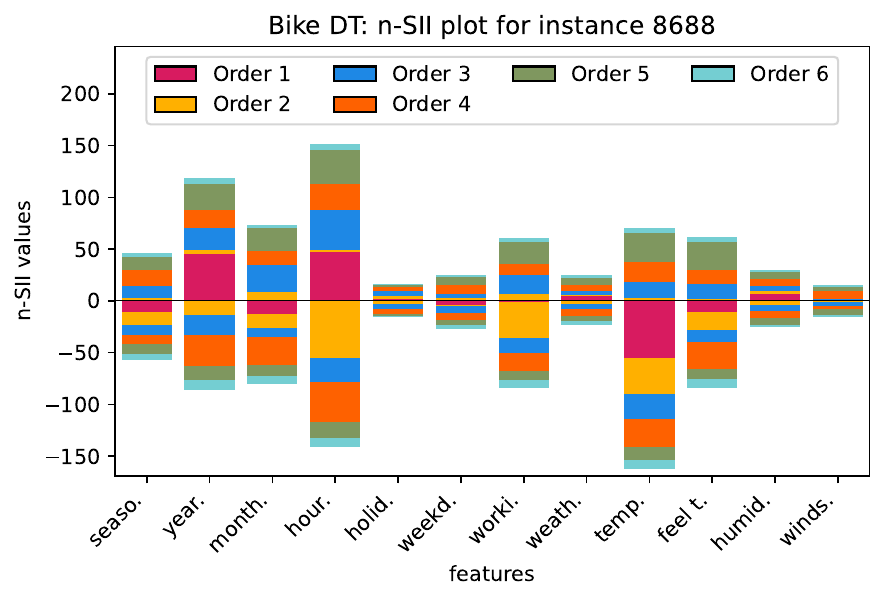}
    \includegraphics[width=0.4\textwidth]{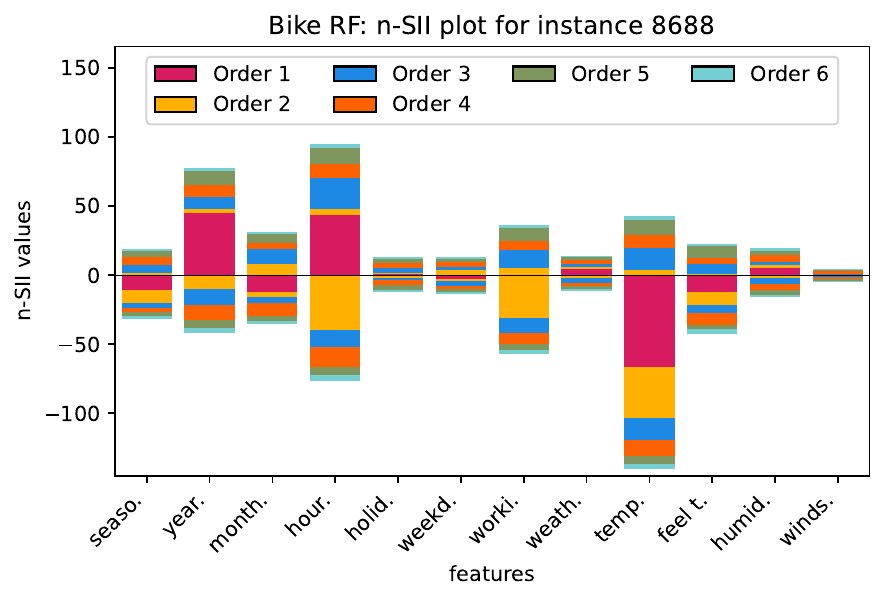}
    \\
    \includegraphics[width=0.4\textwidth]{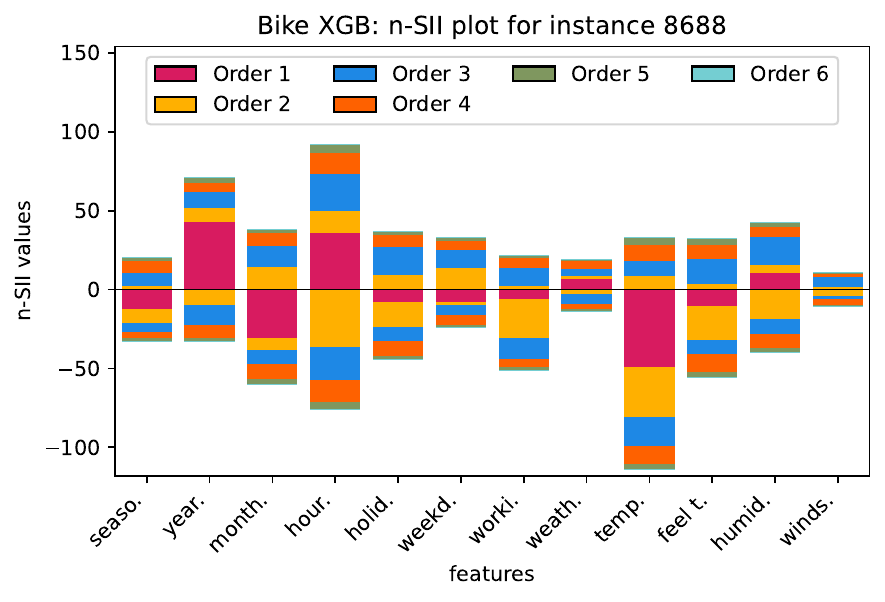}
    \includegraphics[width=0.4\textwidth]{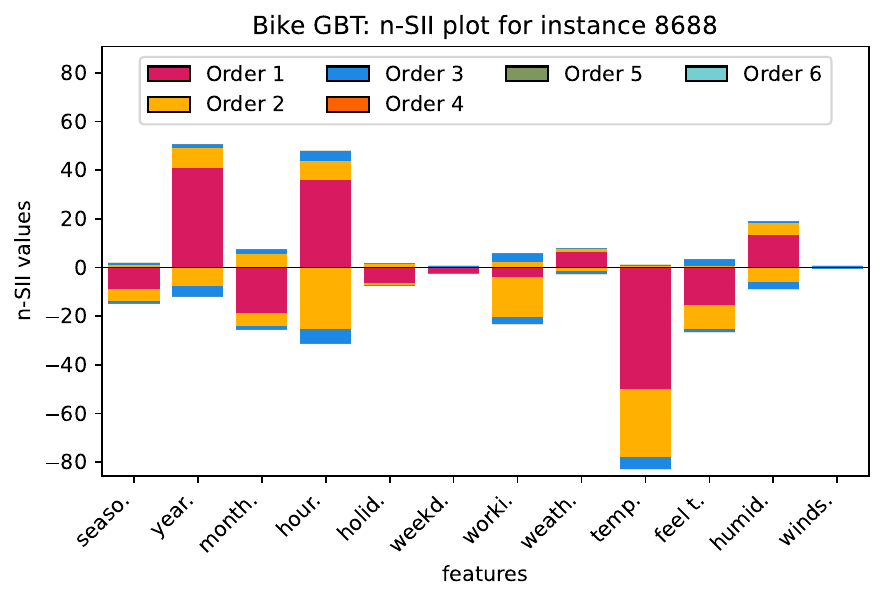}
    \caption{n-SII values up to order $s_0 = 6$ for a randomly selected instance of the \emph{Bike} dataset.}
    \label{fig_app_n_SII_bike_8688}
\end{figure*}

\begin{figure*}[h]
    \centering
    \includegraphics[width=0.4\textwidth]{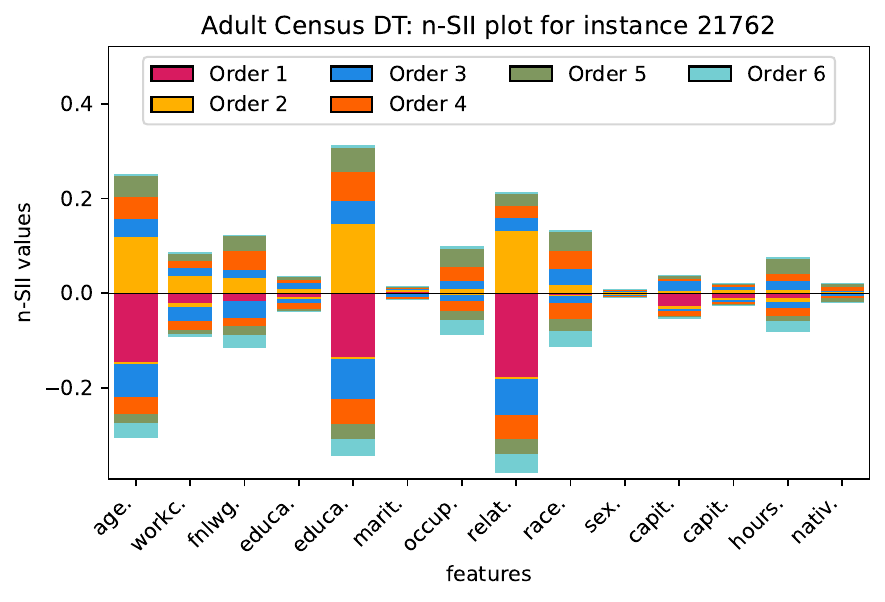}
    \includegraphics[width=0.4\textwidth]{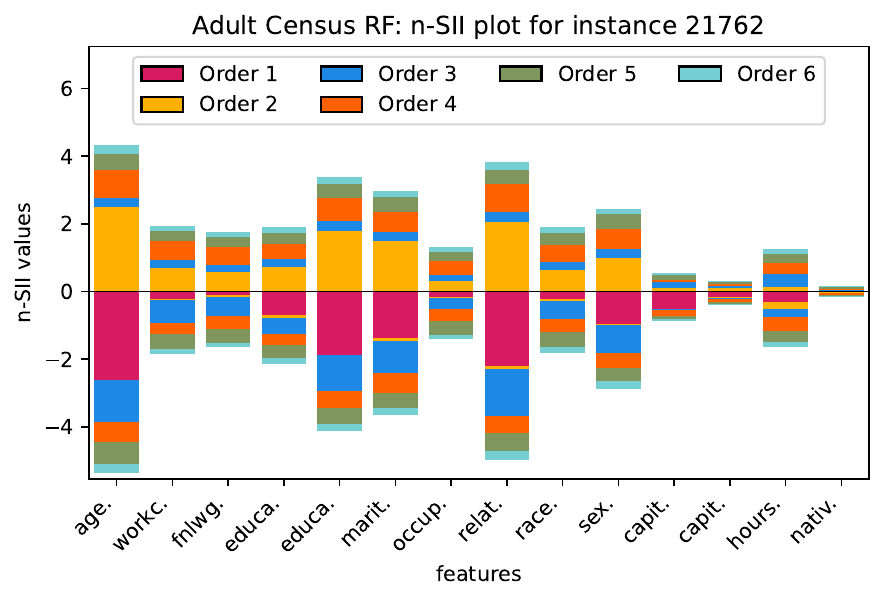}
    \\
    \includegraphics[width=0.4\textwidth]{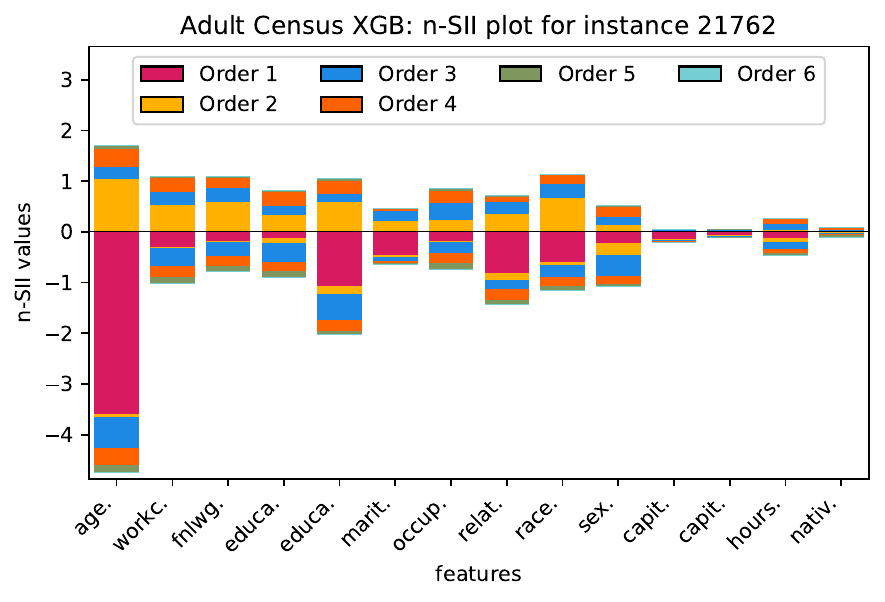}
    \includegraphics[width=0.4\textwidth]{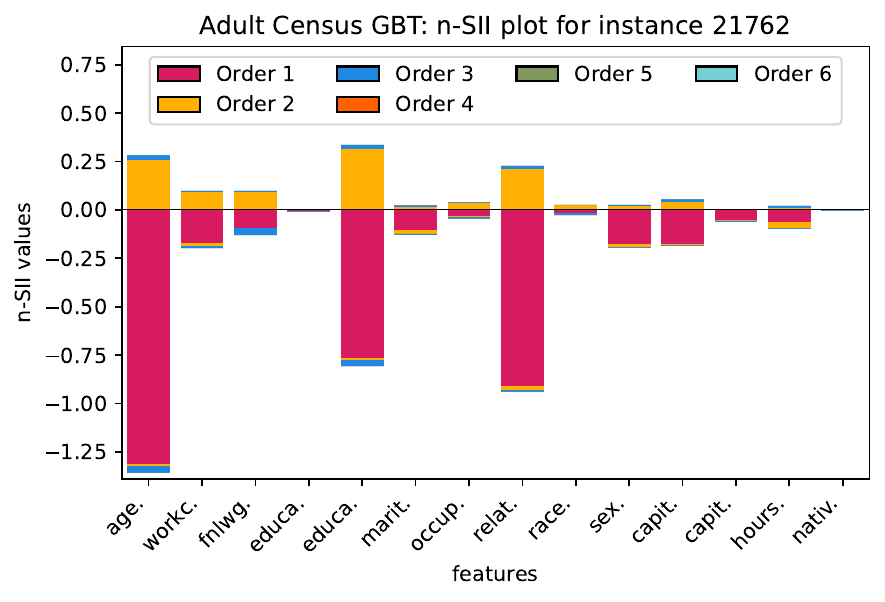}
    \caption{n-SII values up to order $s_0 = 6$ for a randomly selected instance of the \emph{Adult Census} dataset.}
    \label{fig_app_n_SII_adult_21762}
\end{figure*}

\begin{figure*}[h]
    \centering
    \includegraphics[width=0.4\textwidth]{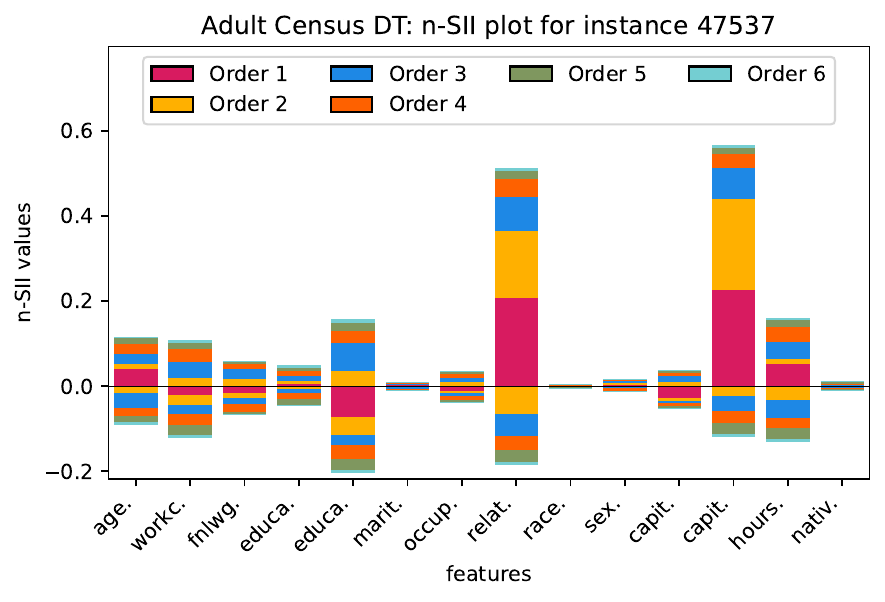}
    \includegraphics[width=0.4\textwidth]{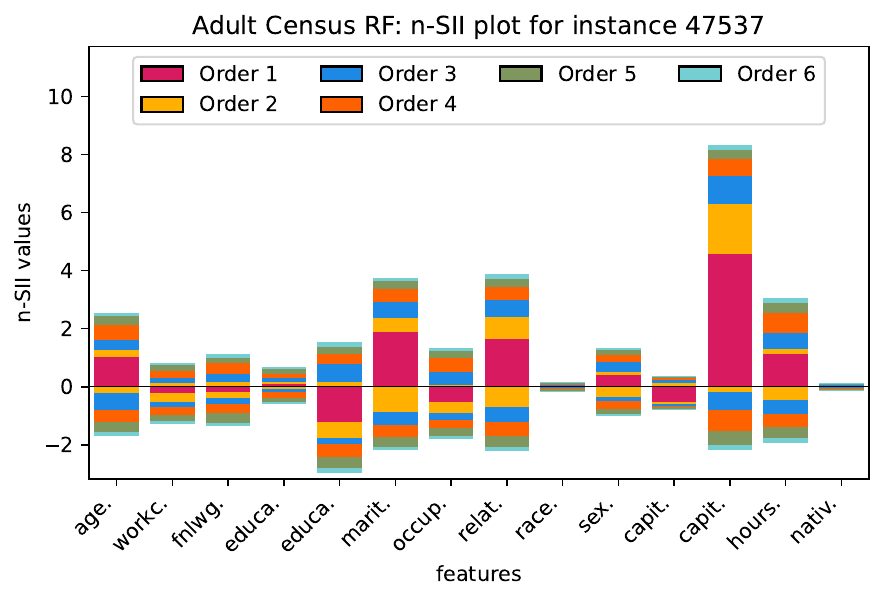}
    \\
    \includegraphics[width=0.4\textwidth]{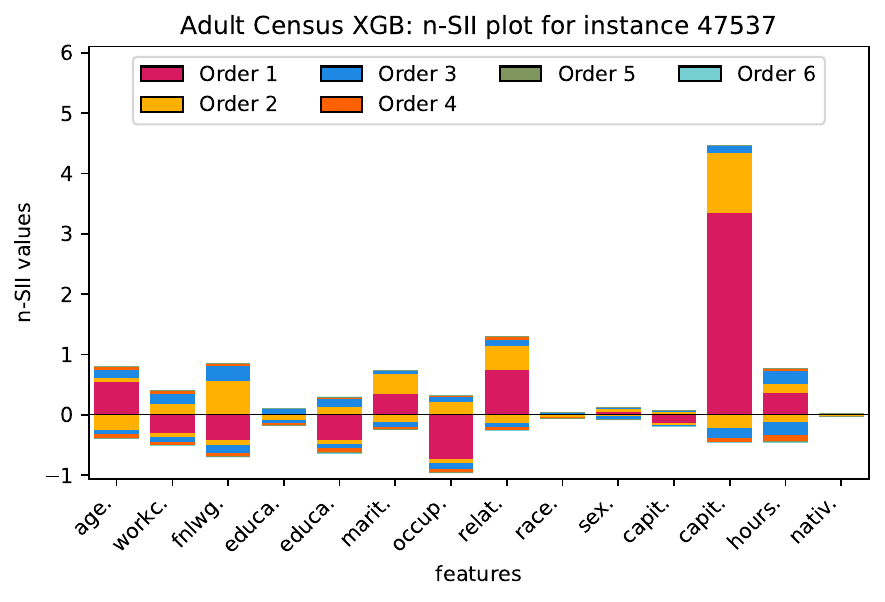}
    \includegraphics[width=0.4\textwidth]{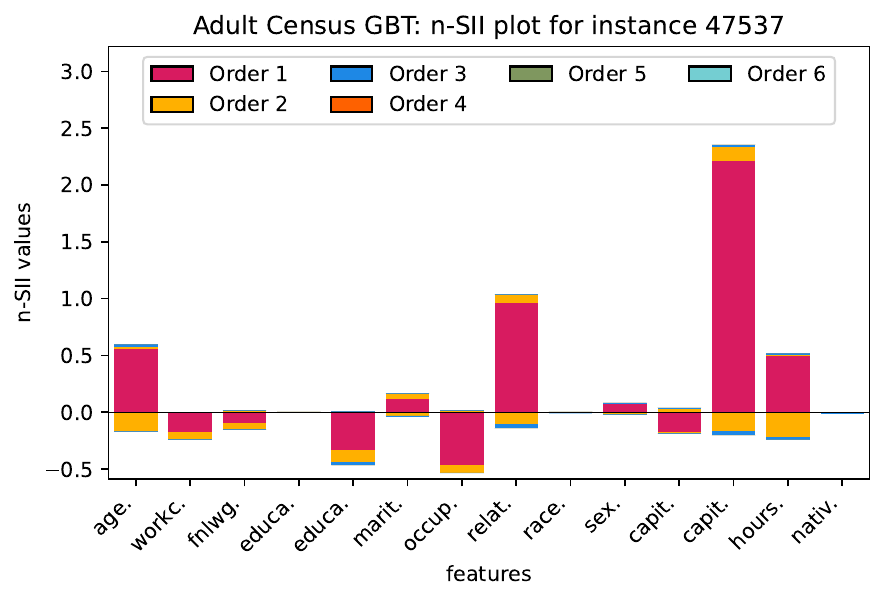}
    \caption{n-SII values up to order $s_0 = 6$ for a randomly selected instance of the \emph{Adult Census} dataset.}
    \label{fig_app_n_SII_adult_47537}
\end{figure*}

In this section, we compare the strength and nature of interactions present in different model architectures.
We generate the n-SII plots of order $s_0=6$ for the predictions of two randomly selected instances in the \emph{Bike} and \emph{Adult Census} dataset.
The results for the \emph{bike} dataset are shown in Figure~\ref{fig_app_n_SII_bike_12830} and Figure~\ref{fig_app_n_SII_bike_8688}.
The results for the \emph{Adult Census} dataset are shown in Figure~\ref{fig_app_n_SII_adult_21762} and Figure~\ref{fig_app_n_SII_adult_47537}.

We observe that the levels of interaction effects differ significantly among different predictions and different model architectures.
For the \emph{Bike} dataset, the prediction in Figure~\ref{fig_app_n_SII_bike_8688} has higher interaction effects than the prediction in Figure~\ref{fig_app_n_SII_bike_12830}.
Further, it can be seen that the well-performing XGB model exhibits a high amount of interaction, whereas the relatively poor-performing GBT has little interactions present.
Furthermore, as expected the DT exhibits a high level of interaction effects among both predictions, as this method is not an esemble of weak learning algorithms, which are expected to have less interaction effects.

In the \emph{Adult Census} dataset, we observe a similar pattern for XBG and GBT.
Notably, for these instances, XGB exhibits less interaction effects than RF, although the overall performance of XGB is superior.
Again, we observe a high amoung of higher order interactions present in the DT.
However, its performance is worse than for all ensemble methods.
Notably, the RF and the DT exhibit more similar interaction effects than the other models.
This could be seen as an indication that the learned functional relationship in gradient boosting approaches differs from classical DT learning schemes.
However, observing two local interaction effects does not allow to conclude this rigorously, which would require a global study of interaction effects in the corresponding models.

\clearpage
\subsection{Further Experimental Results on Benchmark Datasets}

\subsubsection{German Credit Dataset}
\begin{figure*}[h]
    \centering
    \includegraphics[width=0.45\textwidth]{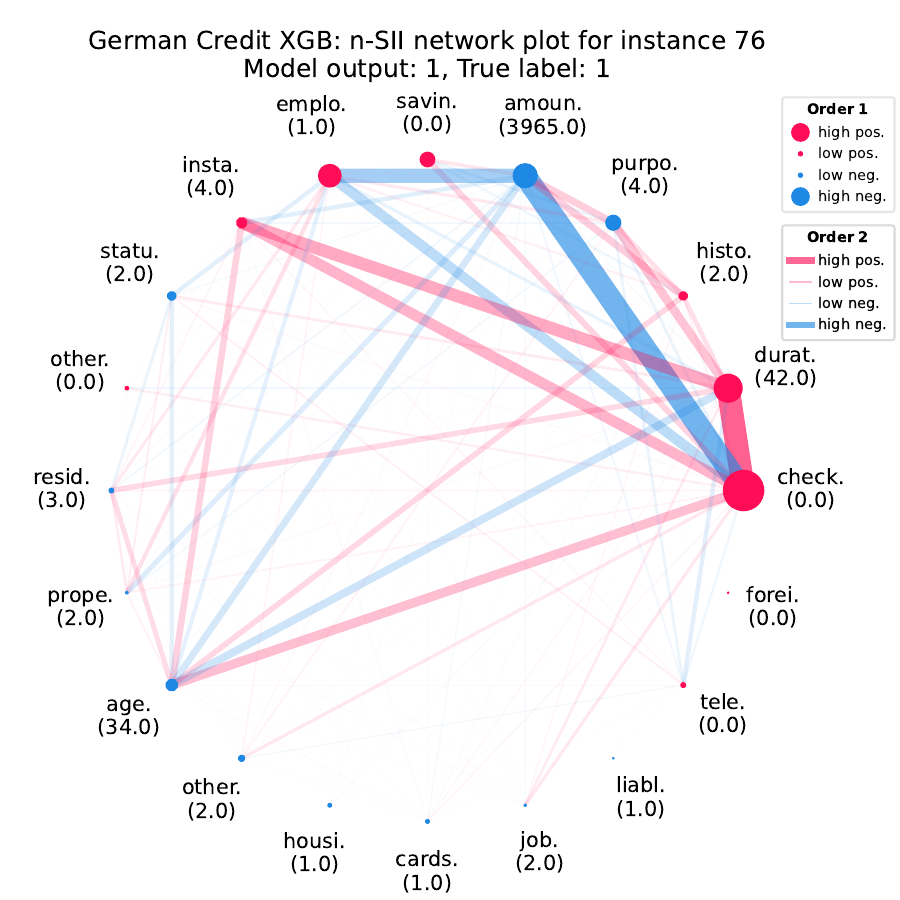}
    \includegraphics[width=0.45\textwidth]{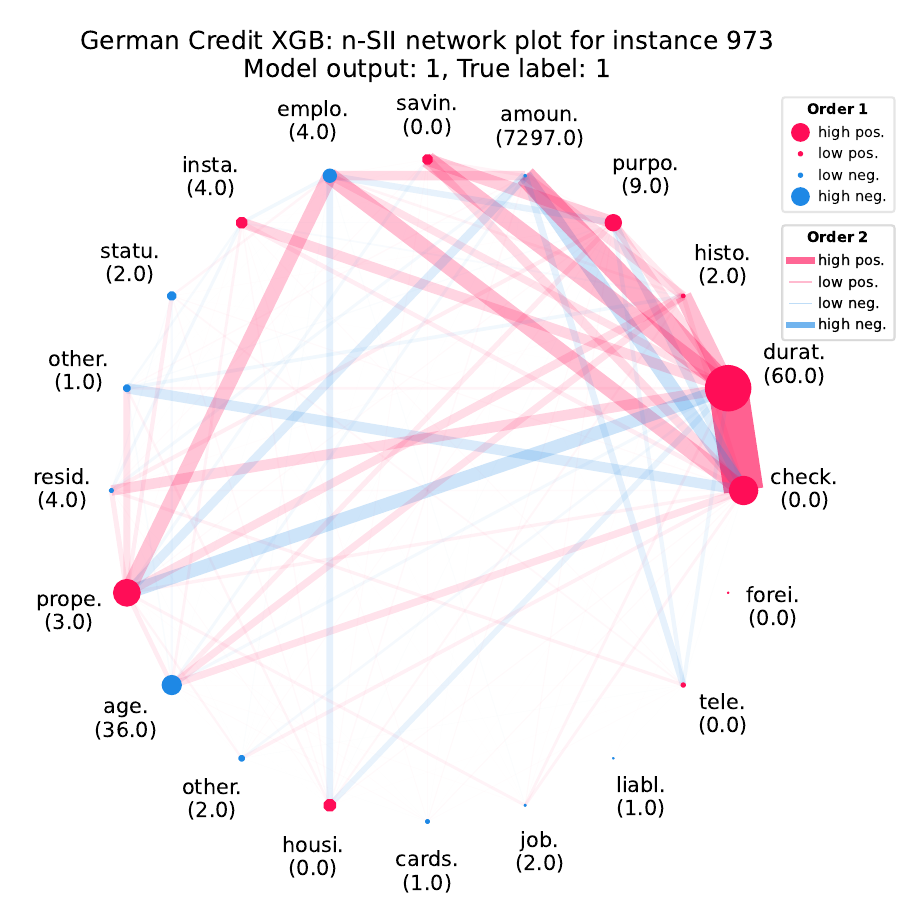}
    \\
    \includegraphics[width=0.45\textwidth]{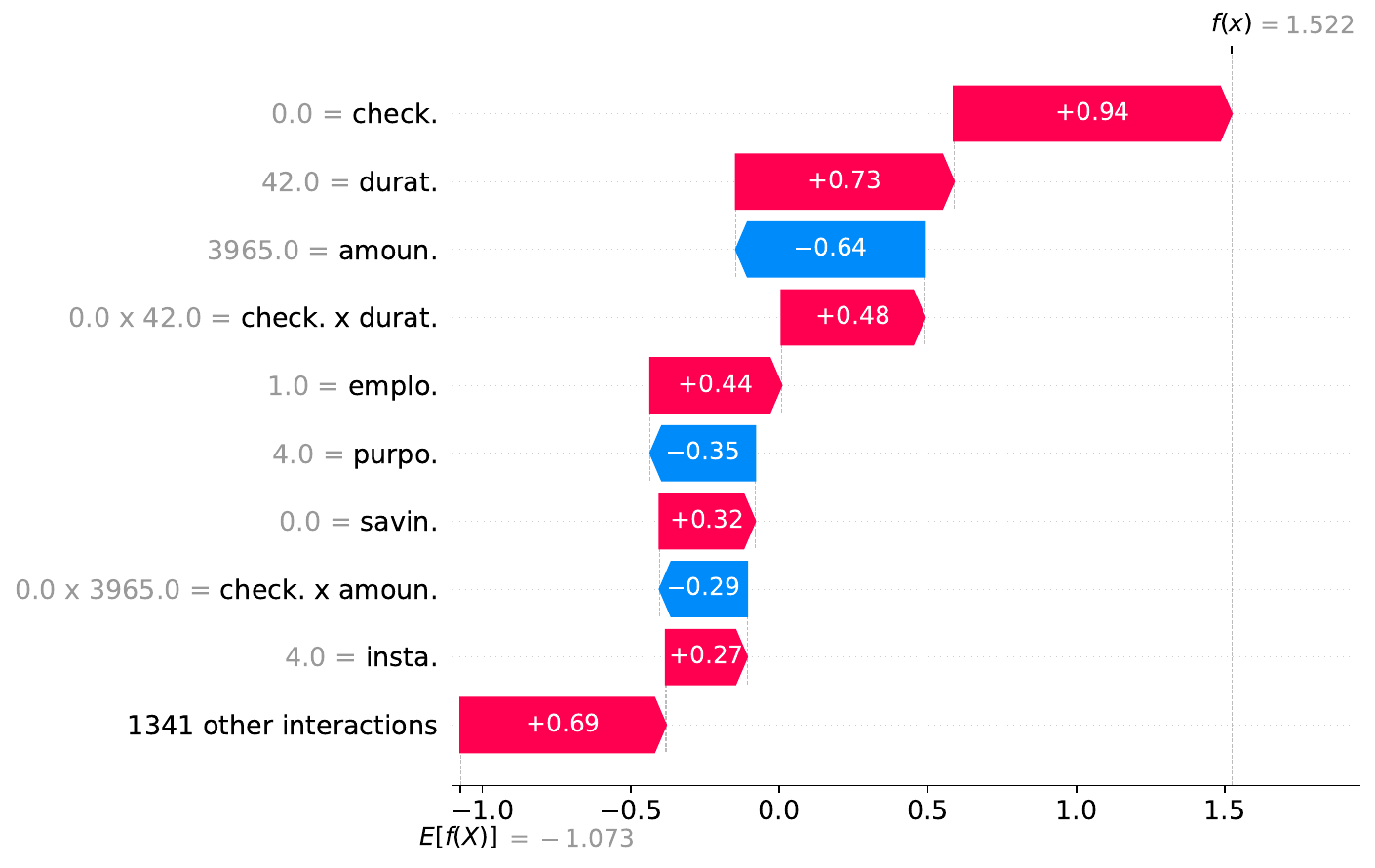}
    \includegraphics[width=0.45\textwidth]{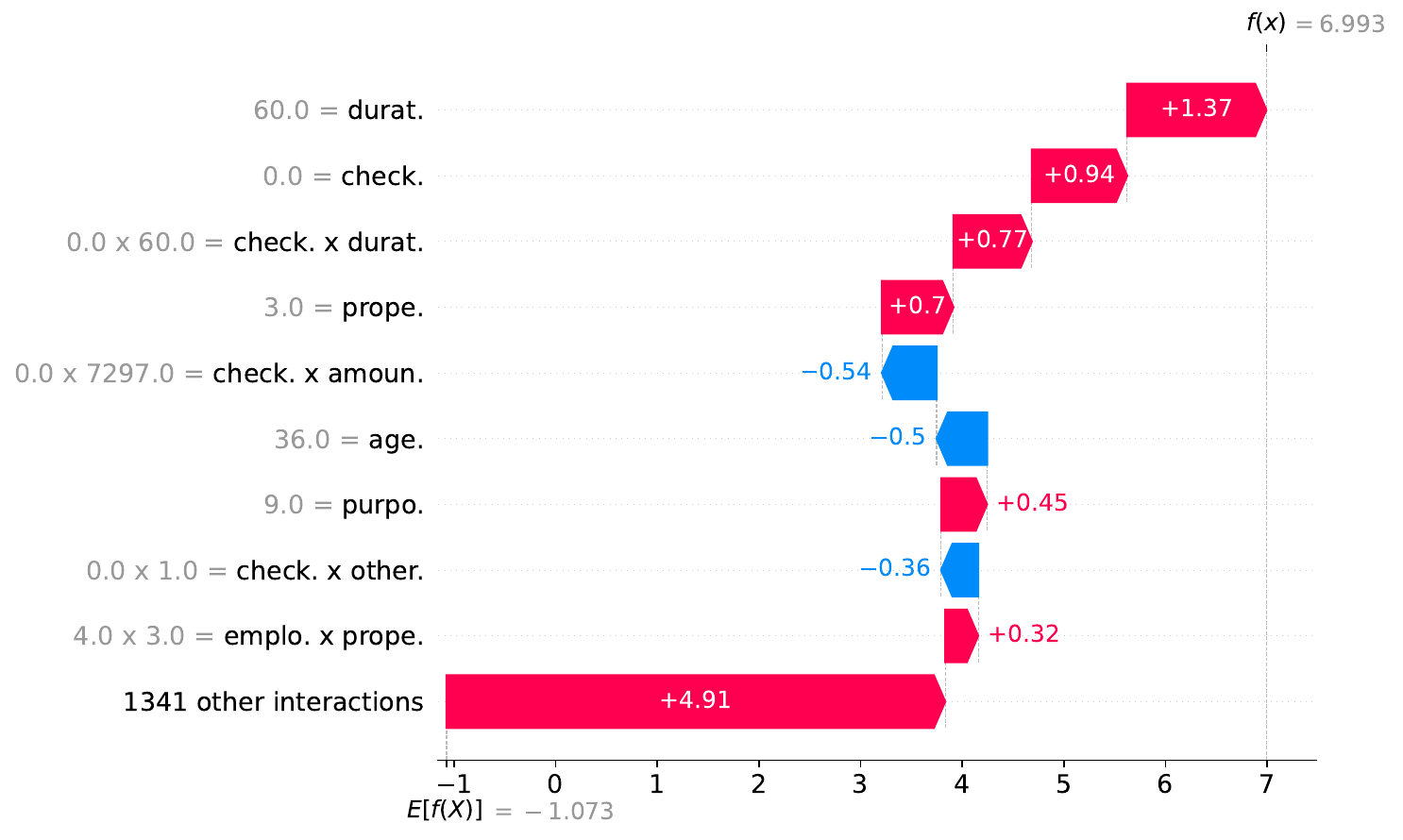}
    \caption{n-SII scores for order $s_0=2$ in a network plot (top) and $s_0=3$ in a waterfall chart (bottom) for two randomly selected instances of the \emph{German Credit} dataset predicted with a XGB}
    \label{fig_app_german}
\end{figure*}

We display n-SII interaction effects up to order $s_0=2$ in a network plot and effects up to order $s_0=3$ in a waterfall chart for two randomly selected instances of the \emph{German Credit} dataset predicted with a XGB.
The results are shown in Figure~\ref{fig_app_german}.

\clearpage
\subsubsection{Bank Dataset}
\begin{figure*}[h]
    \centering
    \includegraphics[width=0.45\textwidth]{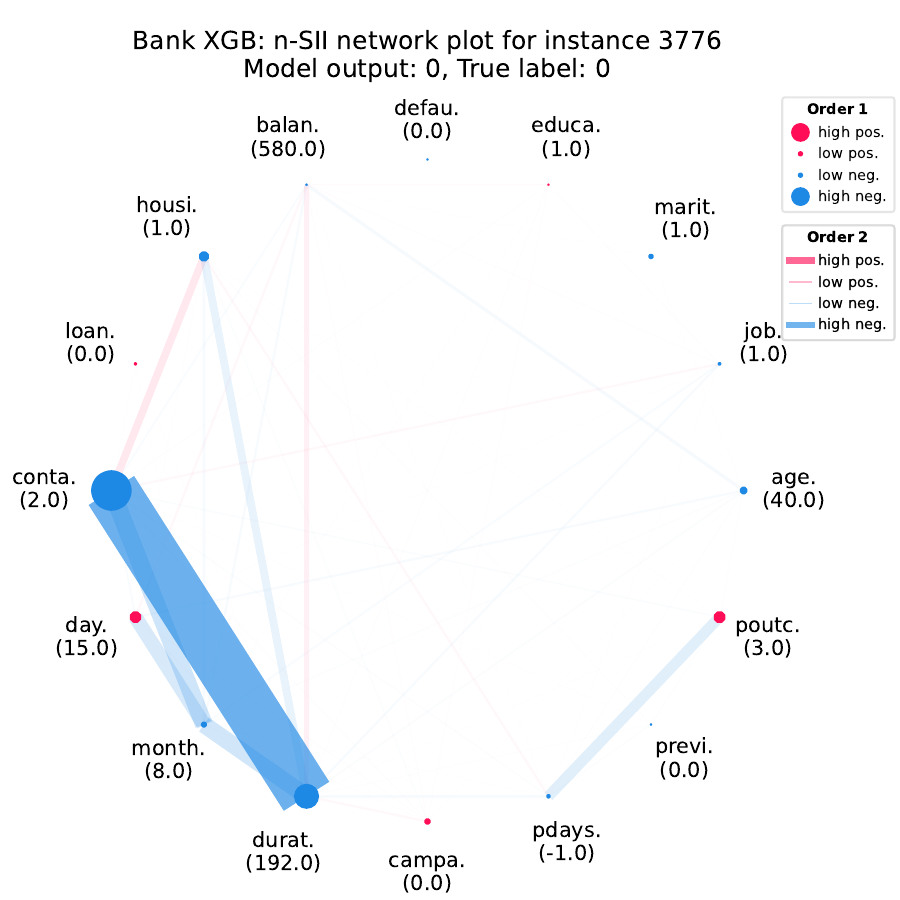}
    \includegraphics[width=0.45\textwidth]{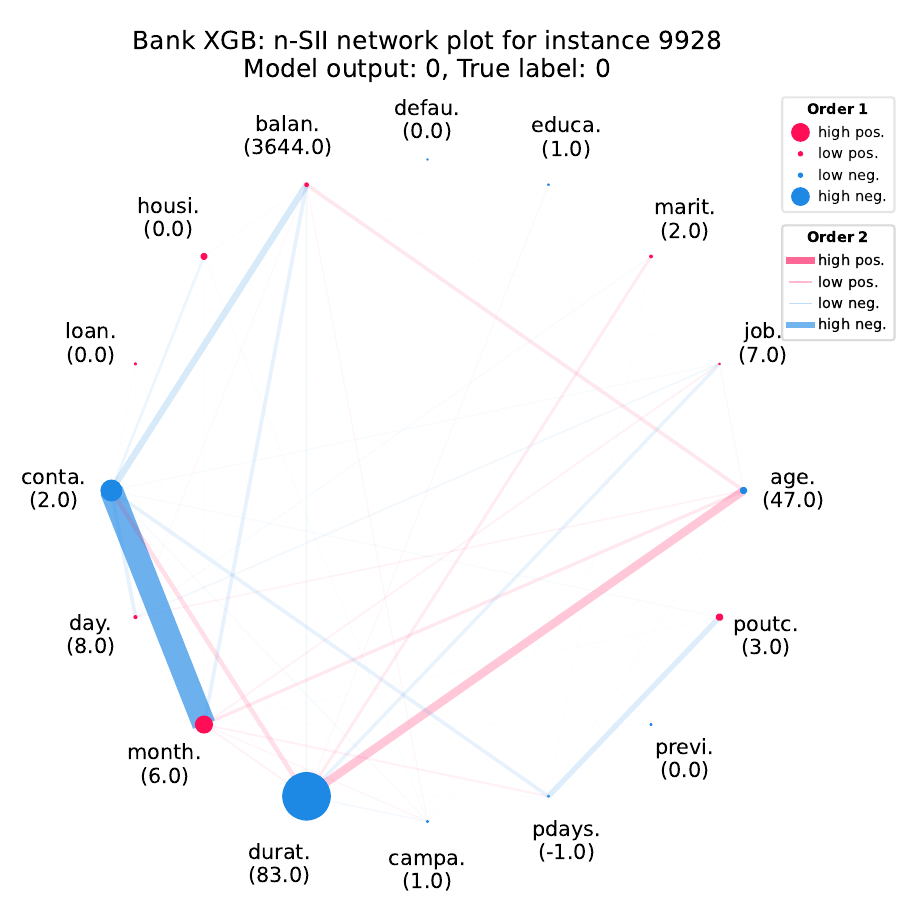}
    \\
    \includegraphics[width=0.45\textwidth]{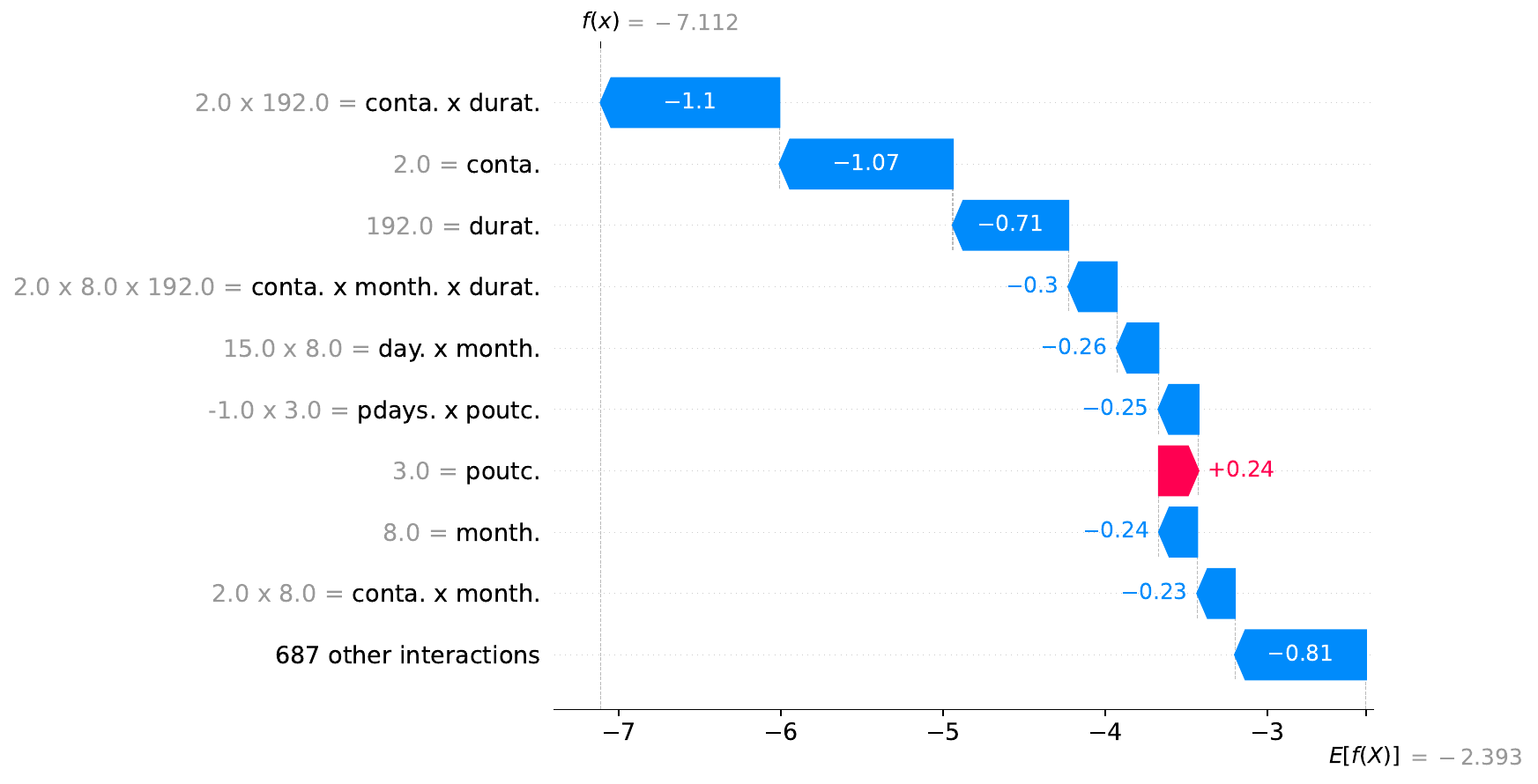}
    \includegraphics[width=0.45\textwidth]{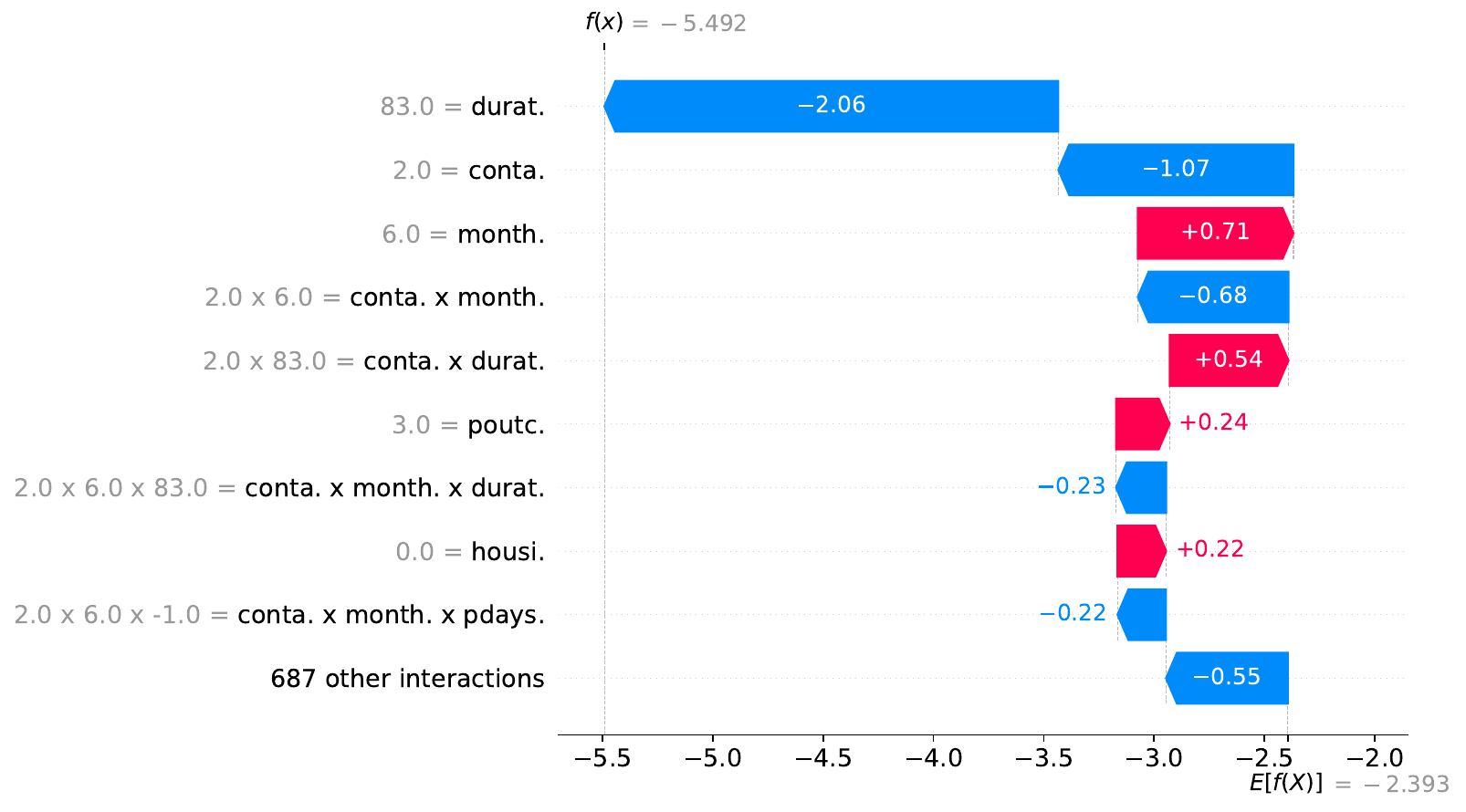}
    \caption{n-SII scores for order $s_0=2$ in a network plot (top) and $s_0=3$ in a waterfall chart (bottom) for two randomly selected instances of the \emph{Bank} dataset predicted with XGB}
    \label{fig_app_bank}
\end{figure*}

We display n-SII interaction effects up to order $s_0=2$ in a network plot and effects up to order $s_0=3$ in a waterfall chart for two randomly selected instances of the \emph{Bank} dataset predicted with XGB.
The results are shown in Figure~\ref{fig_app_bank}.

\clearpage
\subsubsection{Adult Census Dataset}
\begin{figure*}[h]
    \centering
    \includegraphics[width=0.45\textwidth]{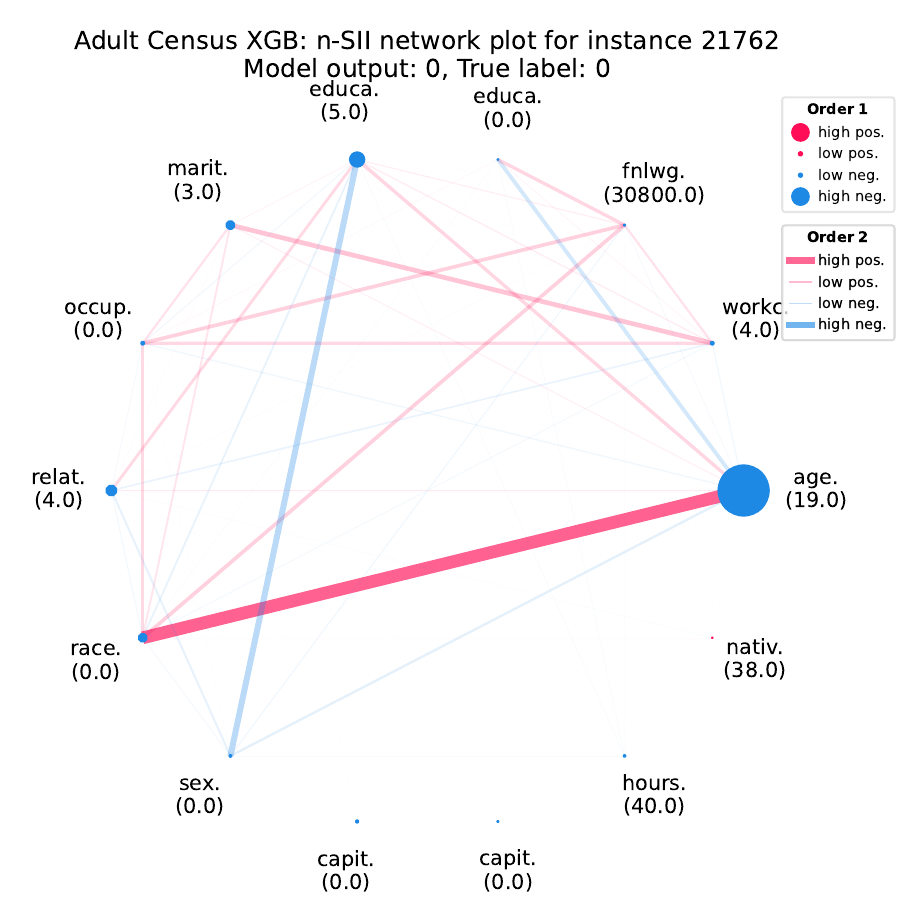}
    \includegraphics[width=0.45\textwidth]{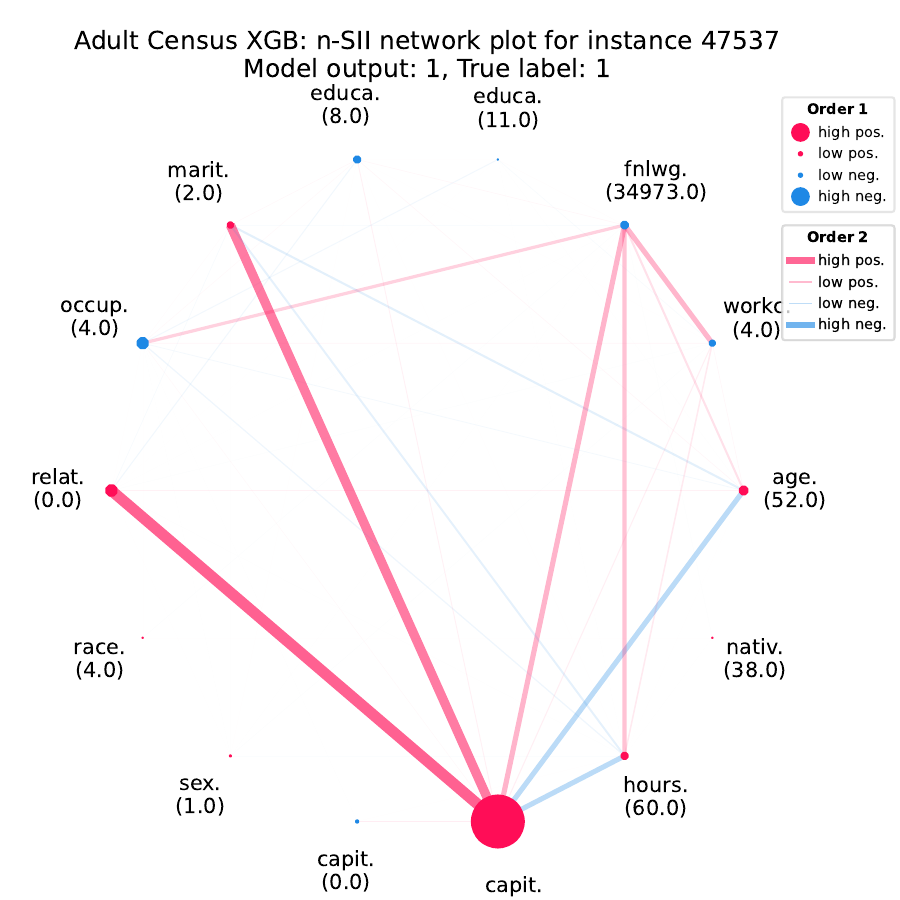}
    \\
    \includegraphics[width=0.45\textwidth]{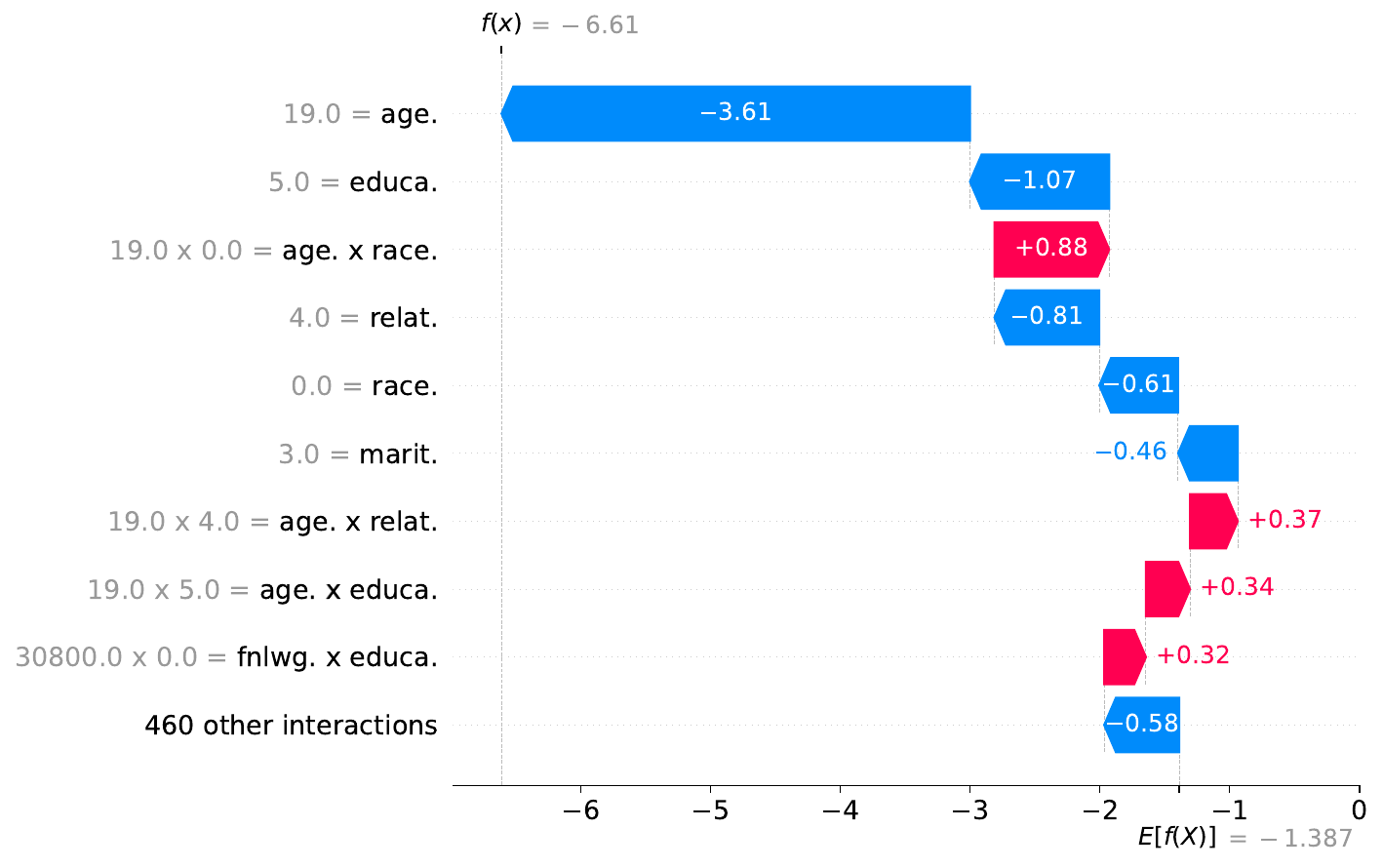}
    \includegraphics[width=0.45\textwidth]{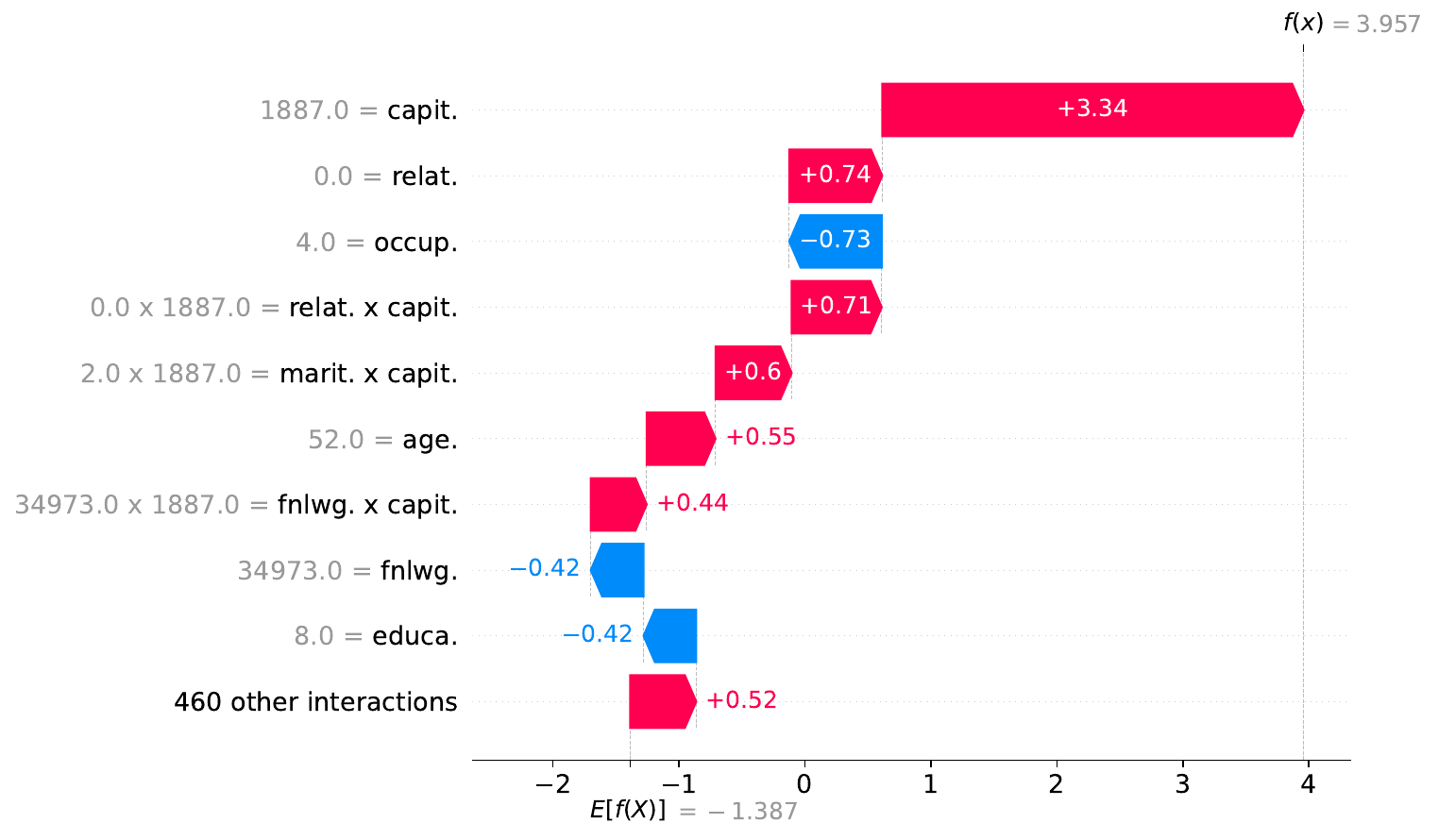}
    \caption{n-SII scores for order $s_0=2$ in a network plot (top) and $s_0=3$ in a waterfall chart (bottom) for two randomly selected instances of the \emph{Adult Census} dataset predicted with XGB}
    \label{fig_app_adult}
\end{figure*}

We display n-SII interaction effects up to order $s_0=2$ in a network plot and effects up to order $s_0=3$ in a waterfall chart for two randomly selected instances of the \emph{Adult Census} dataset predicted with XGB.
The results are shown in Figure~\ref{fig_app_adult}.

\clearpage
\subsubsection{Bike Dataset}
\begin{figure*}[h]
    \centering
    \includegraphics[width=0.45\textwidth]{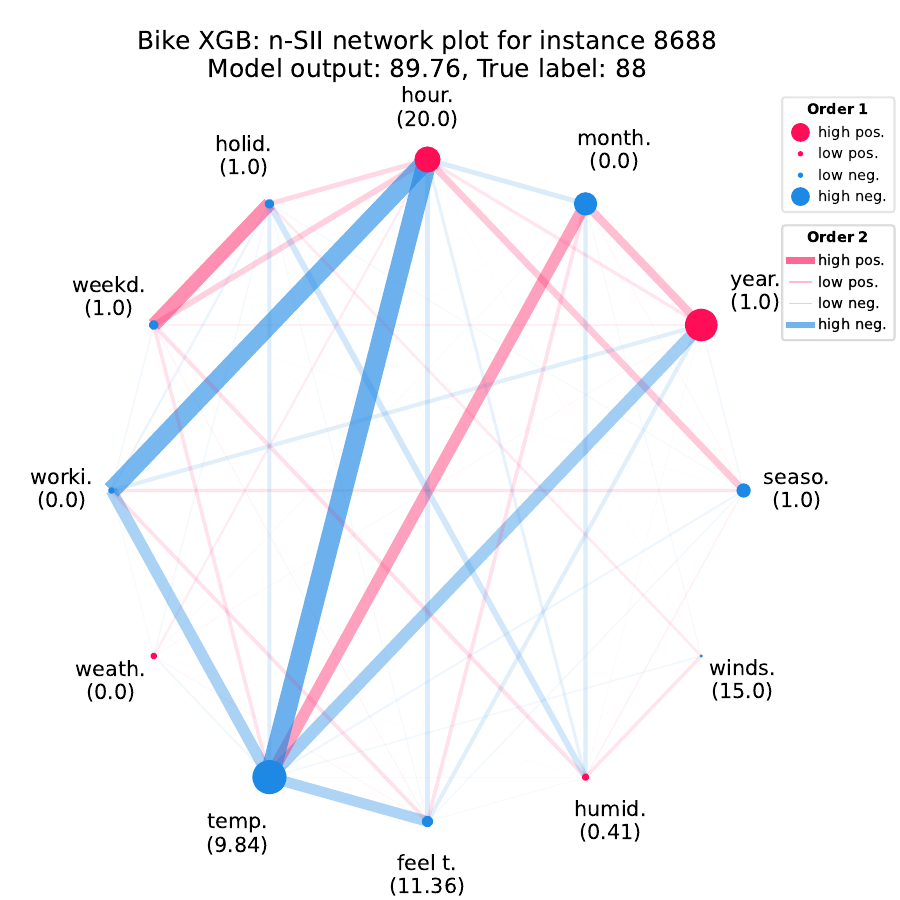}
    \includegraphics[width=0.45\textwidth]{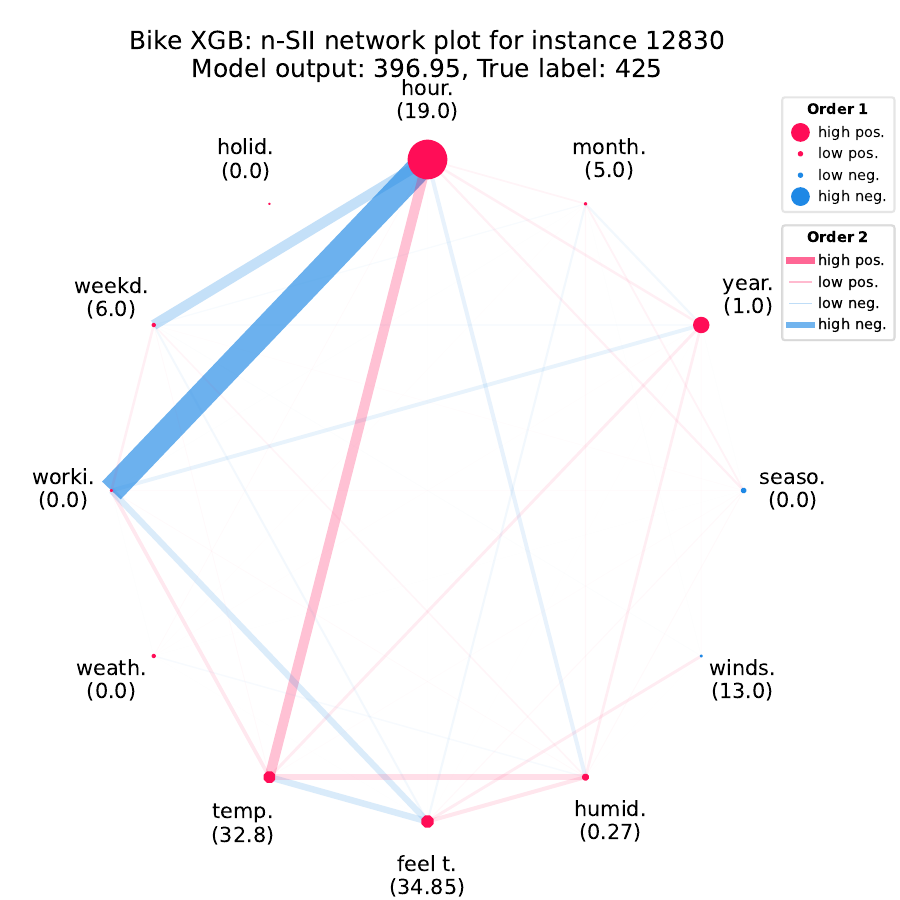}
    \\
    \includegraphics[width=0.45\textwidth]{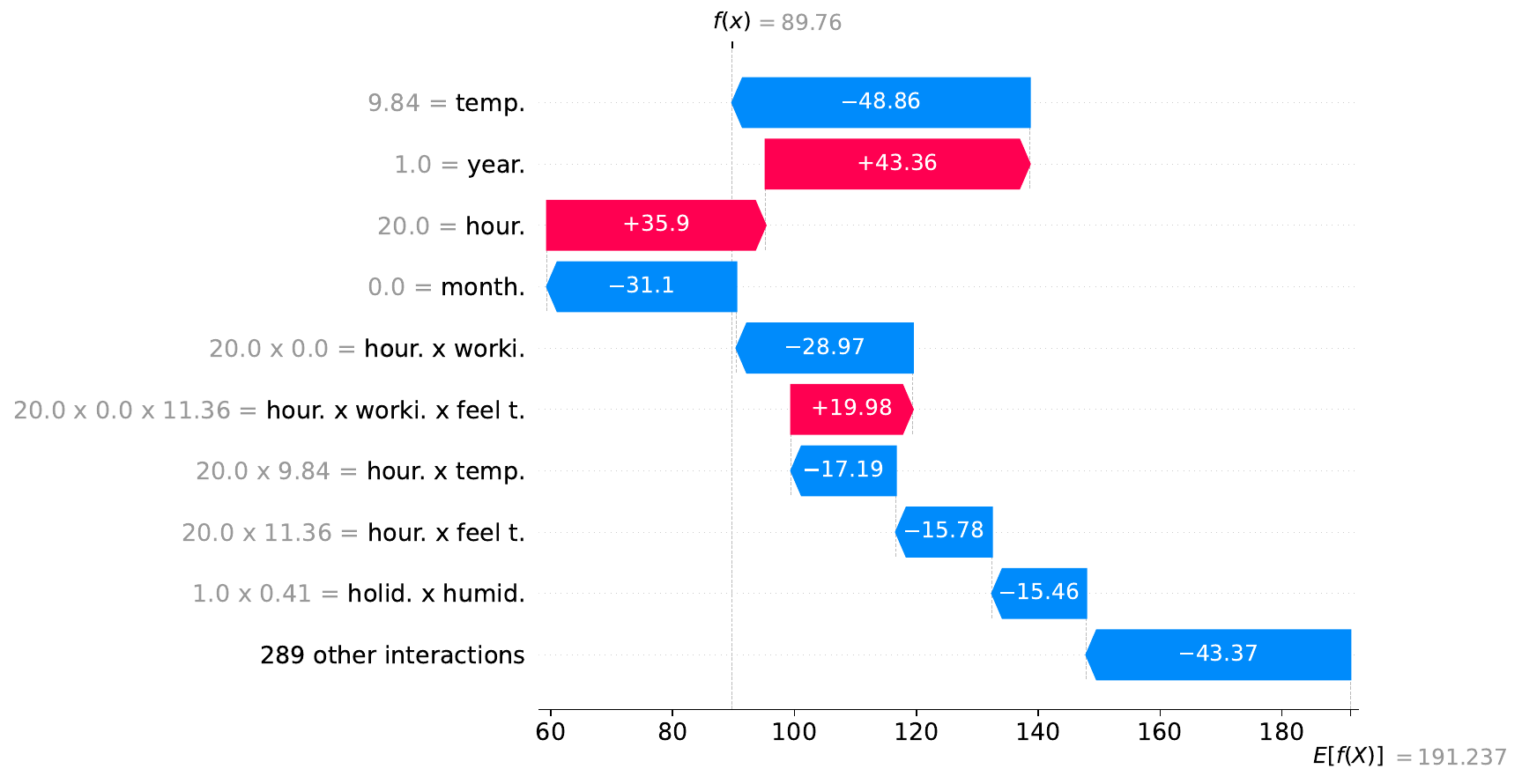}
    \includegraphics[width=0.45\textwidth]{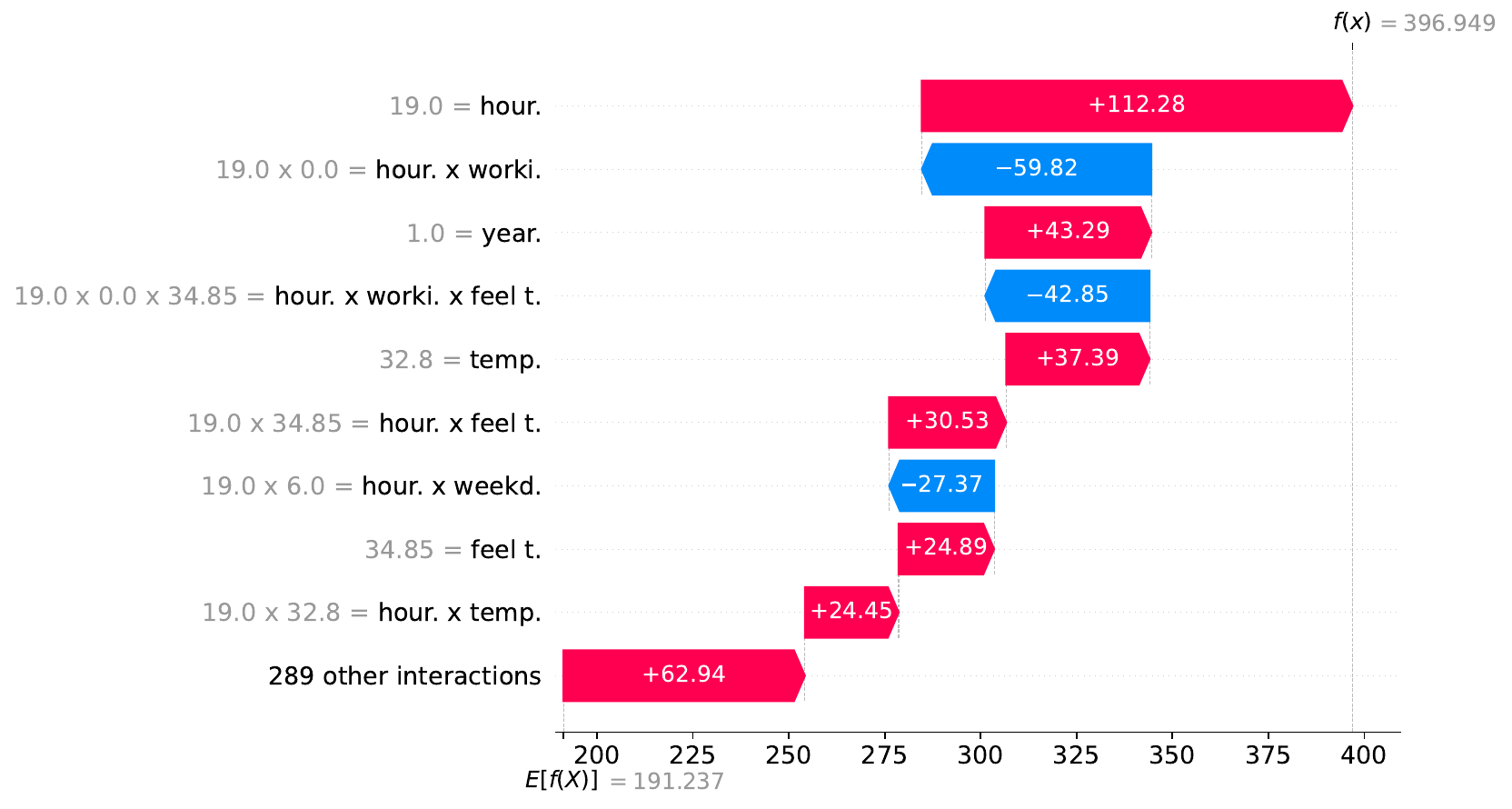}
    \caption{n-SII scores for order $s_0=2$ in a network plot (top) and $s_0=3$ in a waterfall chart (bottom) for two randomly selected instances of the \emph{Bike} dataset predicted with XGB}
    \label{fig_app_bike}
\end{figure*}

We display n-SII interaction effects up to order $s_0=2$ in a network plot and effects up to order $s_0=3$ in a waterfall chart for two randomly selected instances of the \emph{Bike} dataset predicted with XGB.
The results are shown in Figure~\ref{fig_app_bike}.

\clearpage
\subsubsection{COMPAS Dataset}
\begin{figure*}[h]
    \centering
    \includegraphics[width=0.45\textwidth]{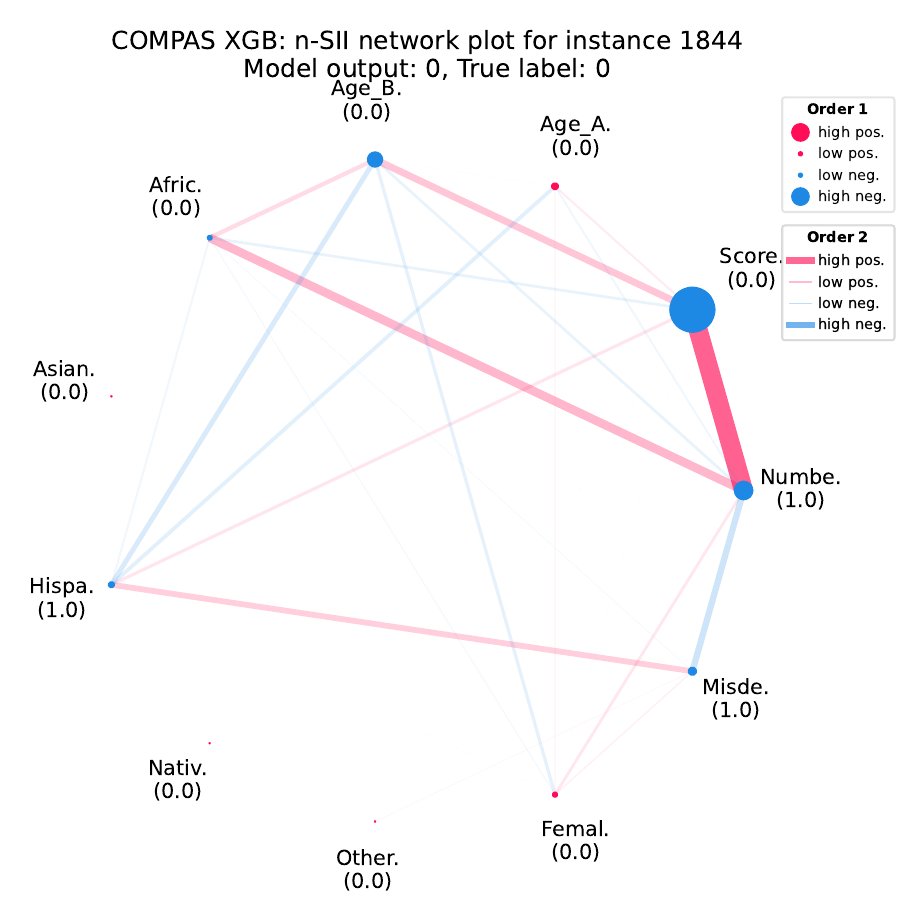}
    \includegraphics[width=0.45\textwidth]{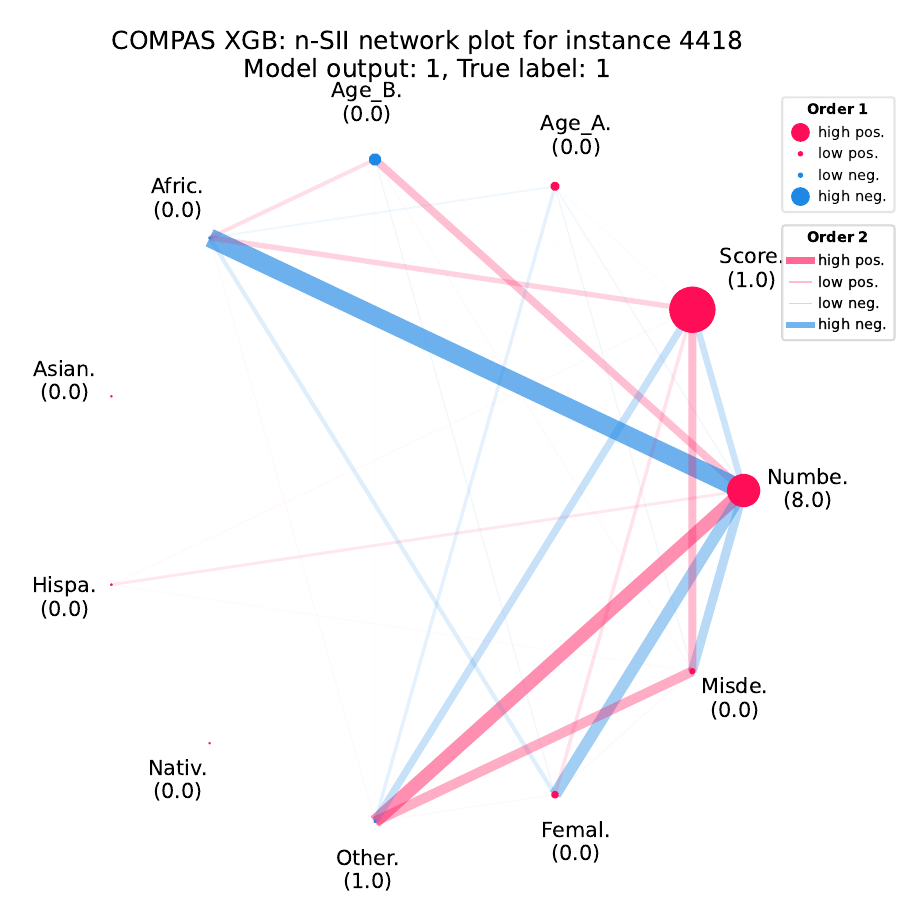}
    \\
    \includegraphics[width=0.45\textwidth]{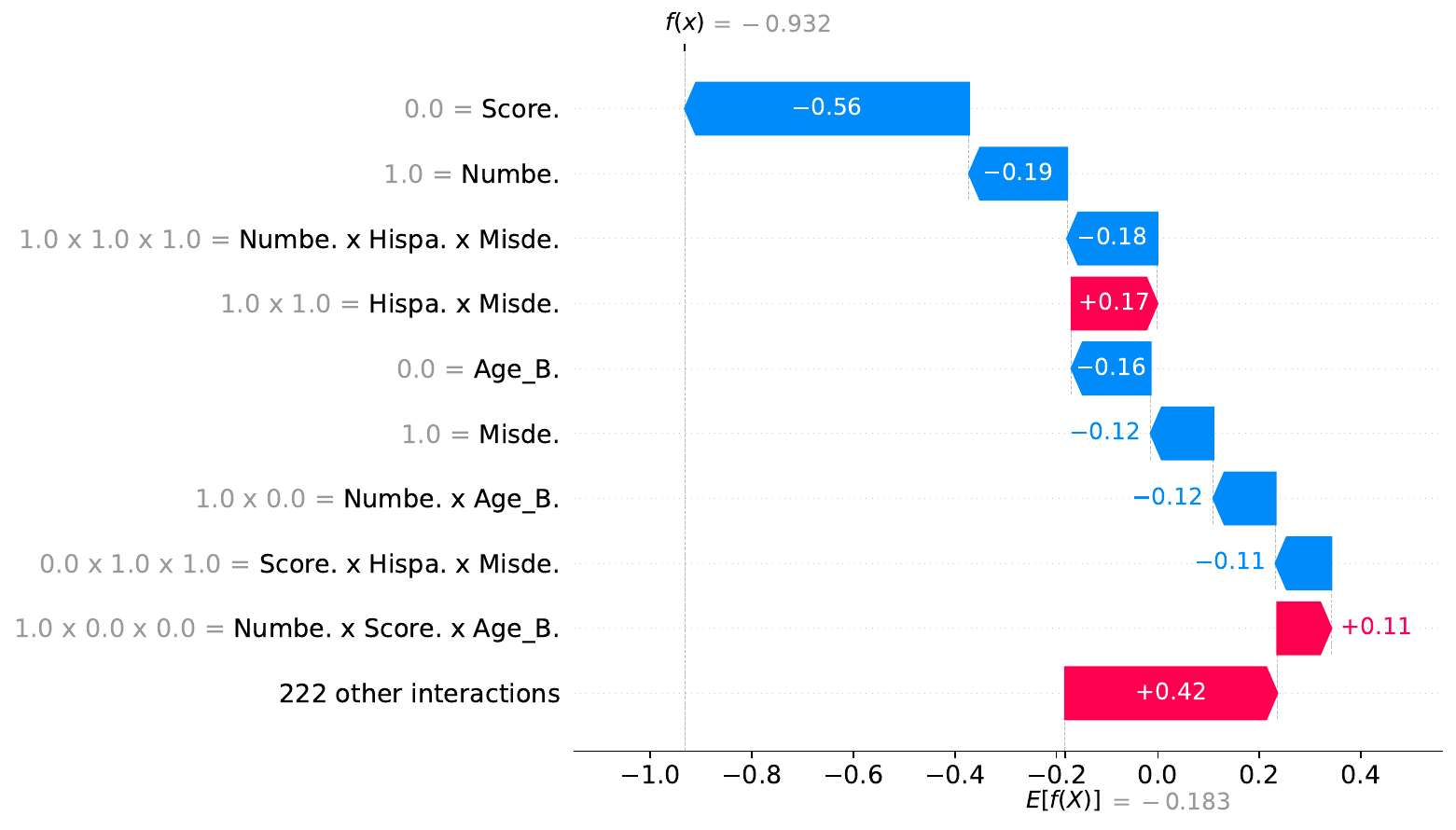}
    \includegraphics[width=0.45\textwidth]{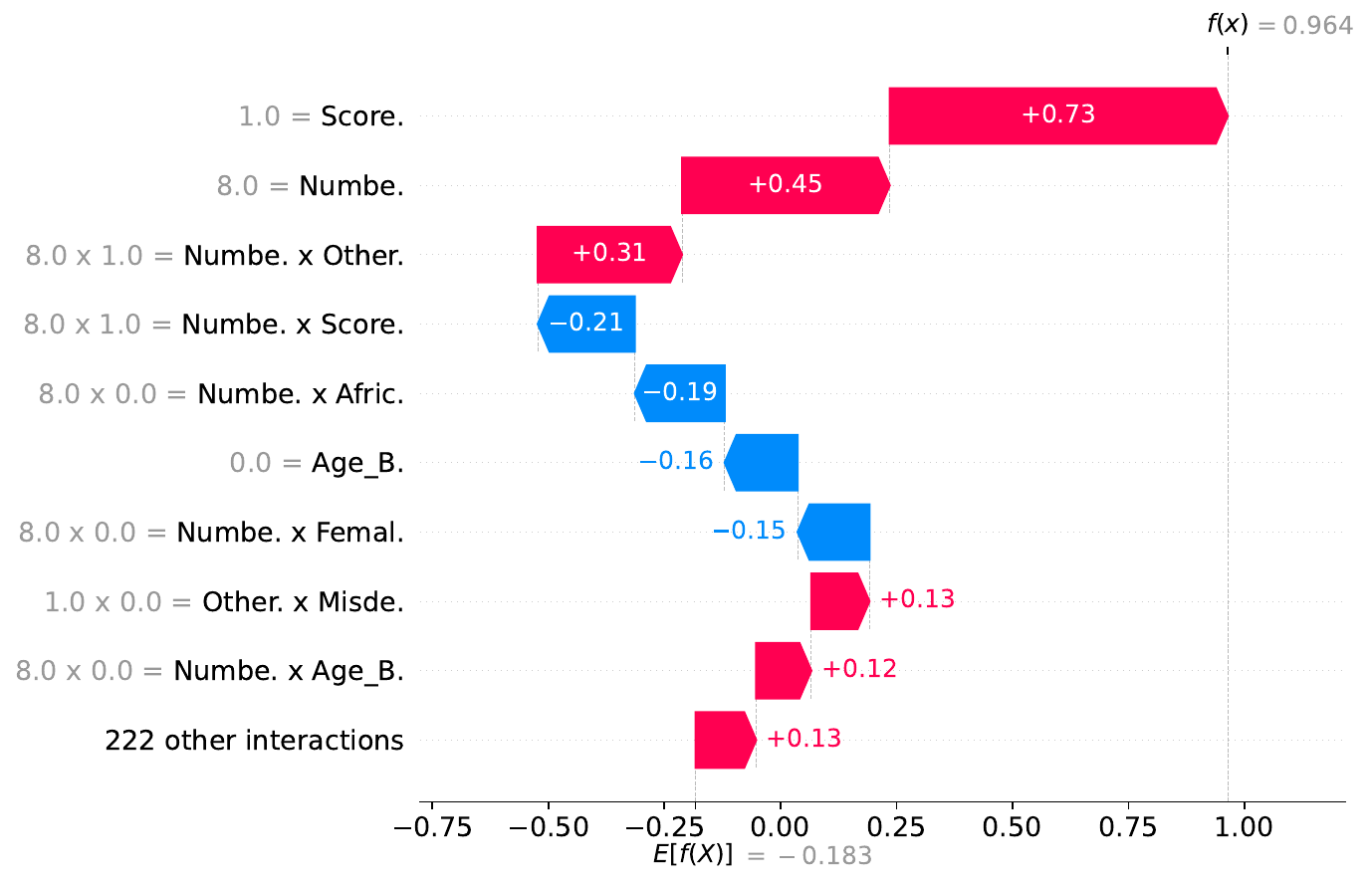}
    \caption{n-SII scores for order $s_0=2$ in a network plot (top) and $s_0=3$ in a waterfall chart (bottom) for two randomly selected instances of the \emph{COMPAS} dataset predicted with a GBT}
    \label{fig_app_compas}
\end{figure*}

We display n-SII interaction effects up to order $s_0=2$ in a network plot and effects up to order $s_0=3$ in a waterfall chart for two randomly selected instances of the \emph{COMPAS} dataset predicted with a GBT.
The results are shown in Figure~\ref{fig_app_compas}.

\clearpage
\subsubsection{Titanic Dataset}
\begin{figure*}[h]
    \centering
    \includegraphics[width=0.45\textwidth]{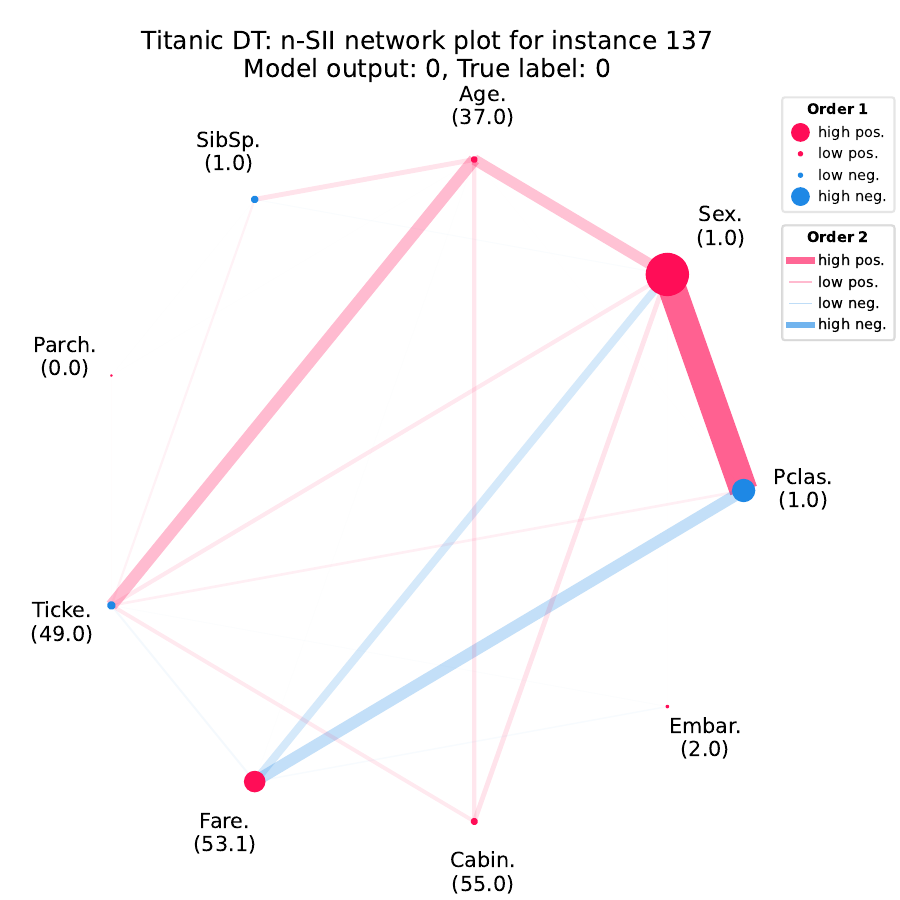}
    \includegraphics[width=0.45\textwidth]{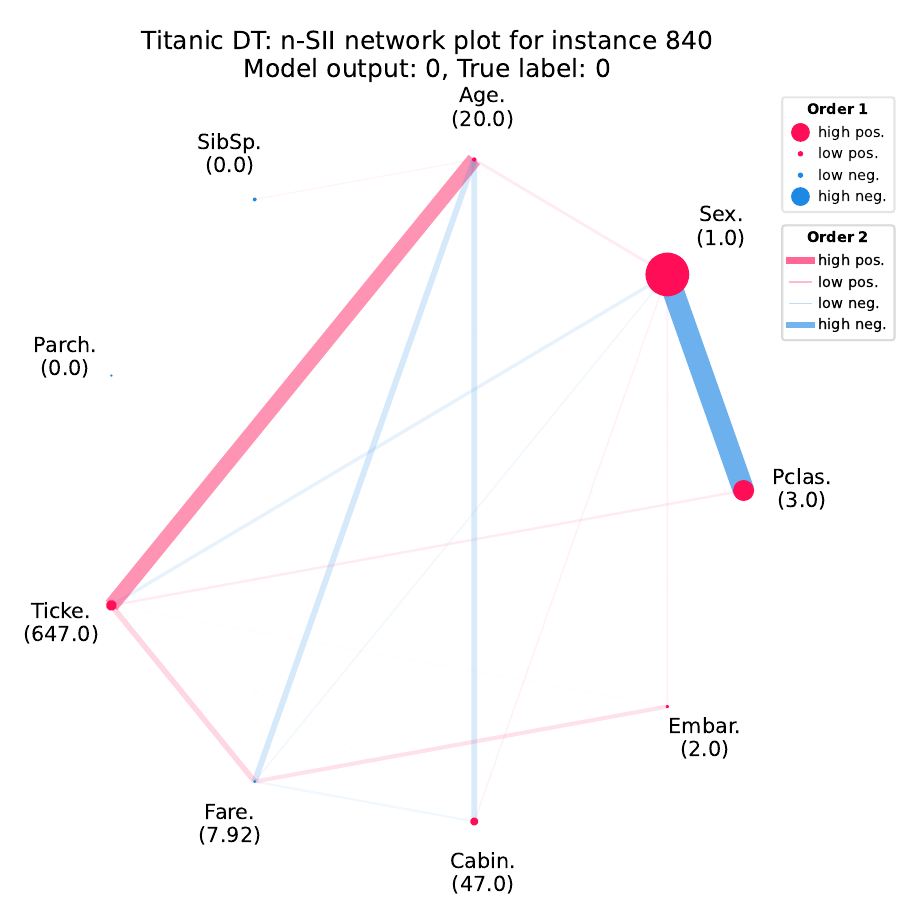}
    \\
    \includegraphics[width=0.45\textwidth]{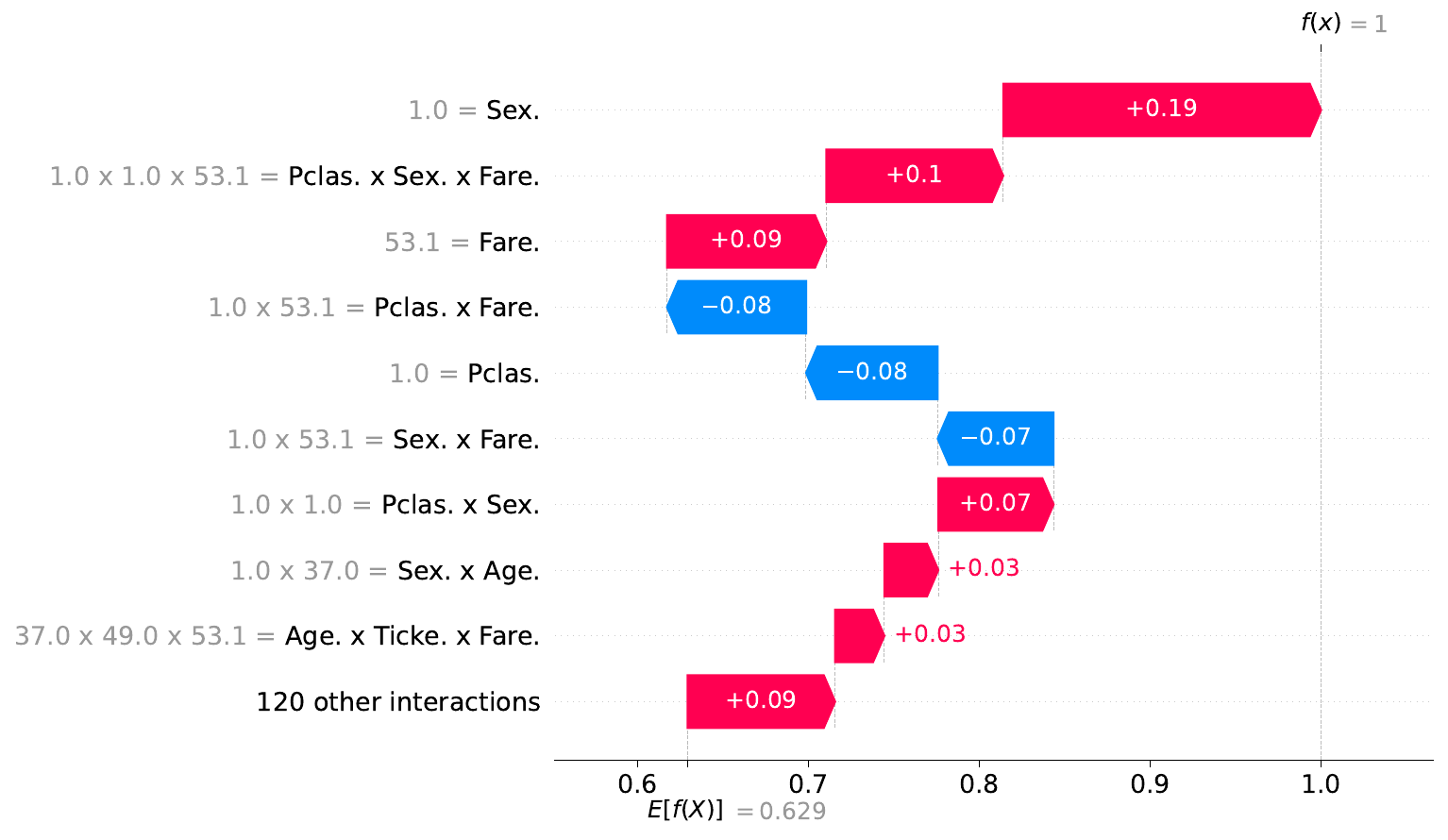}
    \includegraphics[width=0.45\textwidth]{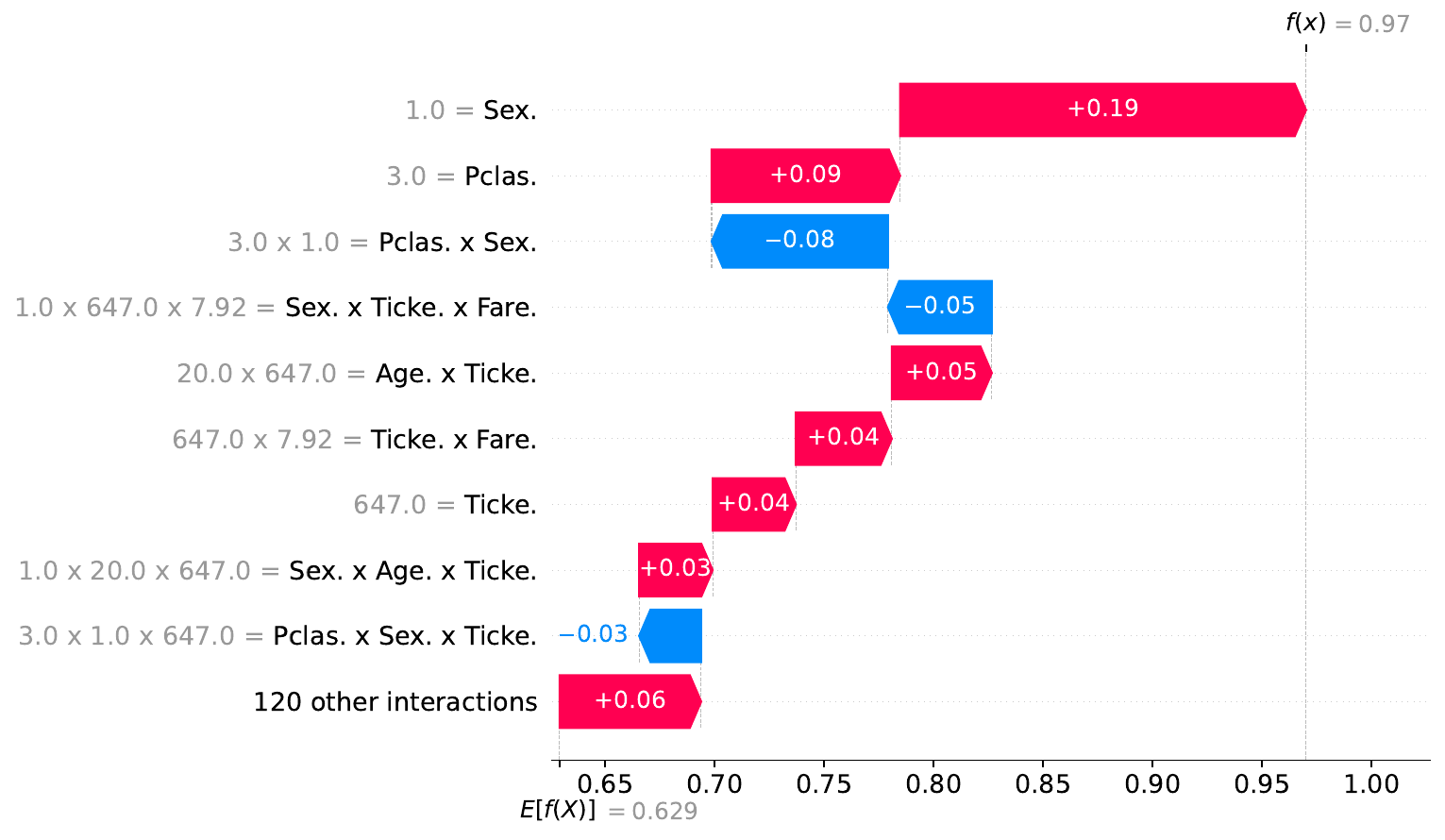}
    \caption{n-SII scores for order $s_0=2$ in a network plot (top) and $s_0=3$ in a waterfall chart (bottom) for two randomly selected instances of the \emph{Titanic} dataset predicted with a DT}
    \label{fig_app_titanic}
\end{figure*}

We display n-SII interaction effects up to order $s_0=2$ in a network plot and effects up to order $s_0=3$ in a waterfall chart for two randomly selected instances of the \emph{Titanic} dataset predicted with a DT.
The results are shown in Figure~\ref{fig_app_titanic}.

\clearpage
\subsubsection{California}
\begin{figure*}[h]
    \centering
    \includegraphics[width=0.45\textwidth]{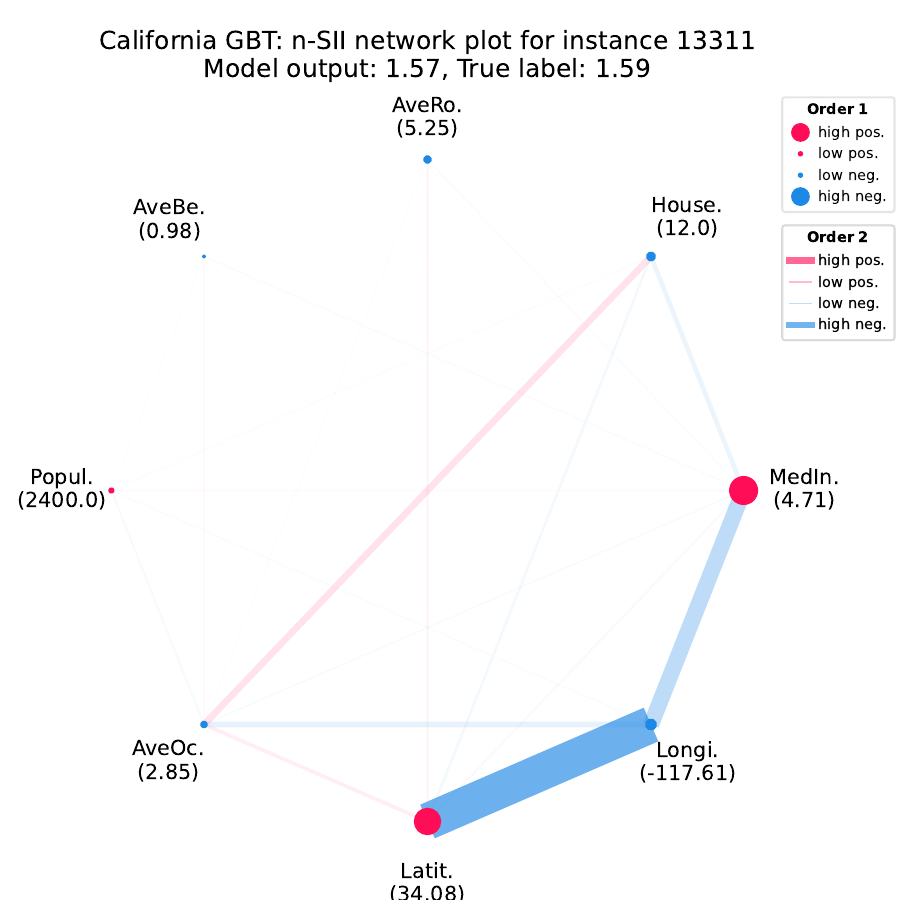}
    \includegraphics[width=0.45\textwidth]{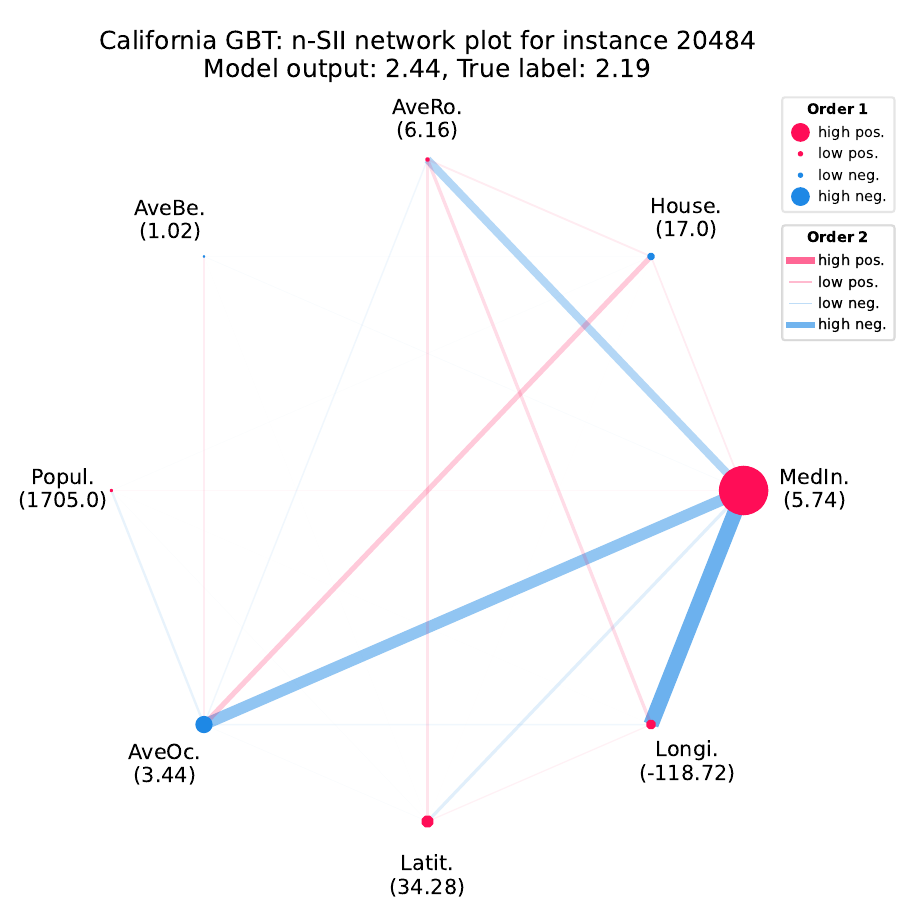}
    \\
    \includegraphics[width=0.45\textwidth]{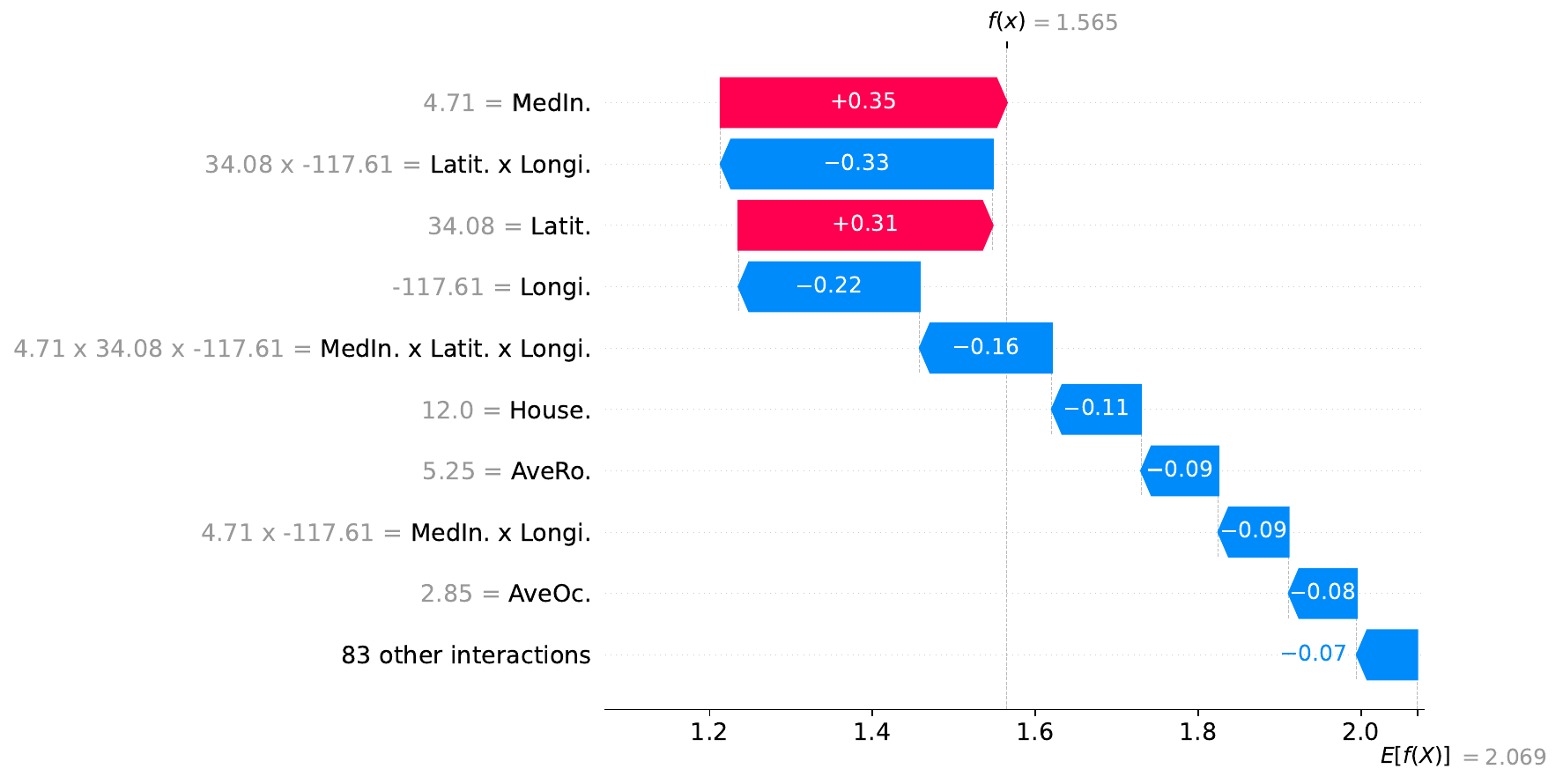}
    \includegraphics[width=0.45\textwidth]{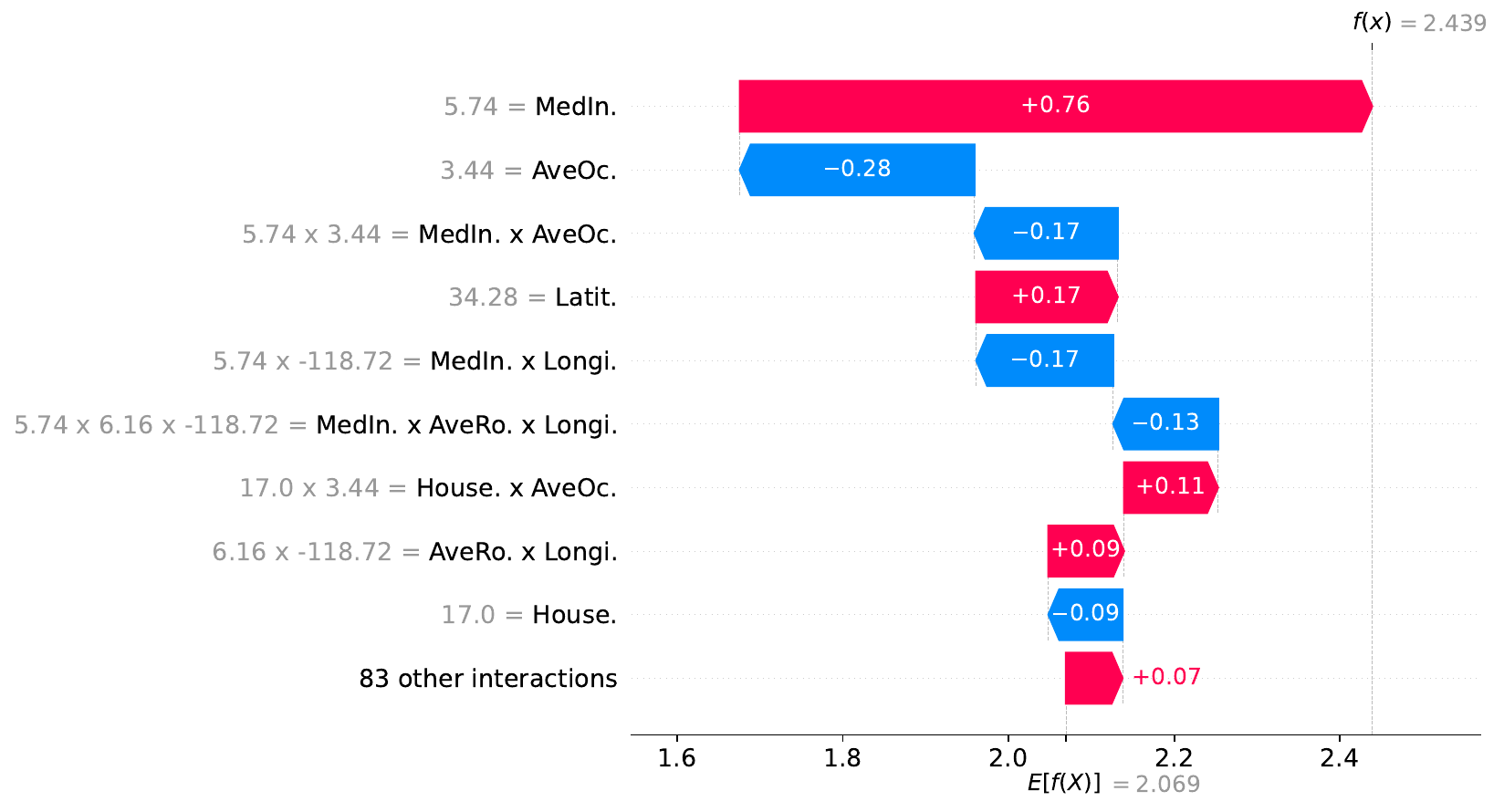}
    \caption{n-SII scores for order $s_0=2$ in a network plot (top) and $s_0=3$ in a waterfall chart (bottom) for two randomly selected instances of the \emph{California} dataset predicted with a GBT}
    \label{fig_app_california}
\end{figure*}

We display n-SII interaction effects up to order $s_0=2$ in a network plot and effects up to order $s_0=3$ in a waterfall chart for two randomly selected instances of the \emph{California} dataset predicted with a GBT.
The results are shown in Figure~\ref{fig_app_california}.

\end{document}